\documentclass[11pt]{article}

\usepackage[final]{acl}

\usepackage{times}
\usepackage{latexsym}

\usepackage[T1]{fontenc}

\usepackage[utf8]{inputenc}

\usepackage{microtype}

\usepackage{inconsolata}

\usepackage{booktabs}
\usepackage{amsfonts}
\usepackage{amsmath}
\usepackage{amssymb}
\usepackage{amsthm}
\usepackage{graphicx}
\usepackage[table]{xcolor}
\usepackage{multirow}
\usepackage{subcaption}
\usepackage{algorithm}
\usepackage{algorithmic}
\usepackage{enumitem}
\usepackage{listings}
\usepackage{tabularx}
\usepackage{array}
\usepackage{pifont}
\usepackage{fontawesome5}
\usepackage{soul}
\definecolor{lightyellow}{rgb}{1,1,0.7}
\sethlcolor{lightyellow}
\usepackage{tikz}
\usetikzlibrary{positioning, shapes.geometric, shapes.misc}
\usepackage{pgfplots}
\usepackage{subcaption}
\usepackage{tablefootnote}
\usepackage[title]{appendix}
\usepackage{titletoc}
\usepackage{etoolbox} 
\usepackage{placeins}

\definecolor{myblue}{RGB}{31, 119, 180}
\definecolor{myred}{RGB}{214, 39, 40}
\definecolor{mygreen}{RGB}{44, 160, 44}
\definecolor{mygray}{RGB}{128, 128, 128}

\usepackage[most]{tcolorbox}
\usepackage{mdframed}
\tcbuselibrary{skins}

\title{Reflection in the Dark: Exposing and Escaping the Black Box in\\ Reflective Prompt Optimization}

\author{
  \textbf{Shiyan Liu}$^{1,2}$\thanks{Equal contribution.}, 
  \textbf{Qifeng Xia}$^{2}$\footnotemark[\value{footnote}], 
  \textbf{Qiyun Xia}$^{3}$\footnotemark[\value{footnote}],
  \textbf{Yisheng Liu}$^{2}$, 
  \textbf{Xinyu Yu}$^{2}$, 
  \textbf{Rui Qu}$^{2}$ \\
  $^1$University of California, Berkeley \\
  $^2$Huazhong University of Science and Technology \\
  $^3$Hefei University of Technology \\
  \texttt{shiyanliu@berkeley.edu, shyl@hust.edu.cn}
}

\begin{document}
\maketitle

\begin{abstract}

Automatic prompt optimization (APO) has emerged as a powerful paradigm for improving LLM performance without manual prompt engineering. Reflective APO methods such as GEPA iteratively refine prompts by diagnosing failure cases, but the optimization process remains black-box and label-free, leading to uninterpretable trajectories and systematic failure. We identify and empirically demonstrate four limitations: on GSM8K with a defective seed, GEPA degrades accuracy from 23.81\% to 13.50\%. We propose VISTA, a multi-agent APO framework that decouples hypothesis generation from prompt rewriting, enabling semantically labeled hypotheses, parallel minibatch verification, and interpretable optimization trace. A two-layer explore-exploit mechanism combining random restart and epsilon-greedy sampling further escapes local optima. VISTA recovers accuracy to 87.57\% on the same defective seed and consistently outperforms baselines across all conditions on GSM8K and AIME2025.

\end{abstract}

\section{Introduction}
\label{sec:intro}

Large language models (LLMs) have demonstrated remarkable capabilities across diverse tasks~\cite{brown2020fewshot, wei2022cot, kojima2022large, wang2022selfconsistency}, yet their performance remains highly sensitive to prompt design: small changes in wording or instruction order can lead to dramatic differences in output quality. Manual prompt engineering is labor-intensive and requires extensive trial and error, motivating a growing body of work on Automatic Prompt Optimization (APO)~\cite{ye2023prompt, chen2023instructzero, sahoo2024systematic}.

\begin{figure}[t]
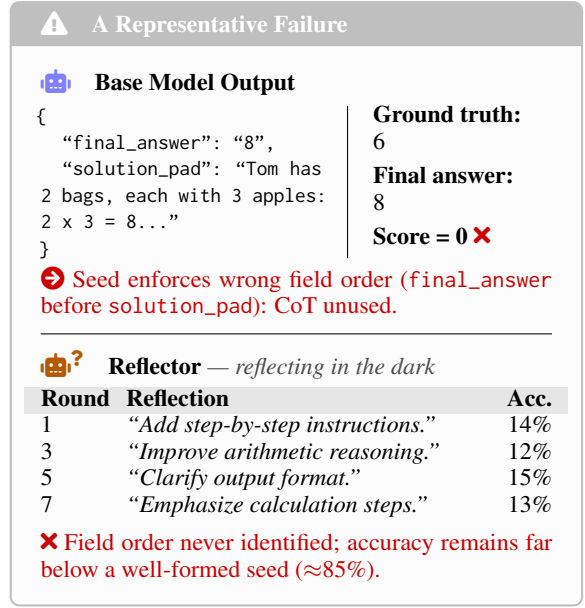

\begin{center}
\begin{minipage}{0.98\linewidth}
\begin{tcolorbox}[
    title={\faExclamationTriangle\quad A Representative Failure},
    fonttitle=\bfseries\small,
    colback=gray!2,
    colframe=gray!50,
    boxrule=0.8pt,
    left=8pt, right=8pt, top=6pt, bottom=6pt,
]
\small
\textcolor{blue!50}{\faRobot}\quad\textbf{Base Model Output}\\[2pt]
\begin{minipage}[t]{0.55\linewidth}
{\fontsize{8}{9}\selectfont\ttfamily
\{\\
\hspace*{1em}``final\_answer'': ``8'',\\
\hspace*{1em}``solution\_pad'': ``Tom has 2 bags, each with 3 apples: 2 x 3 = 8...''\\
\}
}
\end{minipage}%
\hfill%
\vrule%
\hfill%
\begin{minipage}[t]{0.35\linewidth}
\textbf{Ground truth:} \\ 6\\[3pt]
\textbf{Final answer:} \\ 8\\[3pt]
\textbf{Score = 0}\ \textcolor{red!80!black}{\faTimes}
\end{minipage}

\vspace{2pt}
\textcolor{red!80!black}{\faArrowCircleRight\ Seed enforces wrong field order (\texttt{final\_answer} before \texttt{solution\_pad}): CoT unused.}

\medskip\hrule\medskip

\textcolor{orange!70!black}{\faRobot}\textcolor{orange!70!black}{\textsuperscript{\tiny\faQuestion}}\quad\textbf{Reflector} \textcolor{black!70}{\textit{--- reflecting in the dark}}\\[2pt]
\begin{tabular*}{\linewidth}{@{}l@{\hspace{6pt}}p{0.7\linewidth}@{\extracolsep{\fill}}r@{}}
\rowcolor{gray!20}
\textbf{Round} & \textbf{Reflection} & \textbf{Acc.} \\
1 & \textit{``Add step-by-step instructions.''} & 14\% \\
3 & \textit{``Improve arithmetic reasoning.''} & 12\% \\
5 & \textit{``Clarify output format.''} & 15\% \\
7 & \textit{``Emphasize calculation steps.''} & 13\% \\
\end{tabular*}\\[4pt]
\textcolor{red!80!black}{\faTimes\ Field order never identified; accuracy remains far below a well-formed seed ($\approx$85\%).}
\end{tcolorbox}
\end{minipage}
\end{center}
\vspace{-0.5em}
\caption{A representative failure of GEPA under the defective seed.}
\label{case_intro}
\vspace{-1em}
\end{figure}

APO addresses this bottleneck by iteratively refining prompts based on task feedback, replacing human intuition with algorithmic search. Early approaches such as OPRO~\cite{yang2023opro} and ProTeGi~\cite{pryzant2023protegi} demonstrated that LLMs can serve as effective optimizers, proposing prompt edits in natural language guided by performance signals. More recently, reflective APO methods such as GEPA~\cite{agrawal2025gepa} have pushed this further by combining natural language reflection with genetic evolution and Pareto-based candidate selection.

Despite their promise, we identify a fundamental limitation shared by reflective APO methods: the optimization process is entirely black-box and label-free. Failure diagnosis and prompt rewriting are collapsed into a single reflection step, producing no record of what root cause was attributed, no semantic structure on the optimization trajectory, and no mechanism to detect when the search has been trapped from the start. Without this structure, the optimizer has no sense of where it started, what it can diagnose, where it has been, or whether its results will generalize. Figure~\ref{case_intro} illustrates a representative failure. Concretely, GEPA with its official, defective seed prompt on GSM8K~\cite{cobbe2021gsm8k} degrades accuracy from 23.81\% to 13.50\%, and the actual root cause remains unaddressed across all optimization rounds.

We formalize this as four systematic limitations that form a causal chain:
\textbf{[L1] \textsc{Seed trap}}: optimization is sensitive to seed prompts, which silently constrain the search space and trap it in defective regions.
\textbf{[L2] \textsc{Attribution blindspot}}: the reflector's attribution space is doubly bounded---by its prior distribution and by its own capability---systematically missing root causes outside either bound.
\textbf{[L3] \textsc{Trajectory opacity}}: even when attribution points in the right direction, the optimization trajectory is entirely label-free, making the full evolution uninterpretable and preventing the optimizer from accumulating directional experience.
\textbf{[L4] \textsc{Transfer fragility}}: optimized prompts are model-specific, failing silently when transferred across base models.

\begin{figure}[t]
    \centering
    \includegraphics[width=\linewidth]{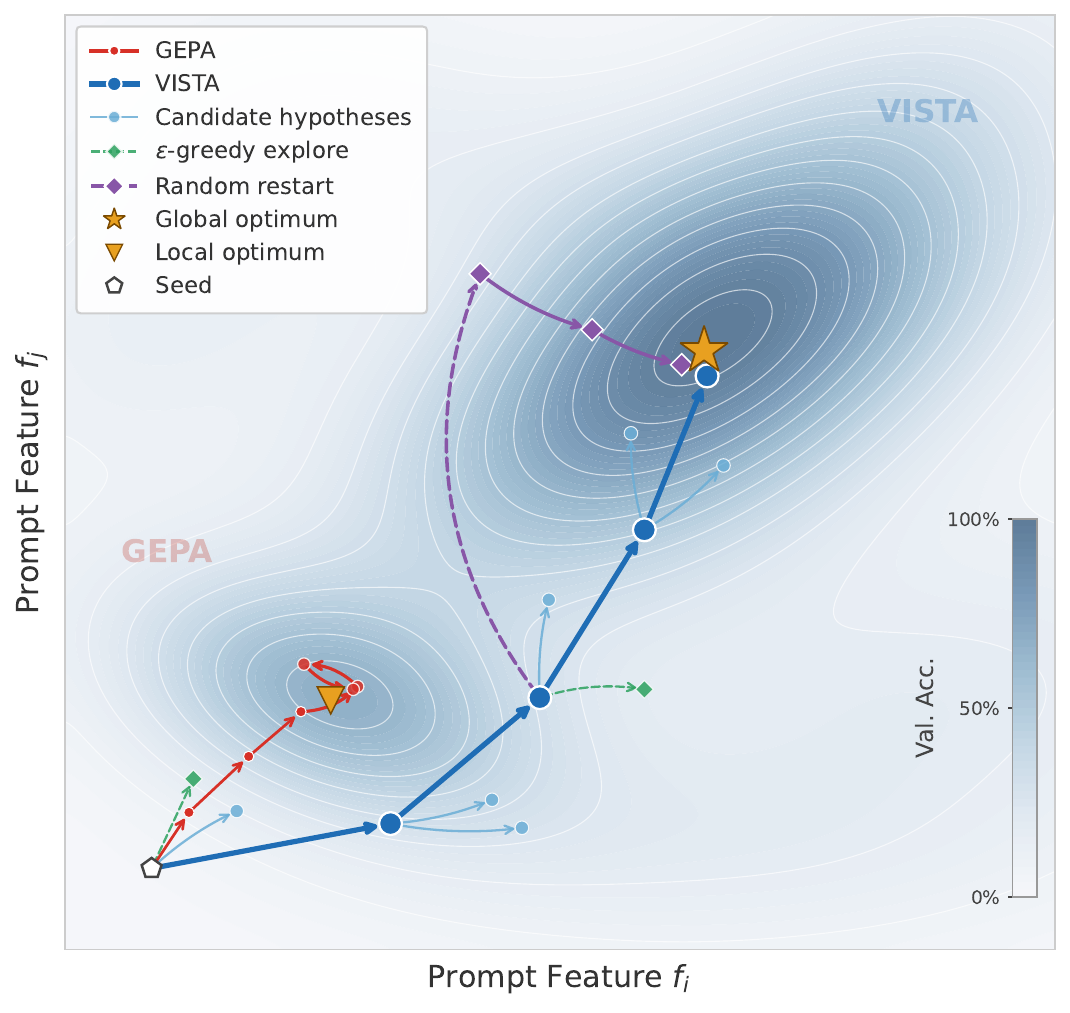}
    \caption{Conceptual illustration of optimization trajectories under the defective seed.}
    \label{fig:landscape}
    \vspace{-1em}
\end{figure}

To address these limitations, we propose \textbf{VISTA} (\textbf{V}erifiable, \textbf{I}nterpretable, \textbf{S}emantic-\textbf{T}r\textbf{A}ce Prompt Optimization). The key insight is to decouple hypothesis generation from prompt rewriting via a multi-agent design: a hypothesis agent proposes semantically labeled hypotheses guided by a heuristic set, while a reflection agent rewrites the prompt targeting each hypothesis independently. Parallel minibatch validation then selects the best hypothesis, constructing an interpretable optimization trace that makes every step auditable and directional. A two-layer explore-exploit mechanism~\cite{sutton2018reinforcement}---combining random restart for escaping seed-induced traps and epsilon-greedy sampling for hypothesis diversity---further ensures robust global search. On the same defective seed, VISTA recovers accuracy to 87.57\%. Figure~\ref{fig:landscape} illustrates how GEPA becomes trapped under a defective seed, whereas VISTA is able to escape.

Our contributions are as follows:
\begin{itemize}
    \item We identify and formalize four systematic limitations of reflective APO under a unified interpretability-blindspot framework.
    \item We propose VISTA, a multi-agent APO framework that decouples hypothesis generation from prompt rewriting, enabling verifiable and interpretable optimization traces.
    \item We provide comprehensive experiments on GSM8K and AIME2025~\cite{aime} validating both the proposed limitations and the effectiveness of VISTA.
\end{itemize}

\section{Related Work}
\label{sec:related}

\subsection{Automatic Prompt Optimization}

Prompt engineering has long relied on manual design~\cite{brown2020fewshot, wei2022cot}, 
but recent work has shifted toward automated approaches. 
Instruction induction methods~\cite{honovich2022instruction, zhou2023largelanguage} 
generate candidate prompts from input-output examples. 
Optimization-based methods treat prompt search as a black-box problem: 
OPRO~\cite{yang2023opro} uses an LLM as a meta-optimizer guided by past scores, 
while APE~\cite{zhou2023largelanguage} searches over instruction candidates via sampling. 
Gradient-inspired methods such as ProTeGi~\cite{pryzant2023protegi} and 
TextGrad~\cite{yuksekgonul2024textgrad} compute textual ``gradients'' from failure feedback 
to guide prompt updates. 
More recently, DSPy~\cite{khattab2024dspy} and MIPROv2~\cite{opsahlong2024miprov2} 
extend APO to multi-module compound systems, jointly optimizing prompts and few-shot 
demonstrations. 

Evolutionary approaches~\cite{fernando2023promptbreeder, guo2023connecting} apply self-referential mutation to prompt search.
GEPA~\cite{agrawal2025gepa} further advances this paradigm by combining 
reflective prompt mutation with Pareto-based candidate selection, achieving strong 
sample efficiency across diverse tasks. EvoX~\cite{liu2026evox} proposes meta-evolution 
of the search strategy itself, jointly optimizing the prompt and the optimization policy. 
While these methods demonstrate strong empirical results, they share a common limitation: 
attribution generation and prompt rewriting are entangled in a single black-box step, 
leaving the optimization process without interpretable structure or verifiable 
root-cause attribution.

\subsection{LLM Self-Correction and Its Limits}

A prominent line of work explores iterative self-refinement of LLM outputs. 
Self-Refine~\cite{madaan2023selfrefine} and Reflexion~\cite{shinn2023reflexion} 
prompt LLMs to critique and revise their own outputs across multiple rounds, 
demonstrating improvements on a range of tasks. 
However, \citet{huang2023selfcorrect} provide a critical counterpoint: 
without access to external feedback, LLMs cannot reliably self-correct reasoning, 
as revisions are bounded by the same prior that produced the original error. 
Subsequent work has further characterized conditions under which self-correction 
succeeds or fails~\cite{olausson2023selfrepair, stechly2023gpt4, kamoi2024can}. 
These findings directly motivate VISTA: reflective APO diagnosis is 
subject to the same prior constraints, making external heuristics 
necessary.

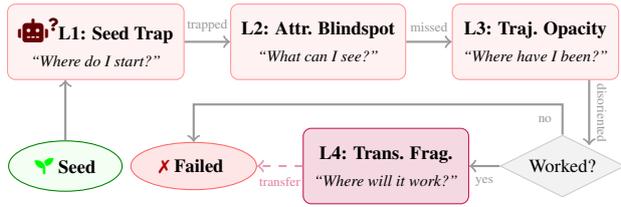
\begin{figure}[t]
\centering
\begin{tikzpicture}[
    node distance=0.6cm,
    every node/.style={font=\fontsize{7}{8}\selectfont},
    box/.style={rectangle, draw=red!40, fill=red!5, rounded corners=3pt,
                minimum width=1.8cm, minimum height=1.0cm, align=center,
                inner sep=4pt},
    purplebox/.style={rectangle, draw=purple!90, fill=purple!15, rounded corners=3pt,
                minimum width=1.8cm, minimum height=1.0cm, align=center,
                inner sep=4pt},
    seedbox/.style={ellipse, draw=green!50!black, fill=green!5,
                minimum width=1.4cm, minimum height=0.6cm, align=center},
    failbox/.style={ellipse, draw=red!60, fill=red!10,
                minimum width=1.4cm, minimum height=0.6cm, align=center},
    decbox/.style={diamond, draw=gray!40, fill=gray!10, aspect=2.0,
                minimum width=1.6cm, minimum height=0.8cm, align=center,
                inner sep=2pt},
    arr/.style={->, thick, black!40},
    darr/.style={->, dashed, thick, purple!40},
]

\node[box] (l1) {
    {\fontsize{9}{10}\selectfont\textcolor{red!40!black}{\faRobot}}%
\raisebox{4pt}{\fontsize{5}{6}\selectfont\textcolor{red!30!black}{\faQuestion}}\textbf{L1: Seed Trap}\\[2pt]
    {\fontsize{6}{7}\selectfont\itshape ``Where do I start?''}
};
\node[box, right=0.6cm of l1] (l2) {
    \textbf{L2: Attr. Blindspot}\\[2pt]
    {\fontsize{6}{7}\selectfont\itshape ``What can I see?''}
};
\node[box, right=0.6cm of l2] (l3) {
    \textbf{L3: Traj. Opacity}\\[2pt]
    {\fontsize{6}{7}\selectfont\itshape ``Where have I been?''}
};
\node[seedbox, below=0.8cm of l1, xshift=-0.4cm] (seed) {\textcolor{green}{\faSeedling} \textbf{Seed}};
\node[failbox, right=0.1cm of seed] (fail) {
    \textcolor{red!70!black}{\ding{55}}\ \textbf{Failed}
};
\node[purplebox, right=0.6cm of fail] (l4) {
    \textbf{L4: Trans. Frag.}\\[2pt]
    {\fontsize{6}{7}\selectfont\itshape ``Where will it work?''}
};

\node[decbox, right=0.4cm of l4] (decision) {Worked?};

\draw[arr] (l1) -- node[above, font=\fontsize{5}{6}\selectfont, text=black!40]
    {trapped} (l2);
\draw[arr] (l2) -- node[above, font=\fontsize{5}{6}\selectfont, text=black!40]
    {missed} (l3);

\draw[arr] (seed.north) -- (seed.north |- l1.south);

\coordinate (fin) at ([xshift=10pt, yshift=-4pt]decision.north);

\draw[arr] (fin |- l3.south) -- (fin)
    node[midway, rotate=-90, yshift=-1pt, anchor=south, font=\fontsize{5}{6}\selectfont, text=black!40] {disoriented};

\draw[arr] (decision.west) -- node[below, font=\fontsize{5}{6}\selectfont, text=black!40]
    {yes} (l4.east);

\draw[darr] (l4.west) -- node[below, font=\fontsize{5}{6}\selectfont, text=purple!50]
    {transfer} (fail.east);

\draw[arr] (decision.north) -- ++(0,0.4) 
    node[midway, left, font=\fontsize{5}{6}\selectfont, text=black!40] {no} 
    -| (fail.north);

\end{tikzpicture}
\caption{Four systematic limitations of reflective APO. L1--L3 form a causal chain, and L4 can apply even when optimization succeeds.}
\label{fig:limitations}
\vspace{-1em}
\end{figure}

\section{Diagnosing the Black Box: Four Limitations}
\label{sec:analysis}

We now expand on each limitation introduced in Section~\ref{sec:intro}. These four limitations form a progressive indictment of reflective APO: the seed constrains where search begins (L1), the reflector constrains what can be hypothesized (L2), and the absence of semantic structure constrains what the optimizer can learn from its own trajectory (L3). Even if L1–L3 were fully resolved and optimization succeeded, the result carries no guarantee of generalization across base models (L4). Figure~\ref{fig:limitations} illustrates how they manifest as a causal chain.

\subsection{L1: Seed Trap}
\label{sec:limitations}

The seed prompt implicitly defines the initial search region. When the seed contains structural defects---such as incorrect output field ordering, malformed schema constraints, or contradictory instructions---the optimizer inherits these defects as implicit constraints. Because reflective APO produces no record of which seed-level assumptions are being carried forward, it has no mechanism to identify or question them.

Consider the official GEPA~\cite{agrawal2025gepa} seed prompt (Figure~\ref{case_intro}). Its output schema specifies \texttt{final\_answer} before \texttt{solution\_pad}, causing the base model to output its final answer \emph{before} its Chain-of-Thought (CoT) reasoning, effectively preventing CoT from influencing the answer. This ordering constraint is silently inherited throughout all optimization rounds and never flagged as a candidate root cause.




\begin{figure}
\begin{tcolorbox}[
    title={\small Case: Attribution Blindspot},
    fonttitle=\bfseries\small,
    colback=gray!2,
    colframe=gray!50,
    boxrule=0.8pt,
    left=6pt, right=6pt, top=4pt, bottom=4pt
]
\small
\textbf{Optimized prompts across rounds:}\\[4pt]
\textcolor{black!70}{Rd.1:} {\fontsize{8}{9}\selectfont\ttfamily ``You are a math assistant. Solve the problem \hl{step by step}. Output in JSON.''}\\[3pt]
\textcolor{black!70}{Rd.3:} {\fontsize{8}{9}\selectfont\ttfamily ``You are a math assistant. Solve the problem step by step. \hl{Verify each calculation.} Output in JSON.''}\\[3pt]
\textcolor{black!70}{Rd.5:} {\fontsize{8}{9}\selectfont\ttfamily ``You are a math assistant. \hl{Break the problem into parts.} Verify each calculation. Output in JSON.''}\\[3pt]
\textcolor{black!70}{Rd.7:} {\fontsize{8}{9}\selectfont\ttfamily ``You are a math assistant. Break the problem into parts. Verify each calculation. \hl{Explain your reasoning in detail before giving the final answer.} Output in JSON.''}\\[0pt]
\hrule\vspace{4pt}
\textcolor{red!80!black}{\faTimes}\; \textbf{Attribution Blindspot:} All attributions target reasoning quality; the structural root cause remains outside the reflector's attribution space across all rounds and reflector configurations.
\end{tcolorbox}
\vspace{-1em}
\caption{An illustrative case of attribution blindspot.}
\label{fig:attr-blind}
\vspace{-1em}
\end{figure}

\subsection{L2: Attribution Blindspot}
\label{attr-blindspot}

The reflector's attribution space is doubly bounded. First, structurally: the reflector can only propose root causes that fall within its prior distribution of plausible failure modes, systematically missing categories that are underrepresented or absent from that distribution. \citet{huang2023selfcorrect} demonstrate that LLMs cannot reliably self-correct without external feedback, as revisions are constrained by the same prior that produced the original error---reflective APO diagnosis is subject to the same bound. Second, by capability: a weaker reflector has a narrower effective attribution space, and even within categories it can in principle reach, attribution quality degrades with model capability. Both constraints are invisible to the optimizer.

\begin{figure}[t]
\centering
\begin{tikzpicture}
\begin{axis}[
    width=\columnwidth, height=5.5cm,
    xmin=0.5, xmax=5.0,
    ymin=0.5, ymax=6.5,
    xtick={1,2,3,4},
    xticklabels={
        \rotatebox{30}{\scriptsize Def. (Qwen)},
        \rotatebox{30}{\scriptsize Rep. (Qwen)},
        \rotatebox{30}{\scriptsize Min. (Qwen)},
        \rotatebox{30}{\scriptsize Def. (GPT)}
    },
    ytick={1,2,3,4,5,6},
    yticklabels={
        {\rotatebox{30}{\fontsize{6}{7}\selectfont\textcolor{red!80!black}{\textbf{Field Order.}}}},
        {\rotatebox{30}{\fontsize{6}{7}\selectfont Edge Cases}},
        {\rotatebox{30}{\fontsize{6}{7}\selectfont Domain Know.}},
        {\rotatebox{30}{\fontsize{6}{7}\selectfont Format/Syn.}},
        {\rotatebox{30}{\fontsize{6}{7}\selectfont Task Instr.}},
        {\rotatebox{30}{\fontsize{6}{7}\selectfont Reasoning}}
    },
    axis line style={color=black!30},
    tick style={color=black!30},
    tick label style={font=\fontsize{6}{7}\selectfont},
    axis lines=left,
    grid=major,
    grid style={dashed, gray!15},
    clip=false,
    xlabel={\fontsize{6}{7}\selectfont Seed},
    ylabel={\fontsize{6}{7}\selectfont Category},
    xlabel style={at={(axis description cs:1.0,0)}, anchor=north east,
                  font=\fontsize{6}{7}\selectfont, yshift=15pt},
    ylabel style={at={(axis description cs:0,1.0)}, anchor=south west,
                  rotate=-90, font=\fontsize{6}{7}\selectfont, xshift=7pt, yshift=-5pt},
]

\fill[red!5] (axis cs:0.5,0.65) rectangle (axis cs:5.0,1.35);
\draw[red!30, dashed] (axis cs:0.5,1.35) -- (axis cs:5.0,1.35);
\draw[red!30, dashed] (axis cs:0.5,0.65) -- (axis cs:5.0,0.65);

\addplot[only marks, mark=*, mark size={sqrt(70)*1.7},
    mark options={fill=red!50, draw=red!60, fill opacity=0.7},
    nodes near coords={\fontsize{6}{7}\selectfont\textcolor{white}{70\%}},
    every node near coord/.style={anchor=center}
] coordinates {(1, 5)};
\addplot[only marks, mark=*, mark size={sqrt(10)*2.2},
    mark options={fill=red!50, draw=red!60, fill opacity=0.7},
    nodes near coords={\fontsize{6}{7}\selectfont\textcolor{white}{10\%}},
    every node near coord/.style={anchor=center}
] coordinates {(1, 4)};
\addplot[only marks, mark=*, mark size={sqrt(20)*2.2},
    mark options={fill=red!50, draw=red!60, fill opacity=0.7},
    nodes near coords={\fontsize{6}{7}\selectfont\textcolor{white}{20\%}},
    every node near coord/.style={anchor=center}
] coordinates {(1, 3)};
\node[font=\fontsize{6}{7}\selectfont, text=red!60!black] at (axis cs:1, 1) {$\varnothing$};

\addplot[only marks, mark=*, mark size={sqrt(10)*2.2},
    mark options={fill=green!50!black, draw=green!60!black, fill opacity=0.7},
    nodes near coords={\fontsize{6}{7}\selectfont\textcolor{white}{10\%}},
    every node near coord/.style={anchor=center}
] coordinates {(2, 6)};
\addplot[only marks, mark=*, mark size={sqrt(75)*1.7},
    mark options={fill=green!50!black, draw=green!60!black, fill opacity=0.7},
    nodes near coords={\fontsize{6}{7}\selectfont\textcolor{white}{75\%}},
    every node near coord/.style={anchor=center}
] coordinates {(2, 5)};
\addplot[only marks, mark=*, mark size={sqrt(15)*2.2},
    mark options={fill=green!50!black, draw=green!60!black, fill opacity=0.7},
    nodes near coords={\fontsize{6}{7}\selectfont\textcolor{white}{15\%}},
    every node near coord/.style={anchor=center}
] coordinates {(2, 3)};
\node[font=\fontsize{6}{7}\selectfont, text=red!60!black] at (axis cs:2, 1) {$\varnothing$};

\addplot[only marks, mark=*, mark size={sqrt(14.29)*2.2},
    mark options={fill=blue!50, draw=blue!60, fill opacity=0.7},
    nodes near coords={\fontsize{6}{7}\selectfont\textcolor{white}{14\%}},
    every node near coord/.style={anchor=center}
] coordinates {(3, 6)};
\addplot[only marks, mark=*, mark size={sqrt(78.57)*1.7},
    mark options={fill=blue!50, draw=blue!60, fill opacity=0.7},
    nodes near coords={\fontsize{6}{7}\selectfont\textcolor{white}{79\%}},
    every node near coord/.style={anchor=center}
] coordinates {(3, 5)};
\addplot[only marks, mark=*, mark size={sqrt(7.14)*2.2},
    mark options={fill=blue!50, draw=blue!60, fill opacity=0.7},
    nodes near coords={\fontsize{6}{7}\selectfont\textcolor{white}{7\%}},
    every node near coord/.style={anchor=center}
] coordinates {(3, 2)};
\node[font=\fontsize{6}{7}\selectfont, text=red!60!black] at (axis cs:3, 1) {$\varnothing$};

\addplot[only marks, mark=*, mark size={sqrt(81.82)*1.7},
    mark options={fill=orange!70, draw=orange!80, fill opacity=0.7},
    nodes near coords={\fontsize{6}{7}\selectfont\textcolor{white}{82\%}},
    every node near coord/.style={anchor=center}
] coordinates {(4, 5)};
\addplot[only marks, mark=*, mark size={sqrt(18.18)*2.2},
    mark options={fill=orange!70, draw=orange!80, fill opacity=0.7},
    nodes near coords={\fontsize{6}{7}\selectfont\textcolor{white}{18\%}},
    every node near coord/.style={anchor=center}
] coordinates {(4, 3)};
\node[font=\fontsize{6}{7}\selectfont, text=red!60!black] at (axis cs:4, 1) {$\varnothing$};

\end{axis}
\end{tikzpicture}
\caption{GEPA attribution distribution on GSM8K (Qwen3-4B base, Qwen3-8B/GPT-4o-mini reflectors).}
\label{fig:asi_distribution}
\vspace{-0.5em}
\end{figure}
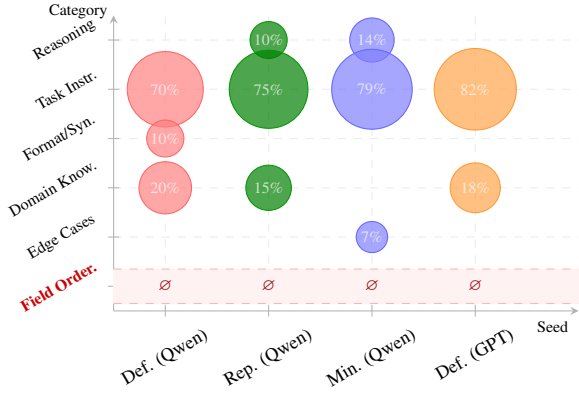

Analysis of GEPA's optimized prompts across all rounds confirms this: despite iterative optimization, no round ever corrects the field ordering defect (Figure~\ref{fig:attr-blind}). The reflector consistently attributes failures to reasoning errors, hallucinations, and instruction-following issues, never proposing structural attributions. Figure~\ref{fig:asi_distribution} shows the distribution of attribution categories across configurations---the actual root cause (\texttt{Field Ordering}) receives zero attributions across all configurations regardless of reflector strength, confirming that the blindspot is prior-structural rather than purely capability-driven.

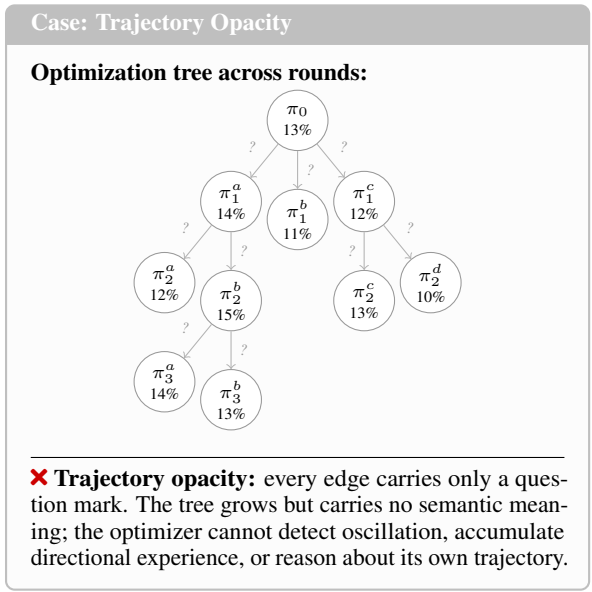
\begin{figure}
\begin{tcolorbox}[
    title={\small Case: Trajectory Opacity},
    fonttitle=\bfseries\small,
    colback=gray!2,
    colframe=gray!50,
    boxrule=0.8pt,
    left=6pt, right=6pt, top=4pt, bottom=4pt
]
\small
\textbf{Optimization tree across rounds:}\\[1pt]
\makebox[\linewidth][c]{
\begin{tikzpicture}[
    node distance=0.5cm and 0.5cm,
    every node/.style={font=\fontsize{7}{8}\selectfont},
    prompt/.style={circle, draw=black!40, fill=white, minimum size=0.8cm, inner sep=1pt, align=center},
    normaledge/.style={->, black!30},
    qlabel/.style={font=\fontsize{6}{7}\selectfont\itshape, text=black!40}
]
\node[prompt] (p0) {$\pi_0$ \\ \tiny 13\%};
\node[prompt, below left=of p0, xshift=0.2cm] (p1a) {$\pi_1^a$ \\ \tiny 14\%};
\node[prompt, below=of p0] (p1b) {$\pi_1^b$ \\ \tiny 11\%};
\node[prompt, below right=of p0, xshift=-0.2cm] (p1c) {$\pi_1^c$ \\ \tiny 12\%};
\node[prompt, below left=of p1a, xshift=0.2cm] (p2a) {$\pi_2^a$ \\ \tiny 12\%};
\node[prompt, below=of p1a] (p2b) {$\pi_2^b$ \\ \tiny 15\%};
\node[prompt, below=of p1c] (p2c) {$\pi_2^c$ \\ \tiny 13\%};
\node[prompt, below right=of p1c, xshift=-0.2cm] (p2d) {$\pi_2^d$ \\ \tiny 10\%};
\node[prompt, below left=of p2b, xshift=0.2cm] (p3a) {$\pi_3^a$ \\ \tiny 14\%};
\node[prompt, below=of p2b] (p3b) {$\pi_3^b$ \\ \tiny 13\%};
\draw[normaledge] (p0) -- (p1a) node[midway, above left, qlabel] {?};
\draw[normaledge] (p0) -- (p1b) node[midway, right, qlabel] {?};
\draw[normaledge] (p0) -- (p1c) node[midway, above right, qlabel] {?};
\draw[normaledge] (p1a) -- (p2a) node[midway, above left, qlabel] {?};
\draw[normaledge] (p1a) -- (p2b) node[midway, right, qlabel] {?};
\draw[normaledge] (p1c) -- (p2c) node[midway, left, qlabel] {?};
\draw[normaledge] (p1c) -- (p2d) node[midway, above right, qlabel] {?};
\draw[normaledge] (p2b) -- (p3a) node[midway, above left, qlabel] {?};
\draw[normaledge] (p2b) -- (p3b) node[midway, right, qlabel] {?};
\end{tikzpicture}
}
\vspace{0pt}\hrule\vspace{4pt}
\textcolor{red!80!black}{\faTimes}\;\textbf{Trajectory opacity:} every edge carries only a question mark. The tree grows but carries no semantic meaning; the optimizer cannot detect oscillation, accumulate directional experience, or reason about its own trajectory.
\end{tcolorbox}
\vspace{-1em}
\caption{An illustrative case of trajectory opacity.}
\label{case:traj}
\vspace{-1em}
\end{figure}

\subsection{L3: Trajectory Opacity}

Even if the reflector's attribution happens to point in the right direction, the optimizer has no way to know it did. GEPA does perform selection across candidate prompts, but the optimization trajectory is entirely label-free (Figure~\ref{case:traj}). Each transition from one prompt to the next is driven by an accuracy-gain signal with no record of what root cause was attributed, what change was made, or why accuracy shifted. The optimizer knows that something worked, but not what, and has no basis for reasoning about what to try next.

This semantic vacuum has two concrete consequences. First, the optimizer cannot detect oscillation: if two conflicting attributions are applied alternately across rounds, there is no signal that the same ground is being revisited. Second, the optimizer cannot accumulate directional experience: each round starts from scratch, with no memory of which attribution categories have been productive or exhausted. The optimization tree grows in size but remains semantically empty---a map with nodes and edges but no labels, making the full evolution uninterpretable.

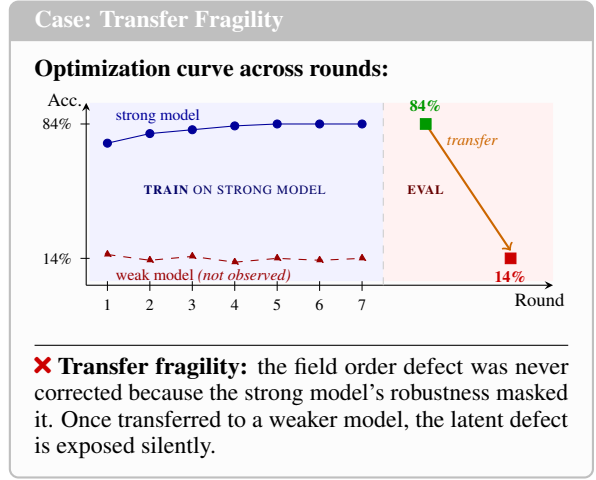
\begin{figure}
\begin{tcolorbox}[
    title={\small Case: Transfer Fragility},
    fonttitle=\bfseries\small,
    colback=gray!2,
    colframe=gray!50,
    boxrule=0.8pt,
    left=6pt, right=6pt, top=4pt, bottom=4pt
]
\small
\textbf{Optimization curve across rounds:}\\[1pt]
\begin{tikzpicture}
\begin{axis}[
    width=1.1\linewidth, height=4.0cm,
    xmin=0.5, xmax=11.5,
    ymin=0, ymax=95,
    xtick={1,2,3,4,5,6,7},
    xticklabels={1,2,3,4,5,6,7},
    ytick={14,84},
    yticklabels={14\%,84\%},
    tick label style={font=\tiny, color=black},
    xlabel={\scriptsize Round},
    ylabel={\scriptsize Acc.},
    xlabel style={
      at={(ticklabel* cs:1.0)},
      yshift=15pt,
      xshift=-5pt,
    },
    every axis y label/.style={
      at={(ticklabel* cs:1.0)},
      yshift=1pt,
      xshift=-8pt,
      rotate=0
    },
    axis line style={color=black},
    tick style={color=black},
    axis lines=left,
    grid=none,
    clip=false,
]

\addplot[fill=blue!5, draw=none, forget plot]
    coordinates {(0.58,2.1)(7.5,2.1)(7.5,95)(0.58,95)} \closedcycle;

\addplot[fill=red!5, draw=none, forget plot]
    coordinates {(7.5,2.1)(11.5,2.1)(11.5,95)(7.5,95)} \closedcycle;

\node[font=\tiny, text=blue!40!black] at (axis cs:4,50) {\textsc{\textbf{train} on strong model}};
\node[font=\tiny, text=red!40!black] at (axis cs:8.5,50) {\textsc{\textbf{eval}}};
\draw[dashed, black!20] (axis cs:7.5,0) -- (axis cs:7.5,95);

\addplot[
    color=blue!60!black,
    mark=*,
    mark size=1.5pt,
    mark options={fill=blue!60!black}
] coordinates {
    (1,74) (2,79) (3,81) (4,83) (5,84) (6,84) (7,84)
};
\node[font=\tiny, text=blue!60!black, anchor=south west] at (axis cs:1,80) {strong model};

\addplot[
    color=red!60!black,
    dashed,
    mark=triangle*,
    mark size=1.5pt,
    mark options={fill=red!60!black}
] coordinates {
    (1,16) (2,13) (3,15) (4,12) (5,14) (6,13) (7,14)
};
\node[font=\tiny, text=red!60!black, anchor=north west] at (axis cs:1,13) {weak model \textit{(not observed)}};

\addplot[
    color=green!60!black,
    mark=square*,
    mark size=2pt,
    mark options={fill=green!60!black},
    only marks
] coordinates {(8.5,84)};
\node[font=\tiny, text=green!60!black, anchor=south] at (axis cs:8.5,86) {\textbf{84\%}};

\addplot[
    color=red!80!black,
    mark=square*,
    mark size=2pt,
    mark options={fill=red!80!black},
    only marks
] coordinates {(10.5,14)};
\node[font=\tiny, text=red!80!black, anchor=north] at (axis cs:10.5,12) {\textbf{14\%}};

\draw[->, thick, orange!80!black] (axis cs:8.5,84) -- (axis cs:10.5,18);
\node[font=\tiny, text=orange!80!black, anchor=west] at (axis cs:8.8,75) {\textit{transfer}};

\end{axis}
\end{tikzpicture}\\[0pt]


\hrule\vspace{4pt}
\textcolor{red!80!black}{\faTimes}\;\textbf{Transfer fragility:} the field order defect was never corrected because the strong model's robustness masked it. Once transferred to a weaker model, the latent defect is exposed silently.
\end{tcolorbox}
\vspace{-1em}
\caption{An illustrative case of transfer fragility.}
\label{case:trans}
\vspace{-1em}
\end{figure}

\begin{figure*}[t]
    \centering
    \includegraphics[width=\linewidth]{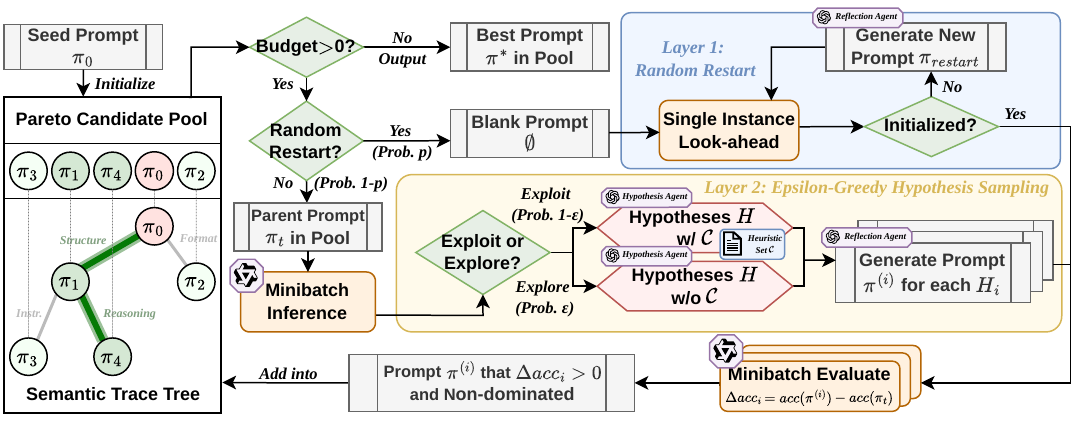}
    \caption{VISTA pipeline overview: a multi-agent framework that decouples hypothesis generation from prompt rewriting via heuristic-guided parallel verification and interpretable semantic trace trees.}
    \label{fig:pipeline}
\vspace{-1em}
\end{figure*}

\subsection{L4: Transfer Fragility}

Prompts optimized under reflective APO are implicitly tailored to the base model's behavior during optimization~\cite{zhao2021calibrate, lu2022prompt}. When the optimized prompt is transferred to a different base model, the structural assumptions encoded in the prompt---output format expectations, reasoning style preferences, instruction sensitivity---may no longer hold, causing silent performance degradation. Because reflective APO produces no record of which model-specific behaviors were exploited during optimization, it provides no signal about where its results will or will not generalize.

This fragility is particularly counterintuitive: a prompt optimized against a stronger model may mask latent defects that the model's robustness compensates for. Once transferred to a weaker model that more faithfully follows the defective schema, the latent defect is exposed---and there is no mechanism to detect this regression without re-running evaluation on the target model (Figure~\ref{case:trans}).

\section{Proposed VISTA Framework}
\label{sec:method}

\subsection{Overview}

VISTA decouples hypothesis generation from prompt rewriting via a multi-agent design~\cite{wu2023autogen}: a \textit{hypothesis agent} proposes semantically labeled hypotheses, and a \textit{reflection agent} rewrites the prompt targeting each hypothesis independently. Each prompt update is thus grounded in an explicit, verifiable hypothesis, making the optimization process interpretable at every step.

The Pareto pool retains non-dominated prompts under per-sample dominance following \citet{agrawal2025gepa}, with prompts sampled proportional to their per-sample win count. In each round, a selected prompt is passed through the two-layer explore-exploit mechanism (Section~\ref{sec:exploit}) to generate $K$ candidate hypotheses (Section~\ref{sec:hypothesis}). The reflection agent produces $K$ candidate prompts, one per hypothesis. The best-verified candidate may be added to the Pareto pool with its root-cause label, extending the semantic trace tree (Section~\ref{sec:trace}). Figure~\ref{fig:pipeline} gives a visual overview, and Algorithm~\ref{alg:vista} formalizes the full procedure.

\begin{algorithm}[t]\small
\caption{VISTA Prompt Optimization}
\label{alg:vista}
\begin{algorithmic}
\REQUIRE dataset $\mathcal{D} = \mathcal{D}_{\text{train}} \cup \mathcal{D}_{\text{val}}$, seed prompt $\pi_0$, heuristic set $\mathcal{C}$, budget $B$, hyperparameters $K, p, \varepsilon, b$
\ENSURE Optimized prompt $\pi^*$
\STATE Initialize Pareto pool $\mathcal{P} \leftarrow \{\pi_0\}$, $\pi^* \leftarrow \pi_0$
\WHILE{budget $B > 0$}
    \STATE $\pi_t \leftarrow \textsc{SelectPareto}(\mathcal{P})$, $\pi' \leftarrow \varnothing$
    \STATE Sample $\mathcal{M} \sim \mathcal{D}_{\text{train}}$ with $|\mathcal{M}| = b$
    \IF{$\text{Uniform}(0,1) < p$}
        \STATE $\pi_{\text{restart}} \leftarrow \textsc{Restart}(\mathcal{D}_{\text{train}})$
        \IF{$\Delta\text{acc}(\pi_{\text{restart}}, \mathcal{M}) > 0$}
            \STATE $\pi' \leftarrow \pi_{\text{restart}}$
        \ENDIF
        \STATE $B' \leftarrow b + 1$
    \ELSE
        \FOR{$i = 1, \ldots, K$}
            \STATE Collect failure cases $\mathcal{F}_t$ from $\mathcal{M}$
            \STATE $H_i \leftarrow \textsc{Sample}(\mathcal{C},\ \mathcal{F}_t,\ \varepsilon)$
            \STATE $\pi^{(i)} \leftarrow \textsc{Reflect}(\pi_t,\ H_i, \mathcal{F}_t)$
        \ENDFOR
        \STATE $\pi' \leftarrow \pi^{(i^*)}$, $i^* = \arg\max_{i \in \{j \,:\, \Delta\text{acc}_j > 0\}} \Delta\text{acc}_i$
        \STATE $B' \leftarrow b \cdot K$
    \ENDIF
    \IF{$\pi' \neq \varnothing$}
        \STATE Evaluate $\pi'$ on $\mathcal{D}_{\text{val}}$
        \IF{$\pi'$ is not dominated by any $\pi \in \mathcal{P}$ on $\mathcal{D}_{\text{val}}$}
            \STATE Add $\pi'$ to $\mathcal{P}$ with label $c^* = c_{i^*}$
            \STATE \textsc{UpdatePareto}($\mathcal{P})$
        \ENDIF
        \IF{$\text{acc}(\pi',\ \mathcal{D}_{\text{val}}) > \text{acc}(\pi^*,\ \mathcal{D}_{\text{val}})$}
            \STATE $\pi^* \leftarrow \pi'$
        \ENDIF
        \STATE $B' \leftarrow B' + |\mathcal{D}_{\text{val}}|$
    \ENDIF
    \STATE $B \leftarrow B - B'$
\ENDWHILE
\RETURN $\pi^*$
\end{algorithmic}
\end{algorithm}

\subsection{Hypothesis Generation}
\label{sec:hypothesis}

Each hypothesis $H_i = (c_i, d_i)$ consists of a category label $c_i \in \mathcal{C}$ and a natural language description $d_i$ of the suspected failure mode. Let $\mathcal{F}_t$ denote the set of failure cases at round $t$. The label set $\mathcal{C}$ is an extensible taxonomy of heuristic failure modes curated from representative cases; each entry specifies a failure mode category, a description, and a suggested fix direction (see Appendix~\ref{sec:appendix_prompts}). 

Let $\mathcal{H}_\theta$ denote the hypothesis space of a reflector with parameters $\theta$, and let $c^*$ denote the true root cause. By \citet{huang2023selfcorrect}, $P(c^* \in \mathcal{H}_\theta) < 1$ for root causes outside the model's prior, and this probability cannot be improved by further prompting alone. An unconstrained hypothesis agent is subject to the same bound. Let $\beta = P(c^* \in \mathcal{C})$ denote the coverage probability of $\mathcal{C}$. The heuristic set introduces an external prior independent of $\theta$, ensuring:
\begin{equation}
    \beta > P(c^* \in \mathcal{H}_\theta)
\end{equation}
for any $c^*$ covered by the heuristic set.

Let $\pi_t$ denote the prompt selected from $\mathcal{P}$ at round $t$. For each hypothesis $H_i$, the reflection agent independently rewrites $\pi_t$ to address the hypothesized root cause, producing candidate $\pi^{(i)}$; we write $\{\pi^{(i)}\}_{i=1}^K$ for the full set of $K$ candidates. Let $\mathcal{D} = \mathcal{D}_{\text{train}} \cup \mathcal{D}_{\text{val}}$ denote the task dataset partitioned into a training split for failure cases collection and a validation split for candidate selection. Let $\mathcal{M} \sim \mathcal{D}_{\text{train}}$ denote a minibatch of size $b$ and $\text{acc}(\pi, \mathcal{M})$ the accuracy of prompt $\pi$ on $\mathcal{M}$. Define the accuracy gain of a prompt $\pi$ on minibatch $\mathcal{M}$ relative to the current prompt $\pi_t$ as:
\begin{equation}
    \Delta\text{acc}(\pi, \mathcal{M}) = \text{acc}(\pi, \mathcal{M}) - \text{acc}(\pi_t, \mathcal{M})
\end{equation}
The accuracy gain of the $i$-th candidate is then $\Delta\text{acc}_i = \Delta\text{acc}(\pi^{(i)}, \mathcal{M})$, and the winning hypothesis is selected by:
\begin{equation}
    i^* = \arg\max_{i \in \{j \,:\, \Delta\text{acc}_j > 0\}} \Delta\text{acc}_i
\end{equation}
The winner $\pi^{(i^*)}$ is then evaluated on $\mathcal{D}_{\text{val}}$ and added to $\mathcal{P}$ with label $c^* = c_{i^*}$ if not dominated by any $\pi \in \mathcal{P}$: the selected root cause is not what the reflector believes caused the failure, but what empirically produces the largest improvement on held-out instances.

\subsection{Semantic Trace}
\label{sec:trace}

VISTA maintains a \textit{semantic trace tree} $\mathcal{T} = (V, E)$, 
where each node $v \in V$ corresponds to a prompt candidate $\pi^{(v)}$ 
and each directed edge $(u, v) \in E$ is annotated with the tuple 
$(c^*, \delta)$, where $\delta = \Delta\text{acc}$ denotes the accuracy 
gain of the optimization step that produced $v$ from $u$. The tree is 
rooted at $\pi_0$ and grows by one node per successful round, giving 
the full optimization history a structured, auditable form: every prompt 
can be traced back to its root-cause chain.

The optimization trajectory up to round $t$ is the root-to-current path:
\begin{equation}\fontsize{10}{9}
    \tau_t = \left(\pi_0 \xrightarrow{(c_1^*,\, \delta_1)} 
    \pi_1 \xrightarrow{(c_2^*,\, \delta_2)} \cdots 
    \xrightarrow{(c_t^*,\, \delta_t)} \pi_t\right)
\end{equation}
where $c_k^* \in \mathcal{C}$ is the selected root-cause label at round $k$. Unlike GEPA's unlabeled sequence, $\tau_t$ is fully interpretable: each transition has a causal explanation and a verified performance delta $\Delta\text{acc}^{(t)}$.

The trace $\tau_t$ is provided to the hypothesis agent as context at each round, enabling it to avoid already-explored directions, identify diminishing-return categories, and detect when two labels alternate without joint resolution. In the latter case, the agent generates a joint hypothesis addressing both dimensions simultaneously, collapsing what would otherwise require multiple sequential rounds into a single update.

\begin{table*}[t]\fontsize{10}{11}\selectfont
\centering
\caption{Accuracy (\%) across benchmarks, conditions, and methods.}
\label{tab:main}
\begin{tabular}{l|ccc|ccc}
\toprule
& \multicolumn{3}{c}{\textbf{GSM8K}} & \multicolumn{3}{c}{\textbf{AIME2025}} \\
\cmidrule(lr){2-4} \cmidrule(lr){5-7}
\textbf{Method} & \textbf{Defective} & \textbf{Repaired} & \textbf{Minimal} & \textbf{Defective} & \textbf{Repaired} & \textbf{Minimal} \\
\midrule
No Opt. & 23.81          & 85.59          & 20.67          & 38.67          & 40.00          & 40.00 \\
GEPA    & 13.50          & 86.53          & 21.68          & 44.00          & 39.33          & 42.00 \\
\rowcolor{blue!6}
VISTA   & \textbf{87.57} & \textbf{87.34} & \textbf{85.67} & \textbf{46.00} & \textbf{46.67} & \textbf{44.00} \\
\bottomrule
\end{tabular}
\vspace{-0.5em}
\end{table*}

\subsection{Two-Layer Explore-Exploit}
\label{sec:exploit}

VISTA introduces a two-layer explore-exploit mechanism~\cite{sutton2018reinforcement} targeting L1 and L2 respectively.

\paragraph{Layer 1: Random Restart.}
At the start of each round, with probability $p \in (0, 1)$, VISTA triggers a random restart~\cite{lourenco2003iterated}. A \textit{blank prompt look-ahead} is executed iteratively: a single training instance is run under the current prompt, collecting the raw output $o_{\text{raw}}$ and any parsing error $e$. The reflection agent constructs a new prompt conditioned on $(o_{\text{raw}}, e)$, which is then used in the next look-ahead step. This loop repeats until $e = \varnothing$, at which point $\pi_{\text{restart}}$ is considered initialized from the model's natural behavior rather than from any inherited seed constraints. Formally, while standard mutation conditions on $\pi_t$:
\begin{equation}
    \pi^{(i)} = f_{\text{reflect}}(\pi_t, H_i, \mathcal{F}_t)
\end{equation}
random restart conditions on the model's natural behavior, using a fixed null 
hypothesis $H_{\text{blank}} = (\texttt{none},\ \text{``initialize from model 
output''})$ that carries no prior attribution:
\begin{equation}
    \pi_{\text{restart}} = f_{\text{reflect}}(\emptyset,\ H_{\text{blank}},\ o_{\text{raw}})
\end{equation}
This ensures the restarted prompt reflects what the model naturally produces, not what the seed constrains it to produce. The resulting $\pi_{\text{restart}}$ is then evaluated against $\pi_t$ on $\mathcal{M}$; it is added to $\mathcal{P}$ only if $\Delta\text{acc}(\pi_{\text{restart}}, \mathcal{M}) > 0$ and not dominated by any $\pi \in \mathcal{P}$ on $\mathcal{D}_{val}$. The cost of a restart is a few look-ahead steps, making it negligible relative to the minibatch validation budget.

\paragraph{Layer 2: Epsilon-Greedy Hypothesis Sampling.}
Within each round, the $K$ hypotheses are drawn according to an epsilon-greedy strategy. For each of the $K$ slots:
\begin{equation}\small
    H_i \sim
    \begin{cases}
        \text{Heuristic}(\mathcal{C},\ \mathcal{F}_t) & \text{with probability } 1 - \varepsilon \\
        \text{Free}(\mathcal{F}_t) & \text{with probability } \varepsilon
    \end{cases}
\end{equation}
The heuristic set branch exploits known failure mode categories; the free branch generates unconstrained hypotheses, allowing discovery of failure modes outside $\mathcal{C}$. In expectation, $\lfloor(1-\varepsilon)K\rfloor$ hypotheses target known modes and $\lceil\varepsilon K\rceil$ explore novel ones.

\paragraph{Sample Efficiency.}
For the purposes of this analysis, we treat $\beta$ as VISTA's effective 
per-round probability of identifying $c^*$, lower-bounded by 
$P(c^* \in \mathcal{C})$ as defined in Section~\ref{sec:hypothesis}. Let 
$\alpha \in [0,1)$ denote GEPA's empirical probability of selecting $c^*$ in 
any given round. The analysis in Section~\ref{attr-blindspot} suggests that 
$\alpha$ is empirically close to zero for structural failure modes, while 
$\beta > \alpha$ follows from $\mathcal{C}$ explicitly covering structural 
categories.

Assuming that each round independently samples a candidate root cause with 
fixed probability (approximated as geometric trials), the expected number of 
rounds until $c^*$ is first identified satisfies

\begin{equation}
\mathbb{E}[N_{\text{VISTA}}] \le \frac{1}{\beta}
< \frac{1}{\alpha}
= \mathbb{E}[N_{\text{GEPA}}], 
\quad \alpha > 0
\end{equation}

where $N$ denotes the number of rounds until $c^*$ is first selected as $i^*$. 
The inequality follows from $\beta > \alpha > 0$. In the limiting case where 
$\alpha \to 0$, $\mathbb{E}[N_{\text{GEPA}}]$ diverges, which is consistent 
with the empirical observation that GEPA fails to identify the structural 
root cause across all rounds.

\section{Experiments}
\label{sec:experiments}

\subsection{Experimental Setup}

\paragraph{Models and Benchmarks.}
For GSM8K~\cite{cobbe2021gsm8k} experiments, we use Qwen3-4B as the base model and Qwen3-8B~\cite{qwen3technical} as the reflector. For AIME2025~\cite{aime} experiments, we use GPT-4.1-mini~\cite{gpt4.1technical} as the base model and GPT-4o-mini~\cite{openai2024gpt4omini} as the reflector.

\paragraph{Conditions.}
We evaluate under three seed conditions to stress-test optimizer robustness: (1) Defective seed: the official GEPA\cite{agrawal2025gepa} seed prompt, which contains an inverted output field order that silently disables CoT reasoning; (2) Repaired seed: a manually corrected version with the field order fixed; (3) Minimal seed: a single-sentence prompt with no structural constraints.

\paragraph{Baselines.}
We compare against two baselines: No optimization, which evaluates the seed prompt directly, and GEPA, the state-of-the-art reflective APO method.

\paragraph{VISTA Configuration.}
Unless otherwise specified, VISTA uses $K=3$ hypotheses per round, restart probability $p=0.2$, and exploration rate $\varepsilon=0.1$. (see Appendix~\ref{sec:appendix_config} for full setup details).

\subsection{Main Results}

Table~\ref{tab:main} reports results across all seed conditions, methods, and benchmarks.

\begin{figure*}[t]
    \centering
    \begin{subfigure}[b]{0.49\linewidth}
        \includegraphics[width=\linewidth]{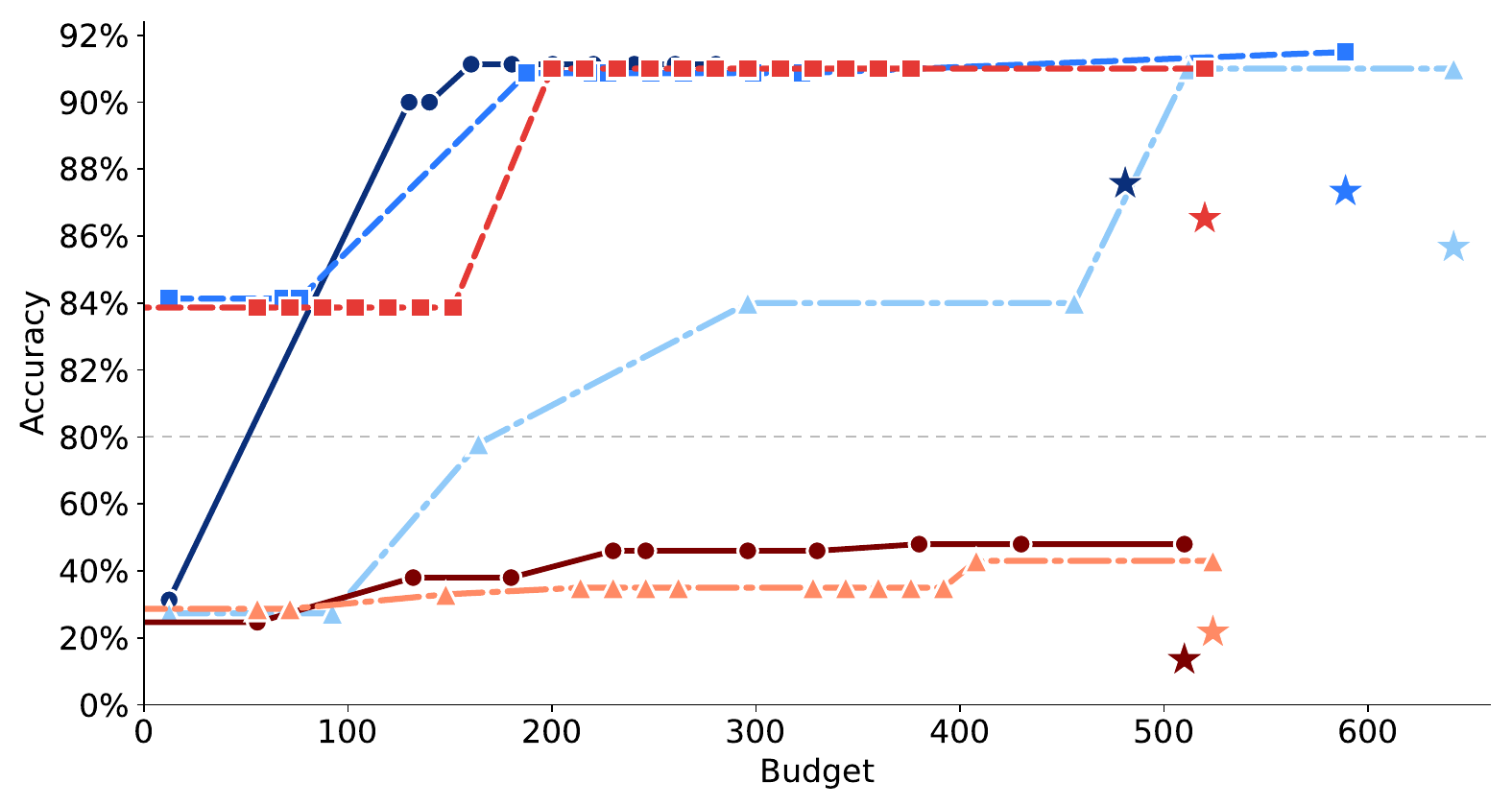}
        \caption{Accuracy (\%) vs.\ metric calls.}
        \label{fig:curve_budget}
    \end{subfigure}
    \hfill
    \begin{subfigure}[b]{0.49\linewidth}
        \includegraphics[width=\linewidth]{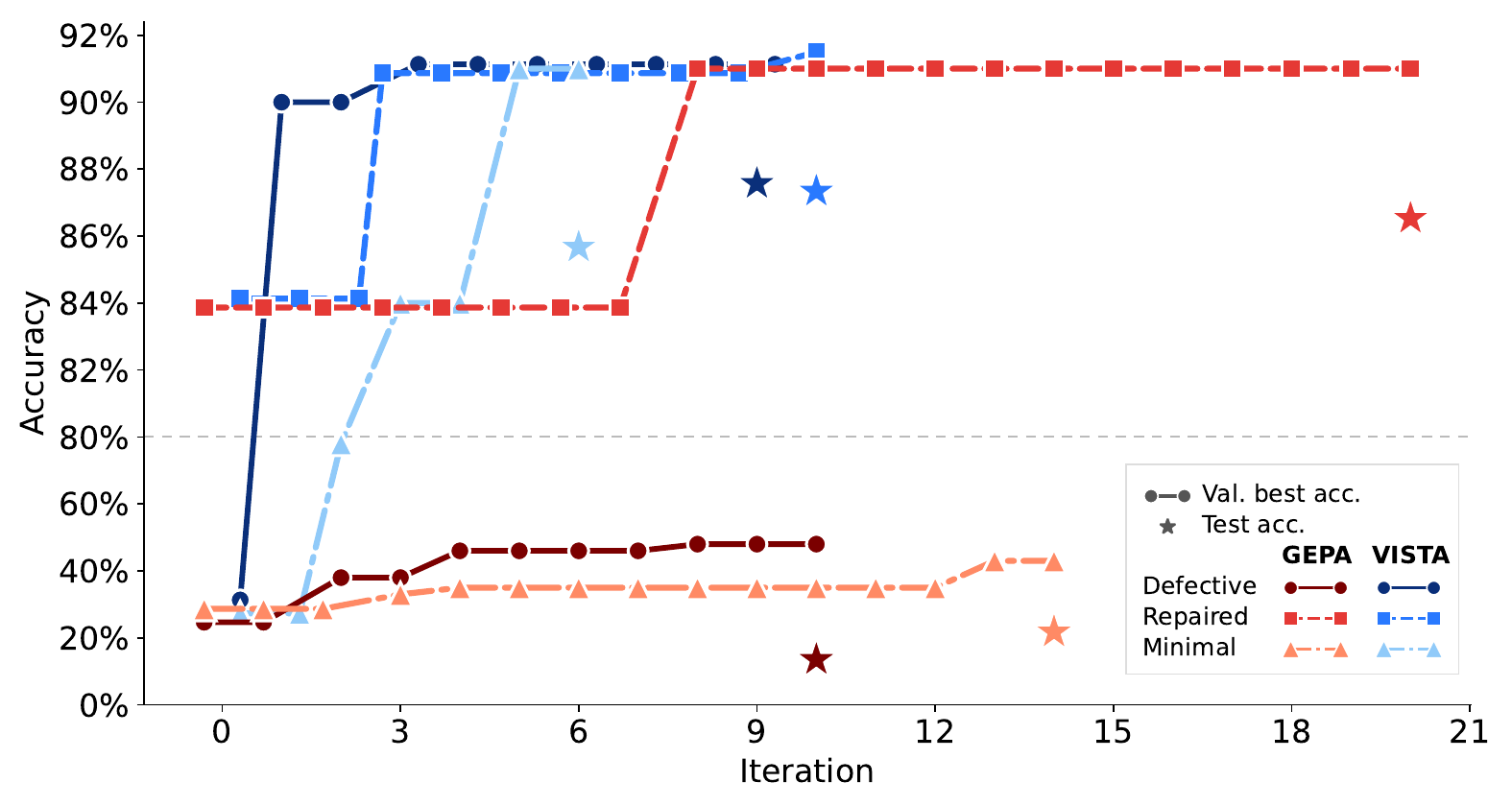}
        \caption{Accuracy (\%) vs.\ optimization rounds.}
        \label{fig:curve_rounds}
    \end{subfigure}
    \caption{Optimization curves on GSM8K under the defective seed (best score to date).}
    \label{fig:curve}
\vspace{-1em}
\end{figure*}

\paragraph{GSM8K.}
Under the defective seed, GEPA degrades accuracy from 23.81\% to 13.50\% while VISTA recovers to 87.57\% (+74.07 points), directly validating L1 and providing evidence for L2. The root cause---inverted output field order---is never hypothesized by GEPA across all optimization rounds even with full failure information, confirming that the hypothesis space is the binding constraint rather than information availability. As shown in Figure~\ref{fig:curve_rounds}, GEPA's best-score trajectory remains flat throughout optimization while VISTA converges within the first few rounds, and this advantage holds under the same budget (Figure~\ref{fig:curve_budget}).

Under the repaired seed, all methods converge to similar accuracy, confirming VISTA introduces no regression on well-formed seeds. Under the minimal seed, GEPA again fails to escape the low-accuracy region (20.67\% $\to$ 21.68\%), while VISTA recovers to 85.67\%, demonstrating that VISTA's gains are not specific to the 
field-ordering defect.

Table~\ref{tab:reflector} validates L2 and L4. Across both reflector configurations, GEPA remains near baseline while VISTA maintains strong performance, confirming the blindspot is prior-structural rather than capability-driven. For L4, GEPA's cross-model accuracy (22.74\%) remains near its 
single-model baseline (13.50\%), confirming that optimization gains 
under GEPA do not transfer across base models. By contrast, VISTA 
maintains strong performance under transfer (86.05\%), suggesting 
that heuristic-guided optimization produces more structurally 
generalizable prompts.

\paragraph{AIME2025.}
Absolute gains are smaller than on GSM8K, reflecting the task's lower sensitivity to prompt structure under high problem difficulty. Nevertheless, VISTA consistently outperforms GEPA across all seed conditions, and GEPA degrades below the no-optimization baseline under the repaired seed (39.33\% vs.\ 40.00\%), further confirming the instability of unconstrained reflection.

\begin{table}[t]\fontsize{10}{11}\selectfont
{\setlength{\tabcolsep}{4pt}
\centering
\caption{GSM8K accuracy (\%) on the defective seed. Columns 1--2: Qwen3-4B base, Qwen3-8B/GPT-4o-mini reflectors. Column 3: trained on GPT-4.1-mini (GPT-4o-mini reflector), evaluated on Qwen3-4B.}
\label{tab:reflector}
\begin{tabular}{l|cc|c}
\toprule
& \multicolumn{2}{c|}{\textbf{Reflector}} & \\
\cmidrule(lr){2-3}
\textbf{Method} & \textbf{Qwen3-8B} & \textbf{GPT-4o-mini} & \textbf{Cross-Model} \\
\midrule
GEPA  & 13.50          & 23.43          & 22.74 \\
\rowcolor{blue!6}
VISTA & \textbf{87.57} & \textbf{87.64} & \textbf{86.05} \\
\bottomrule
\end{tabular}
}
\vspace{-0.5em}
\end{table}

\subsection{Ablation Study}

We conduct ablation experiments on GSM8K under the defective seed. All ablations fix Qwen3-8B as the reflector. Table~\ref{tab:ablation} reports all results.

Two findings stand out. First, heuristic guidance is the dominant factor: removing exploitation ($\varepsilon=1.0$) collapses accuracy to 22.97\%, while removing exploration ($\varepsilon=0$) causes only a modest drop to 85.60\%. Second, the component contribution rows confirm that the heuristic set accounts for the largest single gain ($+$59.81 points), with random restart and parallel sampling contributing cumulatively but secondarily. $K=3$ provides the best tradeoff; performance degrades at $K=5$, suggesting noise from additional candidates outweighs their diversity benefit.

\begin{table}[t]\fontsize{10}{11}\selectfont
{\setlength{\tabcolsep}{5.5pt}
\centering
\caption{Ablation results on GSM8K (\%) under the defective seed (Qwen3-4B base, Qwen3-8B reflector).}
\label{tab:ablation}
\begin{tabular}{l|ccc|c}
\toprule
& \multicolumn{3}{c|}{\textbf{Hyper.}} & \\
\cmidrule(lr){2-4}
\textbf{Configuration} & $K$ & $p$ & $\varepsilon$ & \textbf{Acc.} \\
\midrule
\multicolumn{5}{l}{\cellcolor{gray!15}\textbf{Effect of $K$}} \\
VISTA & 3  & 0.2 & 0.1 & 87.57 \\
VISTA, K=1 & 1 & 0 & 0.1 & 75.97 \\
VISTA, K=3 & 3 & 0 & 0.1 & 86.81 \\
VISTA, K=5 & 5 & 0 & 0.1 & 83.89 \\
\midrule
\multicolumn{5}{l}{\cellcolor{gray!15}\textbf{Effect of Heuristic Set}} \\
VISTA & 3  & 0.2 & 0.1 & 87.57 \\
w/o Exploration  & 3 & 0 & 0 & 85.60 \\
w/o Exploitation & 3 & 0 & 1.0 & 22.97 \\
\midrule
\multicolumn{5}{l}{\cellcolor{gray!15}\textbf{Component Contribution}} \\
GEPA                & -- & --   & --  & 13.50 \\
$+$ Restart           & 0  & 0.2 & 0 & 15.69 \\
$+$ Parallel Sampling & 3  & 0.2   & 0 & 20.17 \\
$+$ Heur.-Guided Reflection     & 3  & 0.2   & 0 & 79.98 \\
\bottomrule
\end{tabular}
}
\vspace{-1em}
\end{table}

\section{Conclusion}
\label{sec:conclusion}

We identified four systematic limitations of reflective APO methods under a unified interpretability-blindspot framework, and demonstrated them concretely: on GSM8K with a defective seed, GEPA degrades accuracy from 23.81\% to 13.50\% while the root cause is never identified across all optimization rounds. We proposed VISTA, a multi-agent APO framework that decouples hypothesis generation from prompt rewriting, replacing black-box reflection with heuristic-guided parallel verification and interpretable semantic trace trees. VISTA recovers accuracy to 87.57\% on the same defective seed and maintains strong performance across all configurations. We hope this work motivates a broader shift toward interpretable, robust optimization in the APO paradigm.





\section*{Limitations}
VISTA directly addresses L1, L2, and L3, and partially mitigates L4: heuristic-guided optimization produces structurally grounded prompts that transfer more reliably across base models than unconstrained reflection, as evidenced by Table~\ref{tab:reflector}. Transfer fragility nonetheless remains open in general---VISTA provides no explicit signal about generalization, and its transfer advantage may not hold across model families with larger capability gaps. Incorporating multi-model minibatch validation, where hypothesis selection is based on accuracy gains averaged across a distribution of base models, is a natural extension. Beyond L4, several further limitations apply.

\paragraph{Heuristic coverage.} The heuristic set has finite coverage; failure modes outside its scope can only be discovered via the $\varepsilon$-exploration branch, which has lower expected sample efficiency than heuristic-guided search. This motivates more principled exploration strategies.

\paragraph{Hyperparameter sensitivity.} VISTA introduces three hyperparameters ($K$, $p$, $\varepsilon$) that interact with the rollout budget in non-obvious ways. Our ablations identify $K{=}3$, $p{=}0.2$, $\varepsilon{=}0.1$ as a robust default, but optimal configurations likely vary across tasks and budget constraints.

\paragraph{Reasoning models.} Our analysis focuses on standard instruction-following models. Modern reasoning models generate explicit chain-of-thought traces before producing output, which may partially circumvent the field-ordering defect in L1 regardless of output schema order. The seed trap and attribution blindspot may therefore be less severe for such models, and the structural gains from VISTA's heuristics could diminish accordingly.

\paragraph{Benchmark coverage.} Experiments are confined to mathematical reasoning benchmarks (GSM8K and AIME2025). Whether the four limitations and VISTA's mitigations generalize to more open-ended domains---such as long-form generation, dialogue, or coding---remains an open question, as these domains may exhibit different failure mode distributions that the current heuristic set does not cover.

\section*{Acknowledgments}
We would like to thank the anonymous reviewers for their insightful comments and constructive suggestions, which helped improve the quality of this paper. The first author is also grateful to the University of California, Berkeley, for providing the valuable opportunity for an academic visit. Furthermore, we extend our sincere gratitude to the professors there for the precious opportunities for academic exchange and insightful discussions.


\bibliography{custom}

@inproceedings{agrawal2025gepa,
    title={{GEPA}: Reflective Prompt Evolution Can Outperform Reinforcement Learning},
    author={Lakshya A Agrawal and Shangyin Tan and Dilara Soylu and Noah Ziems and Rishi Khare and Krista Opsahl-Ong and Arnav Singhvi and Herumb Shandilya and Michael J Ryan and Meng Jiang and Christopher Potts and Koushik Sen and Alex Dimakis and Ion Stoica and Dan Klein and Matei Zaharia and Omar Khattab},
    booktitle={The Fourteenth International Conference on Learning Representations},
    year={2026},
    url={https://openreview.net/forum?id=RQm2KQTM5r}
}

@inproceedings{brown2020fewshot,
    author = {Brown, Tom and Mann, Benjamin and Ryder, Nick and Subbiah, Melanie and Kaplan, Jared D and Dhariwal, Prafulla and Neelakantan, Arvind and Shyam, Pranav and Sastry, Girish and Askell, Amanda and Agarwal, Sandhini and Herbert-Voss, Ariel and Krueger, Gretchen and Henighan, Tom and Child, Rewon and Ramesh, Aditya and Ziegler, Daniel and Wu, Jeffrey and Winter, Clemens and Hesse, Chris and Chen, Mark and Sigler, Eric and Litwin, Mateusz and Gray, Scott and Chess, Benjamin and Clark, Jack and Berner, Christopher and McCandlish, Sam and Radford, Alec and Sutskever, Ilya and Amodei, Dario},
    booktitle = {Advances in Neural Information Processing Systems},
    editor = {H. Larochelle and M. Ranzato and R. Hadsell and M.F. Balcan and H. Lin},
    pages = {1877--1901},
    publisher = {Curran Associates, Inc.},
    title = {Language Models are Few-Shot Learners},
    url = {https://proceedings.neurips.cc/paper_files/paper/2020/file/1457c0d6bfcb4967418bfb8ac142f64a-Paper.pdf},
    volume = {33},
    year = {2020}
}

@inproceedings{wei2022cot,
    author = {Wei, Jason and Wang, Xuezhi and Schuurmans, Dale and Bosma, Maarten and ichter, brian and Xia, Fei and Chi, Ed and Le, Quoc V and Zhou, Denny},
    booktitle = {Advances in Neural Information Processing Systems},
    editor = {S. Koyejo and S. Mohamed and A. Agarwal and D. Belgrave and K. Cho and A. Oh},
    pages = {24824--24837},
    publisher = {Curran Associates, Inc.},
    title = {Chain-of-Thought Prompting Elicits Reasoning in Large Language Models},
    url = {https://proceedings.neurips.cc/paper_files/paper/2022/file/9d5609613524ecf4f15af0f7b31abca4-Paper-Conference.pdf},
    volume = {35},
    year = {2022}
}

@misc{cobbe2021gsm8k,
    title={Training Verifiers to Solve Math Word Problems}, 
    author={Karl Cobbe and Vineet Kosaraju and Mohammad Bavarian and Mark Chen and Heewoo Jun and Lukasz Kaiser and Matthias Plappert and Jerry Tworek and Jacob Hilton and Reiichiro Nakano and Christopher Hesse and John Schulman},
    year={2021},
    eprint={2110.14168},
    archivePrefix={arXiv},
    primaryClass={cs.LG},
    url={https://arxiv.org/abs/2110.14168}, 
}

@article{liu2026evox,
  title={Evox: Meta-evolution for automated discovery},
  author={Liu, Shu and Agarwal, Shubham and Maheswaran, Monishwaran and Cemri, Mert and Li, Zhifei and Mang, Qiuyang and Naren, Ashwin and Boneh, Ethan and Cheng, Audrey and Pan, Melissa Z and others},
  journal={arXiv preprint arXiv:2602.23413},
  year={2026}
}

@inproceedings{honovich2022instruction,
    title = "Instruction Induction: From Few Examples to Natural Language Task Descriptions",
    author = "Honovich, Or  and
      Shaham, Uri  and
      Bowman, Samuel R.  and
      Levy, Omer",
    editor = "Rogers, Anna  and
      Boyd-Graber, Jordan  and
      Okazaki, Naoaki",
    booktitle = "Proceedings of the 61st Annual Meeting of the Association for Computational Linguistics (Volume 1: Long Papers)",
    month = jul,
    year = "2023",
    address = "Toronto, Canada",
    publisher = "Association for Computational Linguistics",
    url = "https://aclanthology.org/2023.acl-long.108/",
    doi = "10.18653/v1/2023.acl-long.108",
    pages = "1935--1952"
}

@inproceedings{zhou2023largelanguage,
    title={Large Language Models are Human-Level Prompt Engineers},
    author={Yongchao Zhou and Andrei Ioan Muresanu and Ziwen Han and Keiran Paster and Silviu Pitis and Harris Chan and Jimmy Ba},
    booktitle={The Eleventh International Conference on Learning Representations },
    year={2023},
    url={https://openreview.net/forum?id=92gvk82DE-}
}

@inproceedings{khattab2024dspy,
    title={{DSP}y: Compiling Declarative Language Model Calls into State-of-the-Art Pipelines},
    author={Omar Khattab and Arnav Singhvi and Paridhi Maheshwari and Zhiyuan Zhang and Keshav Santhanam and Sri Vardhamanan A and Saiful Haq and Ashutosh Sharma and Thomas T. Joshi and Hanna Moazam and Heather Miller and Matei Zaharia and Christopher Potts},
    booktitle={The Twelfth International Conference on Learning Representations},
    year={2024},
    url={https://openreview.net/forum?id=sY5N0zY5Od}
}

@inproceedings{madaan2023selfrefine,
 author = {Madaan, Aman and Tandon, Niket and Gupta, Prakhar and Hallinan, Skyler and Gao, Luyu and Wiegreffe, Sarah and Alon, Uri and Dziri, Nouha and Prabhumoye, Shrimai and Yang, Yiming and Gupta, Shashank and Majumder, Bodhisattwa Prasad and Hermann, Katherine and Welleck, Sean and Yazdanbakhsh, Amir and Clark, Peter},
 booktitle = {Advances in Neural Information Processing Systems},
 editor = {A. Oh and T. Naumann and A. Globerson and K. Saenko and M. Hardt and S. Levine},
 pages = {46534--46594},
 publisher = {Curran Associates, Inc.},
 title = {Self-Refine: Iterative Refinement with Self-Feedback},
 url = {https://proceedings.neurips.cc/paper_files/paper/2023/file/91edff07232fb1b55a505a9e9f6c0ff3-Paper-Conference.pdf},
 volume = {36},
 year = {2023}
}

@inproceedings{olausson2023selfrepair,
    title={Is Self-Repair a Silver Bullet for Code Generation?},
    author={Theo X. Olausson and Jeevana Priya Inala and Chenglong Wang and Jianfeng Gao and Armando Solar-Lezama},
    booktitle={The Twelfth International Conference on Learning Representations},
    year={2024},
    url={https://openreview.net/forum?id=y0GJXRungR}
}

@inproceedings{stechly2023gpt4,
    title={{GPT}-4 Doesn{\textquoteright}t Know It{\textquoteright}s Wrong: An Analysis of Iterative Prompting for Reasoning Problems},
    author={Kaya Stechly and Matthew Marquez and Subbarao Kambhampati},
    booktitle={NeurIPS 2023 Foundation Models for Decision Making Workshop},
    year={2023},
    url={https://openreview.net/forum?id=PMtZjDYB68}
}

@inproceedings{opsahlong2024miprov2,
    title = "Optimizing Instructions and Demonstrations for Multi-Stage Language Model Programs",
    author = "Opsahl-Ong, Krista  and
      Ryan, Michael J  and
      Purtell, Josh  and
      Broman, David  and
      Potts, Christopher  and
      Zaharia, Matei  and
      Khattab, Omar",
    editor = "Al-Onaizan, Yaser  and
      Bansal, Mohit  and
      Chen, Yun-Nung",
    booktitle = "Proceedings of the 2024 Conference on Empirical Methods in Natural Language Processing",
    month = nov,
    year = "2024",
    address = "Miami, Florida, USA",
    publisher = "Association for Computational Linguistics",
    url = "https://aclanthology.org/2024.emnlp-main.525/",
    doi = "10.18653/v1/2024.emnlp-main.525",
    pages = "9340--9366"
}

@inproceedings{pryzant2023protegi,
    title = "Automatic Prompt Optimization with ``Gradient Descent'' and Beam Search",
    author = "Pryzant, Reid  and
      Iter, Dan  and
      Li, Jerry  and
      Lee, Yin  and
      Zhu, Chenguang  and
      Zeng, Michael",
    editor = "Bouamor, Houda  and
      Pino, Juan  and
      Bali, Kalika",
    booktitle = "Proceedings of the 2023 Conference on Empirical Methods in Natural Language Processing",
    month = dec,
    year = "2023",
    address = "Singapore",
    publisher = "Association for Computational Linguistics",
    url = "https://aclanthology.org/2023.emnlp-main.494/",
    doi = "10.18653/v1/2023.emnlp-main.494",
    pages = "7957--7968"
}

@inproceedings{shinn2023reflexion,
    author = {Shinn, Noah and Cassano, Federico and Gopinath, Ashwin and Narasimhan, Karthik and Yao, Shunyu},
    booktitle = {Advances in Neural Information Processing Systems},
    editor = {A. Oh and T. Naumann and A. Globerson and K. Saenko and M. Hardt and S. Levine},
    pages = {8634--8652},
    publisher = {Curran Associates, Inc.},
    title = {Reflexion: language agents with verbal reinforcement learning},
    url = {https://proceedings.neurips.cc/paper_files/paper/2023/file/1b44b878bb782e6954cd888628510e90-Paper-Conference.pdf},
    volume = {36},
    year = {2023}
}

@inproceedings{yang2023opro,
    title={Large Language Models as Optimizers},
    author={Chengrun Yang and Xuezhi Wang and Yifeng Lu and Hanxiao Liu and Quoc V Le and Denny Zhou and Xinyun Chen},
    booktitle={The Twelfth International Conference on Learning Representations},
    year={2024},
    url={https://openreview.net/forum?id=Bb4VGOWELI}
}

@misc{yuksekgonul2024textgrad,
    title={TextGrad: Automatic "Differentiation" via Text}, 
    author={Mert Yuksekgonul and Federico Bianchi and Joseph Boen and Sheng Liu and Zhi Huang and Carlos Guestrin and James Zou},
    year={2024},
    eprint={2406.07496},
    archivePrefix={arXiv},
    primaryClass={cs.CL},
    url={https://arxiv.org/abs/2406.07496}, 
}

@inproceedings{huang2023selfcorrect,
    title={Large Language Models Cannot Self-Correct Reasoning Yet},
    author={Jie Huang and Xinyun Chen and Swaroop Mishra and Huaixiu Steven Zheng and Adams Wei Yu and Xinying Song and Denny Zhou},
    booktitle={The Twelfth International Conference on Learning Representations},
    year={2024},
    url={https://openreview.net/forum?id=IkmD3fKBPQ}
}

@misc{aime,
    title = {MathArena: Evaluating LLMs on Uncontaminated Math Competitions},
    author = {Mislav Balunovi{\'c} and Jasper Dekoninck and Ivo Petrov and Nikola Jovanovi{\'c} and Martin Vechev},
    url = {https://matharena.ai/},
    publisher = {SRI Lab, ETH Zurich},
    year = {2025},
}

@inproceedings{chen2023instructzero,
    author = {Chen, Lichang and Chen, Jiuhai and Goldstein, Tom and Huang, Heng and Zhou, Tianyi},
    title = {INSTRUCTZERO: efficient instruction optimization for black-box large language models6518},
    year = {2024},
    publisher = {JMLR.org},
    booktitle = {Proceedings of the 41st International Conference on Machine Learning},
    articleno = {251},
    numpages = {16},
    location = {Vienna, Austria},
    series = {ICML'24}
}

@inproceedings{fernando2023promptbreeder,
    author = {Fernando, Chrisantha and Banarse, Dylan and Michalewski, Henryk and Osindero, Simon and Rockt\"{a}schel, Tim},
    title = {Promptbreeder: self-referential self-improvement via prompt evolution},
    year = {2024},
    publisher = {JMLR.org},
    booktitle = {Proceedings of the 41st International Conference on Machine Learning},
    articleno = {541},
    numpages = {64},
    location = {Vienna, Austria},
    series = {ICML'24}
}

@misc{gpt4.1technical,
    title = {Introducing {GPT-4.1} in the {API}},
    author = {OpenAI},
    year = {2025},
    url = {https://openai.com/index/gpt-4-1/},
}

@misc{openai2024gpt4omini,
  title = {GPT-4o mini: advancing cost-efficient intelligence},
  author = {OpenAI},
  year = {2024},
  url = {https://openai.com/index/gpt-4o-mini-advancing-cost-efficient-intelligence/}
}

@inproceedings{guo2023connecting,
    title={Connecting Large Language Models with Evolutionary Algorithms Yields Powerful Prompt Optimizers},
    author={Qingyan Guo and Rui Wang and Junliang Guo and Bei Li and Kaitao Song and Xu Tan and Guoqing Liu and Jiang Bian and Yujiu Yang},
    booktitle={The Twelfth International Conference on Learning Representations},
    year={2024},
    url={https://openreview.net/forum?id=ZG3RaNIsO8}
}

@article{kamoi2024can,
    title = "When Can {LLM}s Actually Correct Their Own Mistakes? A Critical Survey of Self-Correction of {LLM}s",
    author = "Kamoi, Ryo  and
      Zhang, Yusen  and
      Zhang, Nan  and
      Han, Jiawei  and
      Zhang, Rui",
    journal = "Transactions of the Association for Computational Linguistics",
    volume = "12",
    year = "2024",
    address = "Cambridge, MA",
    publisher = "MIT Press",
    url = "https://aclanthology.org/2024.tacl-1.78/",
    doi = "10.1162/tacl_a_00713",
    pages = "1417--1440"
}

@inproceedings{kojima2022large,
    author = {Kojima, Takeshi and Gu, Shixiang (Shane) and Reid, Machel and Matsuo, Yutaka and Iwasawa, Yusuke},
    booktitle = {Advances in Neural Information Processing Systems},
    editor = {S. Koyejo and S. Mohamed and A. Agarwal and D. Belgrave and K. Cho and A. Oh},
    pages = {22199--22213},
    publisher = {Curran Associates, Inc.},
    title = {Large Language Models are Zero-Shot Reasoners},
    url = {https://proceedings.neurips.cc/paper_files/paper/2022/file/8bb0d291acd4acf06ef112099c16f326-Paper-Conference.pdf},
    volume = {35},
    year = {2022}
}

@Inbook{lourenco2003iterated,
    author="Louren{\c{c}}o, Helena R.
    and Martin, Olivier C.
    and St{\"u}tzle, Thomas",
    editor="Glover, Fred
    and Kochenberger, Gary A.",
    title="Iterated Local Search",
    bookTitle="Handbook of Metaheuristics",
    year="2003",
    publisher="Springer US",
    address="Boston, MA",
    pages="320--353",
    isbn="978-0-306-48056-0",
    doi="10.1007/0-306-48056-5_11",
    url="https://doi.org/10.1007/0-306-48056-5_11"
}

@inproceedings{lu2022prompt,
    title = "Fantastically Ordered Prompts and Where to Find Them: Overcoming Few-Shot Prompt Order Sensitivity",
    author = "Lu, Yao  and
      Bartolo, Max  and
      Moore, Alastair  and
      Riedel, Sebastian  and
      Stenetorp, Pontus",
    editor = "Muresan, Smaranda  and
      Nakov, Preslav  and
      Villavicencio, Aline",
    booktitle = "Proceedings of the 60th Annual Meeting of the Association for Computational Linguistics (Volume 1: Long Papers)",
    month = may,
    year = "2022",
    address = "Dublin, Ireland",
    publisher = "Association for Computational Linguistics",
    url = "https://aclanthology.org/2022.acl-long.556/",
    doi = "10.18653/v1/2022.acl-long.556",
    pages = "8086--8098"
}

@misc{qwen3technical,
    title={Qwen3 Technical Report}, 
    author={An Yang and Anfeng Li and Baosong Yang and Beichen Zhang and Binyuan Hui and Bo Zheng and Bowen Yu and Chang Gao and Chengen Huang and Chenxu Lv and Chujie Zheng and Dayiheng Liu and Fan Zhou and Fei Huang and Feng Hu and Hao Ge and Haoran Wei and Huan Lin and Jialong Tang and Jian Yang and Jianhong Tu and Jianwei Zhang and Jianxin Yang and Jiaxi Yang and Jing Zhou and Jingren Zhou and Junyang Lin and Kai Dang and Keqin Bao and Kexin Yang and Le Yu and Lianghao Deng and Mei Li and Mingfeng Xue and Mingze Li and Pei Zhang and Peng Wang and Qin Zhu and Rui Men and Ruize Gao and Shixuan Liu and Shuang Luo and Tianhao Li and Tianyi Tang and Wenbiao Yin and Xingzhang Ren and Xinyu Wang and Xinyu Zhang and Xuancheng Ren and Yang Fan and Yang Su and Yichang Zhang and Yinger Zhang and Yu Wan and Yuqiong Liu and Zekun Wang and Zeyu Cui and Zhenru Zhang and Zhipeng Zhou and Zihan Qiu},
    year={2025},
    eprint={2505.09388},
    archivePrefix={arXiv},
    primaryClass={cs.CL},
    url={https://arxiv.org/abs/2505.09388}, 
}

@article{robbins1952some,
    title={Some aspects of the sequential design of experiments},
    author={Robbins, Herbert},
    journal={Bulletin of the American Mathematical Society},
    volume={58},
    number={5},
    pages={527--535},
    year={1952}
}

@article{sahoo2024systematic,
    title={A Systematic Survey of Prompt Engineering in Large Language Models: Techniques and Applications},
    author={Sahoo, Pranab and Singh, Ayush Kumar and Saha, Sriparna and Jain, Vinija and Mondal, Samrat and Chadha, Aman},
    journal={arXiv e-prints},
    pages={arXiv--2402},
    year={2024}
}

@book{sutton2018reinforcement,
	author = {Richard Sutton and Andrew Barto},
	editor = {},
	publisher = {MIT Press},
	title = {Reinforcement Learning: An Introduction},
	year = {2018}
}

@inproceedings{wang2022selfconsistency,
    title={Self-Consistency Improves Chain of Thought Reasoning in Language Models},
    author={Xuezhi Wang and Jason Wei and Dale Schuurmans and Quoc V Le and Ed H. Chi and Sharan Narang and Aakanksha Chowdhery and Denny Zhou},
    booktitle={The Eleventh International Conference on Learning Representations },
    year={2023},
    url={https://openreview.net/forum?id=1PL1NIMMrw}
}

@inproceedings{wu2023autogen,
    title={AutoGen: Enabling Next-Gen {LLM} Applications via Multi-Agent Conversation},
    author={Qingyun Wu and Gagan Bansal and Jieyu Zhang and Yiran Wu and Beibin Li and Erkang Zhu and Li Jiang and Xiaoyun Zhang and Shaokun Zhang and Jiale Liu and Ahmed Hassan Awadallah and Ryen W White and Doug Burger and Chi Wang},
    booktitle={ICLR 2024 Workshop on Large Language Model (LLM) Agents},
    year={2024},
    url={https://openreview.net/forum?id=uAjxFFing2}
}

@inproceedings{ye2023prompt,
    title = "Prompt Engineering a Prompt Engineer",
    author = "Ye, Qinyuan  and
      Ahmed, Mohamed  and
      Pryzant, Reid  and
      Khani, Fereshte",
    editor = "Ku, Lun-Wei  and
      Martins, Andre  and
      Srikumar, Vivek",
    booktitle = "Findings of the Association for Computational Linguistics: ACL 2024",
    month = aug,
    year = "2024",
    address = "Bangkok, Thailand",
    publisher = "Association for Computational Linguistics",
    url = "https://aclanthology.org/2024.findings-acl.21/",
    doi = "10.18653/v1/2024.findings-acl.21",
    pages = "355--385"
}

@InProceedings{zhao2021calibrate,
    title = 	 {Calibrate Before Use: Improving Few-shot Performance of Language Models},
    author =       {Zhao, Zihao and Wallace, Eric and Feng, Shi and Klein, Dan and Singh, Sameer},
    booktitle = 	 {Proceedings of the 38th International Conference on Machine Learning},
    pages = 	 {12697--12706},
    year = 	 {2021},
    editor = 	 {Meila, Marina and Zhang, Tong},
    volume = 	 {139},
    series = 	 {Proceedings of Machine Learning Research},
    month = 	 {18--24 Jul},
    publisher =    {PMLR},
    url = 	 {https://proceedings.mlr.press/v139/zhao21c.html}
}

\clearpage
\newpage
\appendix

    

    

    
    
    

\section{Discussion}
\label{sec:discussion}

\subsection{Data-Driven Heuristics}

The current heuristic set is manually curated from representative failure cases and modes. While this ensures coverage of known out-of-prior failure modes, the approach does not scale and may miss task-specific failure patterns underrepresented in the literature.

A more scalable alternative is data-driven construction: collecting 
large-scale optimization trajectories across diverse tasks and 
reflectors, clustering failure cases by semantic similarity, and 
distilling recurring patterns into heuristic categories automatically. 
Frequently-winning free hypotheses from the $\varepsilon$-exploration 
branch are a natural seed for this process---each successful free 
hypothesis that recurs across multiple runs is a candidate for 
promotion, enabling the heuristic set to grow with accumulated 
experience. 

This process has a bootstrap dependency, however: early 
trajectories are collected under the manually curated heuristic set 
and may underrepresent failure modes outside its scope. Seeding with 
diverse random-restart trajectories, where $p=1$ throughout, provides 
a less biased initialization for the data-driven construction process.

\subsection{Broader Semantic Trace Utility}

The semantic trace tree already enables interpretable optimization histories and oscillation detection, but its potential extends further. The trace encodes a causal graph of which hypotheses led to which improvements, providing rich information that could actively guide future decisions---for instance, informing the System-Aware Merge~\cite{agrawal2025gepa} step by preferring merges between candidates whose root-cause label sequences are complementary, or prioritizing hypothesis categories that have historically been productive. We view the semantic trace as a general-purpose interpretability substrate for APO, and expect its utility to grow as more sophisticated uses are developed.

The semantic trace also has potential for cross-task transfer: 
hypothesis categories that consistently produce large $\delta$ 
gains across multiple tasks constitute a task-agnostic prior 
that can warm-start optimization on new tasks, reducing the 
rounds needed to identify the dominant failure mode.

\subsection{Adaptive Explore-Exploit Scheduling}

The current two-layer mechanism uses fixed $p$ and $\varepsilon$ throughout optimization. In practice, the optimal balance shifts as optimization progresses: early rounds benefit from broader exploration to identify the dominant failure mode, while later rounds benefit from focused exploitation once a productive direction has been found.

The semantic trace provides a natural signal for adaptive scheduling. When the trace shows consistent improvement in a single hypothesis category, $\varepsilon$ could be reduced to concentrate resources on exploitation. When oscillation is detected or improvement plateaus, $p$ could be temporarily increased to trigger more aggressive restart behavior. Formalizing this as a bandit problem~\cite{robbins1952some} over 
the heuristic set---where the reward signal for each category 
$c \in \mathcal{C}$ is its empirical mean $\Delta\text{acc}$ across 
historical selections---is a principled direction for future work.

\section{Experimental Setup (Contd.)}
\label{sec:appendix_config}

\paragraph{Hardware.} All experiments were conducted on a single 
server equipped with one NVIDIA RTX~4090 (24\,GB). Local inference 
for Qwen3-4B and Qwen3-8B was served via Ollama; GPT-4.1-mini and 
GPT-4o-mini were accessed through the OpenAI API.

\paragraph{Experimental Parameters.} All experiments share the 
following configuration: minibatch size $b = 8$, budget $T = 500$, 
training/validation sizes of 50 each, maximum parallelism of 
4 workers, and random seed 0. Evaluation uses exact-match accuracy. 
For GEPA, we use the default hyperparameters from the original 
implementation. For VISTA, the default configuration is $K = 3$, 
$p = 0.2$, and $\varepsilon = 0.1$; ablation groups vary one 
parameter at a time.

\paragraph{Datasets.} For GSM8K, we use the official 
\texttt{openai/gsm8k} dataset; working training and validation 
sets are both sampled from the official training split and fixed 
for each run. The full official test split (1,319 examples) serves 
as the test set. For AIME, working training and validation sets 
are sampled from \texttt{AI-MO/aimo-validation-aime} (problems 
from AIME 2022--2024) and fixed for each run. The test set 
consists of the 30 problems from \texttt{MathArena/aime\_2025} 
(AIME 2025 I \&\ II), each repeated 
5 times to reduce evaluation variance, yielding 150 test instances 
in total.

\paragraph{Model Parameters.} All local models are loaded in 
bfloat16 precision without quantization. The base Qwen-3 model uses 
$\texttt{temperature}{=}0.6$, $\texttt{top-p}{=}0.95$, 
$\texttt{top-k}{=}20$, $\texttt{min-p}{=}0$, and 
$\texttt{presence\_penalty}{=}1.5$, with thinking mode disabled 
(\texttt{reasoning\_effort=none}). Hypothesis and reflection 
agents use default sampling parameters. Maximum generation 
length is set to the model default for all roles.

\section{Computational Cost}
\label{sec:appendix_cost}

All local model inference (Qwen3-4B, Qwen3-8B) incurs no API cost. 
For experiments involving OpenAI models, total expenditure across 
all reported groups amounts to approximately \$34.82--51.92. 
On GSM8K with a local base model (Qwen3-4B) and GPT-4o-mini 
reflector, a single VISTA run ($T{=}500$, $K{=}3$) costs \$0.20 
and a single GEPA run costs \$0.12. On AIME2025 with GPT-4.1-mini 
as base model, a VISTA run costs \$4.1--6.0 and a GEPA run costs 
\$3.7--5.6. The marginal cost of the hypothesis agent relative to 
GEPA is \$0.08--0.40 per run, confirming that VISTA's diagnostic 
capability comes at negligible additional expense.

\onecolumn

\section{Optimization Trees}
\label{sec:appendix_trees}

\tikzset{
    opt node/.style={font=\fontsize{7}{8}\selectfont},
    prompt/.style={circle, draw=blue!50, fill=blue!5, minimum size=0.75cm, inner sep=1pt, align=center},
    rejprompt/.style={circle, draw=black!25, dashed, fill=gray!8, minimum size=0.75cm, inner sep=1pt, align=center},
    bestprompt/.style={circle, draw=blue!50, fill=blue!5, minimum size=0.75cm, inner sep=1pt, align=center},
    selectedge/.style={->, blue!60, thick},
    normaledge/.style={->, blue!60, thick},
    rejedge/.style={->, dashed, black!25, semithick},
    slabel/.style={font=\fontsize{4.8}{5.8}\selectfont, text=blue!70, fill=white, inner sep=1.2pt, sloped, midway, above},
    nlabel/.style={font=\fontsize{4.8}{5.8}\selectfont, text=black!55, fill=white, inner sep=1.2pt, sloped, midway, above},
    qlabel/.style={font=\fontsize{6}{7}\selectfont, text=black!50, fill=white, inner sep=1pt, sloped},
    gepa_normaledge/.style={->, red!60, thick},
    gepa_prompt/.style={circle, draw=red!50, fill=red!5, minimum size=0.75cm, inner sep=1pt, align=center},
}

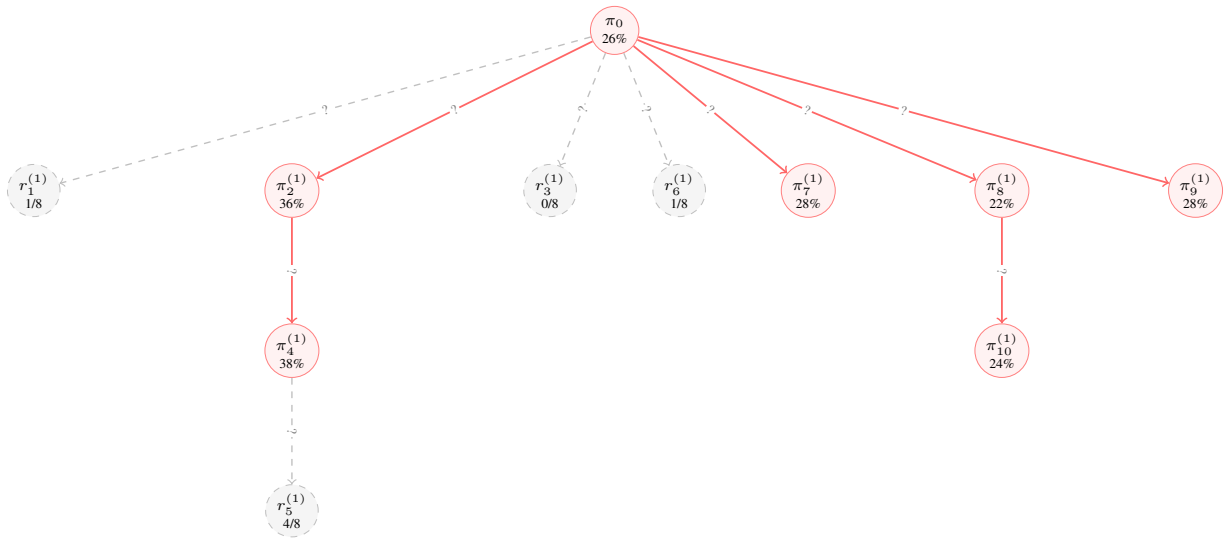
\begin{figure}[H]
\centering
\resizebox{\textwidth}{!}{%
\begin{tikzpicture}[every node/.style=opt node]
    \node[gepa_prompt] (p0) at (0.0, 0.0) {$\pi_{0}$ \\ \tiny 26\%};
    \node[rejprompt] (r1_1) at (-9.0, -2.5) {$r_{1}^{(1)}$ \\ \tiny 1/8};
    \node[gepa_prompt] (p2_1) at (-5.0, -2.5) {$\pi_{2}^{(1)}$ \\ \tiny 36\%};
    \node[rejprompt] (r3_1) at (-1.0, -2.5) {$r_{3}^{(1)}$ \\ \tiny 0/8};
    \node[rejprompt] (r6_1) at (1.0, -2.5) {$r_{6}^{(1)}$ \\ \tiny 1/8};
    \node[gepa_prompt] (p7_1) at (3.0, -2.5) {$\pi_{7}^{(1)}$ \\ \tiny 28\%};
    \node[gepa_prompt] (p8_1) at (6.0, -2.5) {$\pi_{8}^{(1)}$ \\ \tiny 22\%};
    \node[gepa_prompt] (p9_1) at (9.0, -2.5) {$\pi_{9}^{(1)}$ \\ \tiny 28\%};
    \node[gepa_prompt] (p4_1) at (-5.0, -5.0) {$\pi_{4}^{(1)}$ \\ \tiny 38\%};
    \node[gepa_prompt] (p10_1) at (6.0, -5.0) {$\pi_{10}^{(1)}$ \\ \tiny 24\%};
    \node[rejprompt] (r5_1) at (-5.0, -7.5) {$r_{5}^{(1)}$ \\ \tiny 4/8};
    \draw[rejedge] (p0) -- node[midway, qlabel] {?} (r1_1);
    \draw[gepa_normaledge] (p0) -- node[midway, qlabel] {?} (p2_1);
    \draw[rejedge] (p0) -- node[midway, qlabel] {?} (r3_1);
    \draw[rejedge] (p0) -- node[midway, qlabel] {?} (r6_1);
    \draw[gepa_normaledge] (p0) -- node[midway, qlabel] {?} (p7_1);
    \draw[gepa_normaledge] (p0) -- node[midway, qlabel] {?} (p8_1);
    \draw[gepa_normaledge] (p0) -- node[midway, qlabel] {?} (p9_1);
    \draw[gepa_normaledge] (p2_1) -- node[midway, qlabel] {?} (p4_1);
    \draw[gepa_normaledge] (p8_1) -- node[midway, qlabel] {?} (p10_1);
    \draw[rejedge] (p4_1) -- node[midway, qlabel] {?} (r5_1);
\end{tikzpicture}%
}
\caption{GEPA optimization tree under the defective seed. Every edge carries only a question mark; no root-cause label or accuracy gain $\delta$ is recorded for any transition. The structural 
root cause is never identified and optimization stagnates 
at 38\%.}
\label{fig:tree_gepa_defective}
\end{figure}

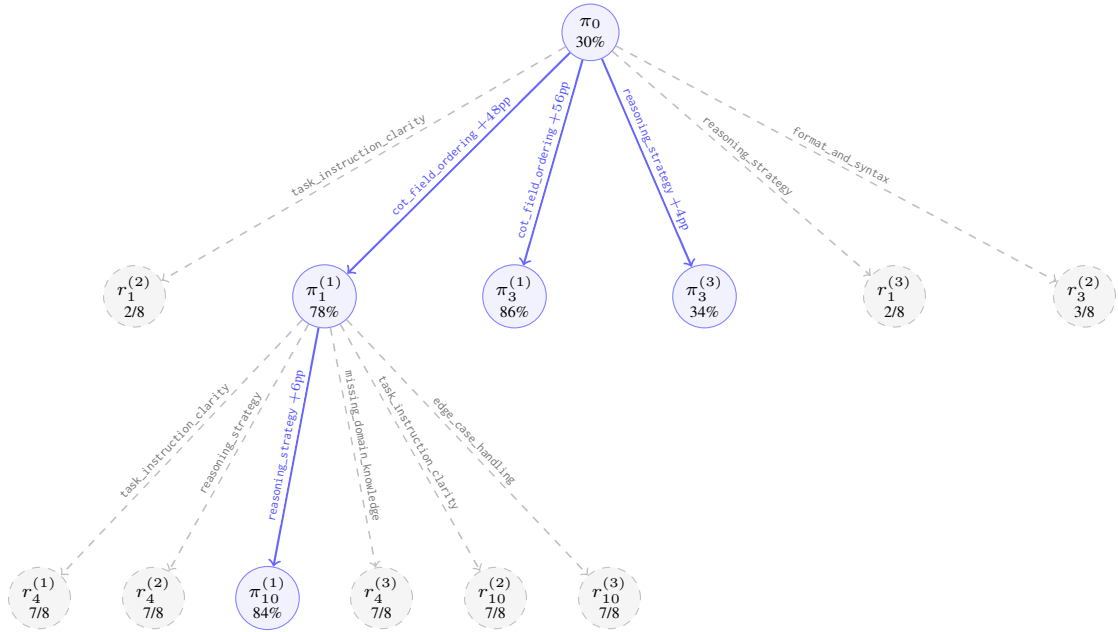
\begin{figure}[H]
\centering
\begin{tikzpicture}[every node/.style=opt node]

\node[prompt] (p0) at (0, 0) {$\pi_{0}$ \\ \tiny 30\%};

\node[rejprompt] (r12) at (-6.0, -3.5) {$r_{1}^{(2)}$ \\ \tiny 2/8};
\node[bestprompt] (p1) at (-3.5, -3.5) {$\pi_{1}^{(1)}$ \\ \tiny 78\%};
\node[bestprompt] (p2) at (-1.0, -3.5) {$\pi_{3}^{(1)}$ \\ \tiny 86\%};
\node[prompt] (p3) at (1.5, -3.5) {$\pi_{3}^{(3)}$ \\ \tiny 34\%};
\node[rejprompt] (r13) at (4.0, -3.5) {$r_{1}^{(3)}$ \\ \tiny 2/8};
\node[rejprompt] (r32) at (6.5, -3.5) {$r_{3}^{(2)}$ \\ \tiny 3/8};

\node[rejprompt] (r41) at (-7.25, -7.5) {$r_{4}^{(1)}$ \\ \tiny 7/8};
\node[rejprompt] (r42) at (-5.75, -7.5) {$r_{4}^{(2)}$ \\ \tiny 7/8};
\node[bestprompt] (p4) at (-4.25, -7.5) {$\pi_{10}^{(1)}$ \\ \tiny 84\%};
\node[rejprompt] (r43) at (-2.75, -7.5) {$r_{4}^{(3)}$ \\ \tiny 7/8};
\node[rejprompt] (r102) at (-1.25, -7.5) {$r_{10}^{(2)}$ \\ \tiny 7/8};
\node[rejprompt] (r103) at (0.25, -7.5) {$r_{10}^{(3)}$ \\ \tiny 7/8};

\draw[rejedge] (p0) -- node[nlabel] {\texttt{task\_instruction\_clarity}} (r12);
\draw[selectedge] (p0) -- node[slabel] {\texttt{cot\_field\_ordering} $+48$pp} (p1);
\draw[selectedge] (p0) -- node[slabel] {\texttt{cot\_field\_ordering} $+56$pp} (p2);
\draw[normaledge] (p0) -- node[slabel] {\texttt{reasoning\_strategy} $+4$pp} (p3);
\draw[rejedge] (p0) -- node[nlabel] {\texttt{reasoning\_strategy}} (r13);
\draw[rejedge] (p0) -- node[nlabel] {\texttt{format\_and\_syntax}} (r32);

\draw[rejedge] (p1) -- node[nlabel] {\texttt{task\_instruction\_clarity}} (r41);
\draw[rejedge] (p1) -- node[nlabel] {\texttt{reasoning\_strategy}} (r42);
\draw[selectedge] (p1) -- node[slabel] {\texttt{reasoning\_strategy} $+6$pp} (p4);
\draw[rejedge] (p1) -- node[nlabel] {\texttt{missing\_domain\_knowledge}} (r43);
\draw[rejedge] (p1) -- node[nlabel] {\texttt{task\_instruction\_clarity}} (r102);
\draw[rejedge] (p1) -- node[nlabel] {\texttt{edge\_case\_handling}} (r103);

\end{tikzpicture}
\caption{VISTA optimization tree under the defective seed. Every edge indicates a hypothesis annotated with a root-cause label and $\delta$. Iteration~1 immediately identifies \texttt{cot\_field\_ordering} as the structural root cause, achieving a $+48$pp jump to 78\% accuracy; iteration~3 further reaches 86\% via the same diagnosis.}
\label{fig:tree_vista_defective}
\end{figure}

Figures~\ref{fig:tree_gepa_defective} and \ref{fig:tree_vista_defective} show the optimization trees for GEPA and VISTA under the defective seed (Qwen3-4B base, Qwen3-8B reflector, GSM8K), where each node $v \in V$ corresponds to a prompt candidate and each edge $(u,v) \in E$ records the root-cause label and accuracy gain $\delta$ of the optimization step that produced $v$ from $u$. We denote the initial base prompt as $\pi_0$. For all subsequent nodes, we denote accepted prompts as $\pi_i^{(k)}$ and rejected candidates as $r_i^{(k)}$, where the subscript $i$ is the iteration index in which the candidate was generated, and the superscript $k$ is its candidate index within that iteration. GEPA's tree carries no semantic labels on any edge; VISTA's tree annotates every transition with $(c^*, \delta)$. Red and blue edges indicate accepted updates within an iteration, for GEPA and VISTA, respectively.

\begin{figure}[H]
\centering
\begin{tikzpicture}[every node/.style=opt node]

\node[gepa_prompt] (s0) at (0, 0) {$\pi_{0}$ \\ \tiny 86\%};

\node[rejprompt] (r1) at (-7.0, -3.5) {$r_{1}^{(1)}$ \\ \tiny 6/8};
\node[rejprompt] (r2) at (-6.0, -3.5) {$r_{2}^{(1)}$ \\ \tiny 3/8};
\node[rejprompt] (r3) at (-5.0, -3.5) {$r_{3}^{(1)}$ \\ \tiny 0/8};
\node[rejprompt] (r4) at (-4.0, -3.5) {$r_{4}^{(1)}$ \\ \tiny 7/8};
\node[gepa_prompt] (p8) at (-2.5, -3.5) {$\pi_{8}^{(1)}$ \\ \tiny 88\%};
\node[rejprompt] (r5) at (-1.0, -3.5) {$r_{5}^{(1)}$ \\ \tiny 0/8};
\node[rejprompt] (r6) at (0.0, -3.5) {$r_{6}^{(1)}$ \\ \tiny 5/8};
\node[rejprompt] (r7) at (1.0, -3.5) {$r_{7}^{(1)}$ \\ \tiny 8/8};
\node[rejprompt] (r9) at (2.0, -3.5) {$r_{9}^{(1)}$ \\ \tiny 7/8};
\node[gepa_prompt] (p20) at (3.5, -3.5) {$\pi_{20}^{(1)}$ \\ \tiny 88\%};
\node[rejprompt] (r14) at (5.0, -3.5) {$r_{14}^{(1)}$ \\ \tiny 5/8};
\node[rejprompt] (r17) at (6.0, -3.5) {$r_{17}^{(1)}$ \\ \tiny 8/8};
\node[rejprompt] (r19) at (7.0, -3.5) {$r_{19}^{(1)}$ \\ \tiny 7/8};

\node[rejprompt] (r10) at (-5.5, -7.5) {$r_{10}^{(1)}$ \\ \tiny 7/8};
\node[rejprompt] (r11) at (-4.5, -7.5) {$r_{11}^{(1)}$ \\ \tiny 8/8};
\node[rejprompt] (r12) at (-3.5, -7.5) {$r_{12}^{(1)}$ \\ \tiny 7/8};
\node[gepa_prompt] (p15) at (-2.5, -7.5) {$\pi_{15}^{(1)}$ \\ \tiny 86\%};
\node[rejprompt] (r13) at (-1.5, -7.5) {$r_{13}^{(1)}$ \\ \tiny 8/8};
\node[rejprompt] (r16) at (-0.5, -7.5) {$r_{16}^{(1)}$ \\ \tiny 7/8};
\node[rejprompt] (r18) at (0.5, -7.5) {$r_{18}^{(1)}$ \\ \tiny 8/8};

\draw[rejedge] (s0) -- node[midway, qlabel] {?} (r1);
\draw[rejedge] (s0) -- node[midway, qlabel] {?} (r2);
\draw[rejedge] (s0) -- node[midway, qlabel] {?} (r3);
\draw[rejedge] (s0) -- node[midway, qlabel] {?} (r4);
\draw[gepa_normaledge] (s0) -- node[midway, qlabel] {?} (p8);
\draw[rejedge] (s0) -- node[midway, qlabel] {?} (r5);
\draw[rejedge] (s0) -- node[midway, qlabel] {?} (r6);
\draw[rejedge] (s0) -- node[midway, qlabel] {?} (r7);
\draw[rejedge] (s0) -- node[midway, qlabel] {?} (r9);
\draw[gepa_normaledge] (s0) -- node[midway, qlabel] {?} (p20);
\draw[rejedge] (s0) -- node[midway, qlabel] {?} (r14);
\draw[rejedge] (s0) -- node[midway, qlabel] {?} (r17);
\draw[rejedge] (s0) -- node[midway, qlabel] {?} (r19);

\draw[rejedge] (p8) -- node[midway, qlabel] {?} (r10);
\draw[rejedge] (p8) -- node[midway, qlabel] {?} (r11);
\draw[rejedge] (p8) -- node[midway, qlabel] {?} (r12);
\draw[gepa_normaledge] (p8) -- node[midway, qlabel] {?} (p15);
\draw[rejedge] (p8) -- node[midway, qlabel] {?} (r13);
\draw[rejedge] (p8) -- node[midway, qlabel] {?} (r16);
\draw[rejedge] (p8) -- node[midway, qlabel] {?} (r18);

\end{tikzpicture}
\caption{GEPA optimization tree under the repaired seed. The trace follows a single-candidate mutation path; two independent branches successfully reach 88\% accuracy, while one subsequent update on the left branch regresses to 86\%.}
\label{fig:tree_gepa_repaired}
\end{figure}
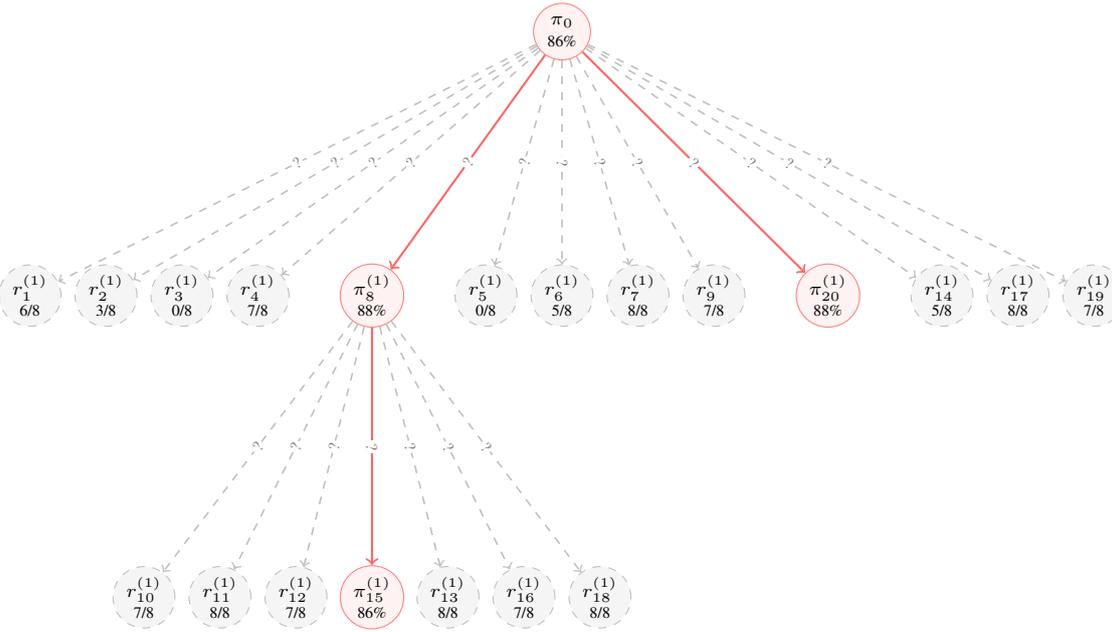

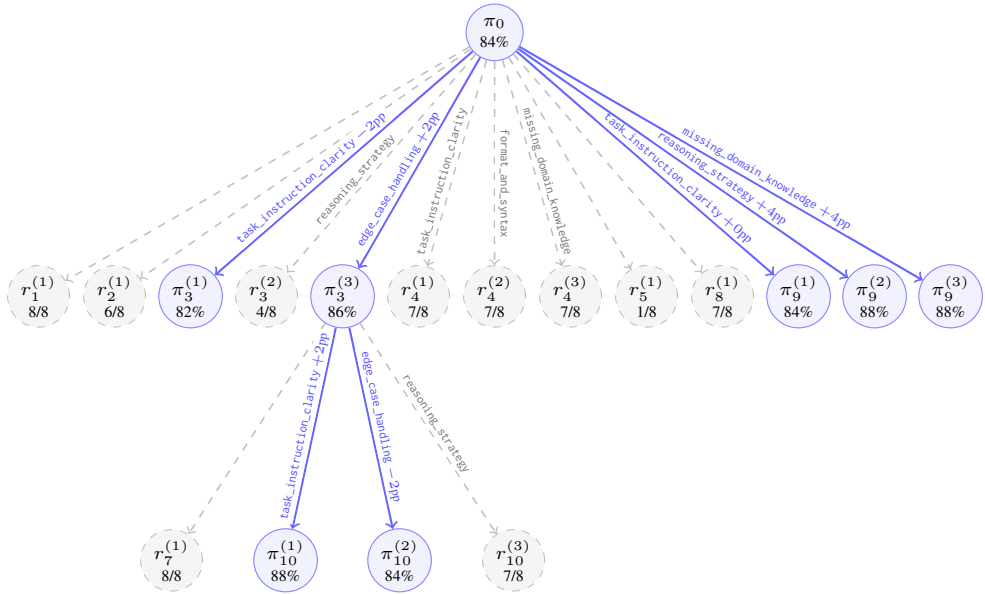
\begin{figure}[H]
\centering
\begin{tikzpicture}[every node/.style=opt node]

\node[prompt] (v0) at (0, 0) {$\pi_{0}$ \\ \tiny 84\%};

\node[rejprompt] (r1_1) at (-6.0, -3.5) {$r_{1}^{(1)}$ \\ \tiny 8/8};
\node[rejprompt] (r2_1) at (-5.0, -3.5) {$r_{2}^{(1)}$ \\ \tiny 6/8};
\node[prompt] (p3_1) at (-4.0, -3.5) {$\pi_{3}^{(1)}$ \\ \tiny 82\%};
\node[rejprompt] (r3_2) at (-3.0, -3.5) {$r_{3}^{(2)}$ \\ \tiny 4/8};
\node[bestprompt] (p3_3) at (-2.0, -3.5) {$\pi_{3}^{(3)}$ \\ \tiny 86\%};
\node[rejprompt] (r4_1) at (-1.0, -3.5) {$r_{4}^{(1)}$ \\ \tiny 7/8};
\node[rejprompt] (r4_2) at (0.0, -3.5) {$r_{4}^{(2)}$ \\ \tiny 7/8};
\node[rejprompt] (r4_3) at (1.0, -3.5) {$r_{4}^{(3)}$ \\ \tiny 7/8};
\node[rejprompt] (r5_1) at (2.0, -3.5) {$r_{5}^{(1)}$ \\ \tiny 1/8};
\node[rejprompt] (r8_1) at (3.0, -3.5) {$r_{8}^{(1)}$ \\ \tiny 7/8};
\node[prompt] (p9_1) at (4.0, -3.5) {$\pi_{9}^{(1)}$ \\ \tiny 84\%};
\node[bestprompt] (p9_2) at (5.0, -3.5) {$\pi_{9}^{(2)}$ \\ \tiny 88\%};
\node[bestprompt] (p9_3) at (6.0, -3.5) {$\pi_{9}^{(3)}$ \\ \tiny 88\%};

\node[rejprompt] (r7_1) at (-4.25, -7.0) {$r_{7}^{(1)}$ \\ \tiny 8/8};
\node[bestprompt] (p10_1) at (-2.75, -7.0) {$\pi_{10}^{(1)}$ \\ \tiny 88\%};
\node[prompt] (p10_2) at (-1.25, -7.0) {$\pi_{10}^{(2)}$ \\ \tiny 84\%};
\node[rejprompt] (r10_3) at (0.25, -7.0) {$r_{10}^{(3)}$ \\ \tiny 7/8};

\draw[rejedge] (v0) -- (r1_1);
\draw[rejedge] (v0) -- (r2_1);
\draw[normaledge] (v0) -- node[slabel, pos=0.6] {\texttt{task\_instruction\_clarity} $-2$pp} (p3_1);
\draw[rejedge] (v0) -- node[nlabel, pos=0.6] {\texttt{reasoning\_strategy}} (r3_2);
\draw[selectedge] (v0) -- node[slabel, pos=0.6] {\texttt{edge\_case\_handling} $+2$pp} (p3_3);
\draw[rejedge] (v0) -- node[nlabel, pos=0.6] {\texttt{task\_instruction\_clarity}} (r4_1);
\draw[rejedge] (v0) -- node[nlabel, pos=0.6] {\texttt{format\_and\_syntax}} (r4_2);
\draw[rejedge] (v0) -- node[nlabel, pos=0.6] {\texttt{missing\_domain\_knowledge}} (r4_3);
\draw[rejedge] (v0) -- (r5_1);
\draw[rejedge] (v0) -- (r8_1);
\draw[normaledge] (v0) -- node[slabel, pos=0.6] {\texttt{task\_instruction\_clarity} $+0$pp} (p9_1);
\draw[selectedge] (v0) -- node[slabel, pos=0.6] {\texttt{reasoning\_strategy} $+4$pp} (p9_2);
\draw[selectedge] (v0) -- node[slabel, pos=0.6] {\texttt{missing\_domain\_knowledge} $+4$pp} (p9_3);

\draw[rejedge] (p3_3) -- (r7_1);
\draw[selectedge] (p3_3) -- node[slabel] {\texttt{task\_instruction\_clarity} $+2$pp} (p10_1);
\draw[normaledge] (p3_3) -- node[slabel] {\texttt{edge\_case\_handling} $-2$pp} (p10_2);
\draw[rejedge] (p3_3) -- node[nlabel] {\texttt{reasoning\_strategy}} (r10_3);

\end{tikzpicture}
\caption{VISTA optimization tree under the repaired seed. The run branches early through parallel hypotheses in iterations 3, 4, and 9, reaching 88\% peak accuracy via multiple distinct, semantically tagged paths. Unlabeled edges indicate iterations where no failure cases were collected on the minibatch of its parent, falling back to a single-mutation step as in GEPA.}
\label{fig:tree_vista_repaired}
\end{figure}

Figures~\ref{fig:tree_gepa_repaired} and \ref{fig:tree_vista_repaired} show the optimization trees for GEPA and VISTA under the repaired seed, following the same experimental setting and notation as above.

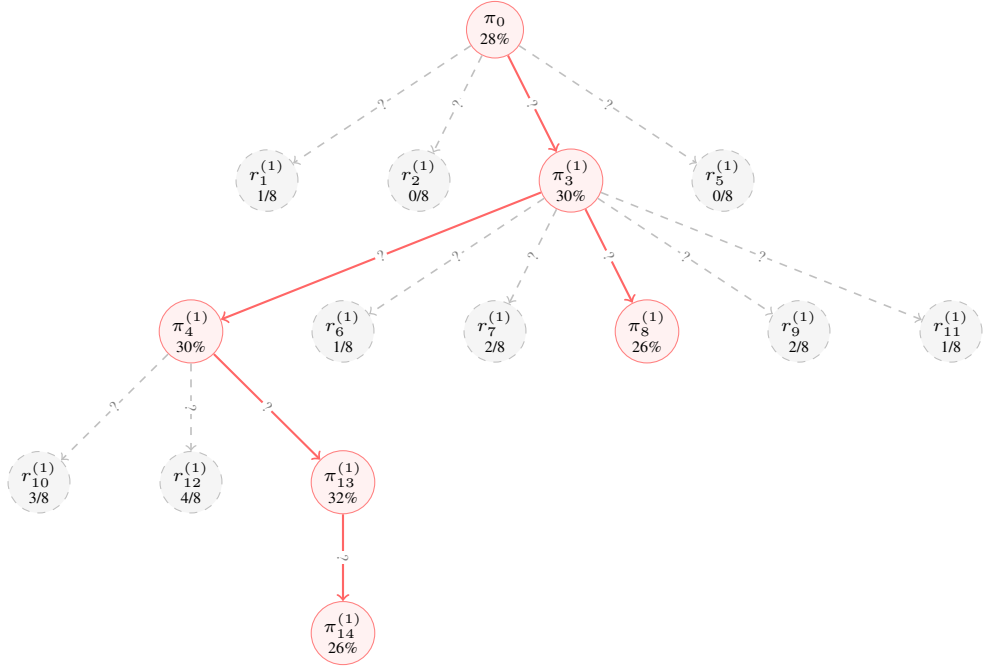
\begin{figure}[H]
\centering
\begin{tikzpicture}[every node/.style=opt node]

\node[gepa_prompt] (s0) at (0, 0) {$\pi_{0}$ \\ \tiny 28\%};

\node[rejprompt] (r1) at (-3.0, -2.0) {$r_{1}^{(1)}$ \\ \tiny 1/8};
\node[rejprompt] (r2) at (-1.0, -2.0) {$r_{2}^{(1)}$ \\ \tiny 0/8};
\node[gepa_prompt] (p3) at (1.0, -2.0) {$\pi_{3}^{(1)}$ \\ \tiny 30\%};
\node[rejprompt] (r5) at (3.0, -2.0) {$r_{5}^{(1)}$ \\ \tiny 0/8};

\node[gepa_prompt] (p4) at (-4.0, -4.0) {$\pi_{4}^{(1)}$ \\ \tiny 30\%};
\node[rejprompt] (r6) at (-2.0, -4.0) {$r_{6}^{(1)}$ \\ \tiny 1/8};
\node[rejprompt] (r7) at (0.0, -4.0) {$r_{7}^{(1)}$ \\ \tiny 2/8};
\node[gepa_prompt] (p8) at (2.0, -4.0) {$\pi_{8}^{(1)}$ \\ \tiny 26\%};
\node[rejprompt] (r9) at (4.0, -4.0) {$r_{9}^{(1)}$ \\ \tiny 2/8};
\node[rejprompt] (r11) at (6.0, -4.0) {$r_{11}^{(1)}$ \\ \tiny 1/8};

\node[rejprompt] (r10) at (-6.0, -6.0) {$r_{10}^{(1)}$ \\ \tiny 3/8};
\node[rejprompt] (r12) at (-4.0, -6.0) {$r_{12}^{(1)}$ \\ \tiny 4/8};
\node[gepa_prompt] (p13) at (-2.0, -6.0) {$\pi_{13}^{(1)}$ \\ \tiny 32\%};

\node[gepa_prompt] (p14) at (-2.0, -8.0) {$\pi_{14}^{(1)}$ \\ \tiny 26\%};

\draw[rejedge] (s0) -- node[midway, qlabel] {?} (r1);
\draw[rejedge] (s0) -- node[midway, qlabel] {?} (r2);
\draw[gepa_normaledge] (s0) -- node[midway, qlabel] {?} (p3);
\draw[rejedge] (s0) -- node[midway, qlabel] {?} (r5);

\draw[gepa_normaledge] (p3) -- node[midway, qlabel] {?} (p4);
\draw[rejedge] (p3) -- node[midway, qlabel] {?} (r6);
\draw[rejedge] (p3) -- node[midway, qlabel] {?} (r7);
\draw[gepa_normaledge] (p3) -- node[midway, qlabel] {?} (p8);
\draw[rejedge] (p3) -- node[midway, qlabel] {?} (r9);
\draw[rejedge] (p3) -- node[midway, qlabel] {?} (r11);

\draw[rejedge] (p4) -- node[midway, qlabel] {?} (r10);
\draw[rejedge] (p4) -- node[midway, qlabel] {?} (r12);
\draw[gepa_normaledge] (p4) -- node[midway, qlabel] {?} (p13);

\draw[gepa_normaledge] (p13) -- node[midway, qlabel] {?} (p14);

\end{tikzpicture}
\caption{GEPA optimization tree under the minimal seed. The main chain reaches a peak of 32\% at iteration 13 before regressing to 26\% in the next update.}
\label{fig:tree_gepa_minimal}
\end{figure}

\begin{figure}[H]
\centering
\begin{tikzpicture}[every node/.style=opt node]

\node[prompt] (v0) at (0, 0) {$\pi_{0}$ \\ \tiny 26\%};

\node[rejprompt] (r1_1) at (-6.0, -3.5) {$r_{1}^{(1)}$ \\ \tiny 0/8};
\node[rejprompt] (r1_2) at (-4.8, -3.5) {$r_{1}^{(2)}$ \\ \tiny 2/8};
\node[rejprompt] (r1_3) at (-3.6, -3.5) {$r_{1}^{(3)}$ \\ \tiny 1/8};
\node[bestprompt] (p2_1) at (-2.4, -3.5) {$\pi_{2}^{(1)}$ \\ \tiny 76\%};
\node[rejprompt] (r2_2) at (-1.2, -3.5) {$r_{2}^{(2)}$ \\ \tiny 1/8};
\node[rejprompt] (r2_3) at (0.0, -3.5) {$r_{2}^{(3)}$ \\ \tiny 2/8};
\node[prompt] (p3_1) at (1.2, -3.5) {$\pi_{3}^{(1)}$ \\ \tiny 32\%};
\node[prompt] (p3_2) at (2.4, -3.5) {$\pi_{3}^{(2)}$ \\ \tiny 30\%};
\node[rejprompt] (r3_3) at (3.6, -3.5) {$r_{3}^{(3)}$ \\ \tiny 0/8};

\node[rejprompt] (r4_1) at (-4.8, -7.0) {$r_{4}^{(1)}$ \\ \tiny 6/8};
\node[rejprompt] (r4_2) at (-3.2, -7.0) {$r_{4}^{(2)}$ \\ \tiny 5/8};
\node[rejprompt] (r4_3) at (-1.6, -7.0) {$r_{4}^{(3)}$ \\ \tiny 5/8};
\node[bestprompt] (p5_1) at (0.0, -7.0) {$\pi_{5}^{(1)}$ \\ \tiny 86\%};
\node[prompt] (p5_2) at (1.6, -7.0) {$\pi_{5}^{(2)}$ \\ \tiny 72\%};
\node[rejprompt] (r5_3) at (3.2, -7.0) {$r_{5}^{(3)}$ \\ \tiny 0/8};

\node[prompt] (p6_1) at (-0.8, -10.5) {$\pi_{6}^{(1)}$ \\ \tiny 74\%};
\node[prompt] (p6_2) at (1.6, -10.5) {$\pi_{6}^{(2)}$ \\ \tiny 76\%};
\node[bestprompt] (p6_3) at (4.0, -10.5) {$\pi_{6}^{(3)}$ \\ \tiny 80\%};

\draw[rejedge] (v0) -- node[nlabel] {\texttt{cot\_field\_ordering}} (r1_1);
\draw[rejedge] (v0) -- node[nlabel] {\texttt{format\_and\_syntax}} (r1_2);
\draw[rejedge] (v0) -- node[nlabel] {\texttt{task\_instruction\_clarity}} (r1_3);
\draw[selectedge] (v0) -- node[slabel] {\texttt{cot\_field\_ordering} $+50$pp} (p2_1);
\draw[rejedge] (v0) -- node[nlabel] {\texttt{task\_instruction\_clarity}} (r2_2);
\draw[rejedge] (v0) -- node[nlabel] {\texttt{missing\_domain\_knowledge}} (r2_3);
\draw[normaledge] (v0) -- node[slabel] {\texttt{format\_and\_syntax} $+6$pp} (p3_1);
\draw[normaledge] (v0) -- node[slabel] {\texttt{task\_instruction\_clarity} $+4$pp} (p3_2);
\draw[rejedge] (v0) -- node[nlabel] {\texttt{missing\_domain\_knowledge}} (r3_3);

\draw[rejedge] (p2_1) -- node[nlabel] {\texttt{format\_and\_syntax}} (r4_1);
\draw[rejedge] (p2_1) -- node[nlabel] {\texttt{task\_instruction\_clarity}} (r4_2);
\draw[rejedge] (p2_1) -- node[nlabel] {\texttt{reasoning\_strategy}} (r4_3);
\draw[selectedge] (p2_1) -- node[slabel] {\texttt{format\_and\_syntax} $+10$pp} (p5_1);
\draw[normaledge] (p2_1) -- node[slabel] {\texttt{reasoning\_strategy} $-4$pp} (p5_2);
\draw[rejedge] (p2_1) -- node[nlabel] {\texttt{task\_instruction\_clarity}} (r5_3);

\draw[normaledge] (p5_2) -- node[slabel] {\texttt{format\_and\_syntax} $+2$pp} (p6_1);
\draw[normaledge] (p5_2) -- node[slabel] {\texttt{reasoning\_strategy} $+4$pp} (p6_2);
\draw[selectedge] (p5_2) -- node[slabel] {\texttt{edge\_case\_handling} $+8$pp} (p6_3);

\end{tikzpicture}
\caption{VISTA optimization tree under the minimal seed. Iteration~2 immediately identifies \texttt{cot\_field\_ordering} as a high-impact root cause ($+50$pp), then improves further through semantically tagged branches.}
\label{fig:tree_vista_minimal}
\end{figure}
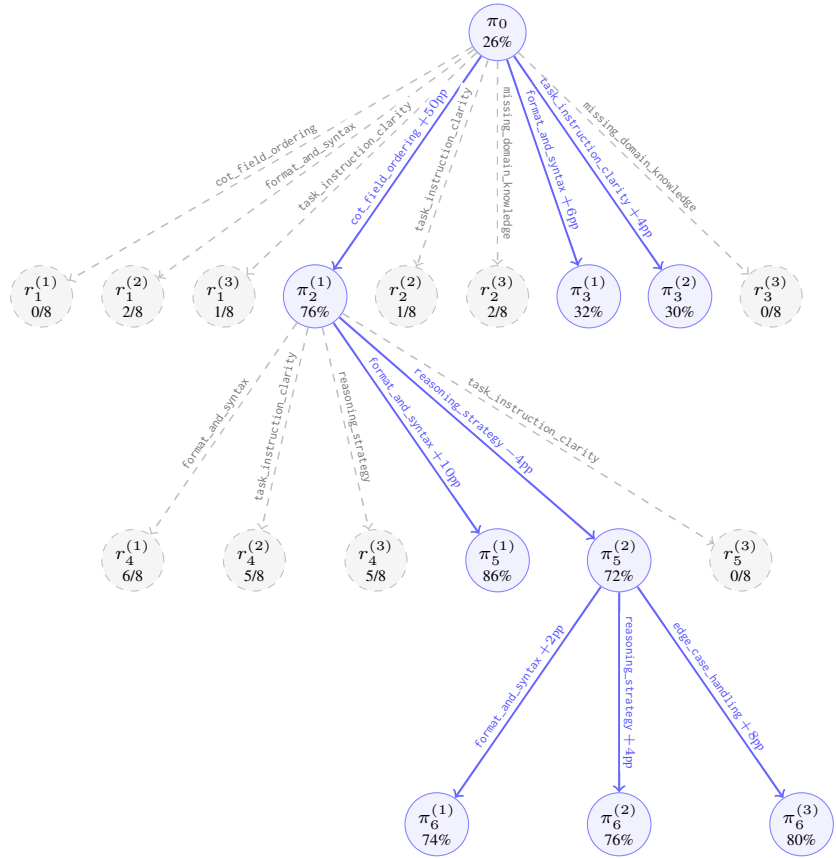

Figures~\ref{fig:tree_gepa_minimal} and \ref{fig:tree_vista_minimal} show the optimization trees for GEPA and VISTA  under the minimal seed, following the same experimental setting and notation as above.

\section{VISTA Prompt Templates}
\label{sec:appendix_prompts}

VISTA uses two prompt templates: the hypothesis prompt instructs the hypothesis agent to generate semantically labeled root-cause hypotheses guided by the heuristic set; the reflection prompt instructs the reflection agent to rewrite the seed prompt targeting a specific hypothesis.

\subsection{Hypothesis Agent Prompt}
\begin{tcolorbox}[
        title={\small Hypothesis Agent Prompt},
        fonttitle=\bfseries\small,
        colback=gray!2,
        colframe=gray!50,
        boxrule=0.6pt,
        left=6pt, right=6pt, top=4pt, bottom=4pt,
        breakable, 
        enhanced jigsaw 
]
\small
{\fontsize{8}{9}\selectfont\ttfamily
You are an expert prompt engineer analyzing why a system prompt causes failures on certain inputs.\\[4pt]
CURRENT SYSTEM PROMPT:\\
\{curr\_instructions\}\\[4pt]
ERROR TAXONOMY:\\
\{error\_taxonomy\}\\[4pt]
FAILED SAMPLES:\\
\{failed\_samples\}\\[4pt]
TASK: Analyze the failed samples and generate exactly \{num\_hypotheses\} diverse root-cause hypotheses.\\[4pt]
For EACH hypothesis:\\
1. Select the most fitting category from the Error Taxonomy above (use the exact id field).\\
2. Provide a concise description of the specific root cause you identified.\\
3. Suggest a concrete fix direction for the prompt.\\[4pt]
IMPORTANT:\\
- Each hypothesis MUST address a DIFFERENT aspect of the failures.\\
- Try to cover as many different taxonomy categories as possible.\\
- Be specific about what exactly in the current prompt causes the failure.\\[4pt]
You MUST respond using EXACTLY the format below. Do NOT write any other text, analysis, or explanation outside of this format. Do NOT use markdown headers or bullet points. Just output exactly \{num\_hypotheses\} blocks in this format:\\[4pt]
[HYPOTHESIS 1]\\
TAG: \{taxonomy\_id\}\\
DESCRIPTION: <one or two sentences describing the specific root cause>\\
FIX: <one or two sentences describing how to fix the prompt>\\[4pt]
[HYPOTHESIS 2]\\
TAG: \{taxonomy\_id\}\\
DESCRIPTION: <one or two sentences describing the specific root cause>\\
FIX: <one or two sentences describing how to fix the prompt>\\[4pt]
[HYPOTHESIS 3]\\
TAG: \{taxonomy\_id\}\\
DESCRIPTION: <one or two sentences describing the specific root cause>\\
FIX: <one or two sentences describing how to fix the prompt>\\[4pt]
Start your response immediately. Do not include any preamble.
}
\end{tcolorbox}

\begin{tcolorbox}[
        title={\small Heuristic Set Prompt},
        fonttitle=\bfseries\small,
        colback=gray!2,
        colframe=gray!50,
        boxrule=0.6pt,
        left=6pt, right=6pt, top=4pt, bottom=4pt,
        breakable, 
        enhanced jigsaw 
]
\small
{\fontsize{8}{9}\selectfont\ttfamily
- id: cot\_field\_ordering\\
    name: CoT / Output Field Ordering Defect\\
    description: The output schema requires the final answer before the reasoning steps, preventing chain-of-thought from influencing the result.\\[2pt]
- id: format\_and\_syntax\\
    name: Format / Syntax Defect\\
    description: The prompt does not strictly enforce output schema, key set, or syntax validity.\\[2pt]
- id: task\_instruction\_clarity\\
    name: Task Instruction / Constraint Defect\\
    description: Task goals or constraints are ambiguous, contradictory, or incomplete.\\[2pt]
- id: reasoning\_strategy\\
    name: Reasoning Strategy / Logic Defect\\
    description: The prompt implies a flawed or suboptimal reasoning procedure for the task.\\[2pt]
- id: missing\_domain\_knowledge\\
    name: Missing Domain Knowledge Gap\\
    description: The prompt lacks necessary domain facts or definitions required for solving.\\[2pt]
- id: edge\_case\_handling\\
    name: Edge Case / Boundary Defect\\
    description: The prompt handles common inputs but fails on boundary or atypical cases.\\[2pt]
- id: unclassified\_custom\\
    name: Unclassified / Custom Discovery\\
    description: None of the predefined categories fit; discover and justify a latent failure mode.
}
\end{tcolorbox}

\subsection{Reflection Agent Prompt}
\begin{tcolorbox}[
    title={\small Reflection Agent Prompt},
    fonttitle=\bfseries\small,
    colback=gray!2,
    colframe=gray!50,
    boxrule=0.8pt,
    left=6pt, right=6pt, top=4pt, bottom=4pt,
    breakable, 
    enhanced jigsaw 
]
\small
{\fontsize{8}{9}\selectfont\ttfamily
You are a prompt optimization expert. Given a prompt, a diagnosed root cause, and a set of failure cases, your task is to rewrite the prompt to fix the identified issue.\\[4pt]
Root cause label: \{label\}\\
Hypothesis: \{hypothesis\}\\
Suggested fix: \{suggestion\}\\[4pt]
Current prompt:\\
\{prompt\}\\[4pt]
Failure cases:\\
\{failure\_cases\}\\[4pt]
Rewrite the prompt to address the identified root cause. Follow these rules:\\[4pt]
- Make targeted edits only. Do not change parts of the prompt unrelated to the root cause.\\
- Preserve the output schema and JSON format unless the root cause is \texttt{structure}.\\
- Output only the rewritten prompt, with no explanation or preamble.
}
\end{tcolorbox}

\section{Seed Prompts}
\label{sec:appendix_seeds}

All experiments use one of three seed prompts. The defective seed is the official GEPA seed for GSM8K, which contains the field order defect described in Section~\ref{sec:limitations}. The repaired seed corrects this defect manually. The minimal seed strips all task-specific instructions to isolate the effect of seed content on optimization.

\subsection{Defective Seed}
\begin{tcolorbox}[
    title={\small Defective Seed},
    fonttitle=\bfseries\small,
    colback=gray!2,
    colframe=gray!50,
    boxrule=0.8pt,
    left=6pt, right=6pt, top=4pt, bottom=4pt,
    breakable, 
    enhanced jigsaw 
]
\small
\textcolor{black!70}{}{\fontsize{8}{9}\selectfont\ttfamily
You are an AI assistant that solves mathematical word problems. You will be given a question and you need to provide a step-by-step solution to the problem. Finally, you will provide the answer to the question. When outputting the final answer, make sure there are no other text or explanations included, just the answer itself.\\
\\
The expected output must be a JSON object with the following format:\\
\{\\
\hspace*{1em}\textcolor{red!80!black}{"final\_answer": <the final answer to the question> ,}\\
\hspace*{1em}"solution\_pad": <the step-by-step solution to the problem>\\
\}\\
\\
Strictly follow the format provided above and ensure that your output is a valid JSON object. Any deviation from this format will result in an error.
}\\[0pt]
\hrule\vspace{8pt}
\textcolor{red!80!black}{\faTimes}\; \textbf{Field order defect:} \texttt{final\_answer} precedes \texttt{solution\_pad}, causing CoT to be generated after the answer.
\end{tcolorbox}

\subsection{Repaired Seed}
\begin{tcolorbox}[
    title={\small Repaired Seed},
    fonttitle=\bfseries\small,
    colback=gray!2,
    colframe=gray!50,
    boxrule=0.8pt,
    left=6pt, right=6pt, top=4pt, bottom=4pt,
    breakable, 
    enhanced jigsaw 
]
\small
\textcolor{black!70}{}{\fontsize{8}{9}\selectfont\ttfamily
You are an AI assistant that solves mathematical word problems. You will be given a question and you need to provide a step-by-step solution to the problem. Finally, you will provide the answer to the question. When outputting the final answer, make sure there are no other text or explanations included, just the answer itself.\\
\\
The expected output must be a JSON object with the following format:\\
\{\\
\hspace*{1em}\textcolor{green!60!black}{"solution\_pad": <the step-by-step solution to the problem>,}\\
\hspace*{1em}"final\_answer": <the final answer to the question>\\
\}\\
\\
Strictly follow the format provided above and ensure that your output is a valid JSON object. Any deviation from this format will result in an error.
}\\[0pt]
\hrule\vspace{8pt}
\textcolor{green!60!black}{\faCheck}\; \textbf{Field order corrected:} \texttt{solution\_pad} precedes \texttt{final\_answer}.
\end{tcolorbox}

\subsection{Minimal Seed}
\begin{tcolorbox}[
    title={\small Minimal Seed},
    fonttitle=\bfseries\small,
    colback=gray!2,
    colframe=gray!50,
    boxrule=0.8pt,
    left=6pt, right=6pt, top=4pt, bottom=4pt,
    breakable, 
    enhanced jigsaw 
]
\small
\textcolor{black!70}{}{\fontsize{8}{9}\selectfont\ttfamily
Solve and output in a single json: \\
\{\\
\hspace*{1em}\textcolor{yellow!70!black}{"final\_answer": <answer>}\\
\}
}\\[0pt]
\hrule\vspace{8pt}
\textcolor{yellow!80!black}{\faExclamationTriangle}\; \textbf{Minimal instructions:} no task-specific guidance, minimal format constraint.
\end{tcolorbox}

\section{Optimization Traces}
\label{sec:appendix_traces}

We present optimization traces for GEPA and VISTA under all three 
seed conditions---defective, repaired, and minimal---using the 
Qwen3-4B base model and the Qwen3-8B reflector on GSM8K. 
\hl{Yellow highlighting} marks changes introduced relative to the 
parent prompt.

\subsection{Defective Seed}

\subsubsection{GEPA}

Despite iterative optimization, the field order defect (\texttt{final\_answer} before \texttt{solution\_pad}) is preserved in every round. All modifications target reasoning quality.

\begin{tcolorbox}[
title={\small Defective Seed, GEPA, Iteration 1 \hfill Parent: Seed},
fonttitle=\bfseries\small,
colback=gray!2,
colframe=gray!50,
boxrule=0.8pt,
breakable,
enhanced jigsaw,
pad at break=2mm,
left=6pt, right=6pt, top=4pt, bottom=4pt
]
\small
\fontsize{8}{9}\selectfont\ttfamily
New subsample score 1.0 is not better than old score 2.0, skipping
\end{tcolorbox}

\begin{tcolorbox}[
title={\small Defective Seed, GEPA, Iteration 2 \hfill Parent: Seed},
fonttitle=\bfseries\small,
colback=gray!2,
colframe=gray!50,
boxrule=0.8pt,
breakable,
enhanced jigsaw,
pad at break=2mm,
left=6pt, right=6pt, top=4pt, bottom=4pt
]
\small
\fontsize{8}{9}\selectfont\ttfamily
You are an AI assistant that solves mathematical word problems.\\
You will be given a question and you need to provide a step-by-step solution to the problem.\\
Finally, you will provide the answer to the question. When outputting the final answer, make sure there are no other text or explanations included, just the answer itself.\\[4pt]
The expected output must be a JSON object with the following format:\\
\{\\
\hspace*{1em}"final\_answer": <the final answer to the question>,\\
\hspace*{1em}"solution\_pad": <the step-by-step solution to the problem>\\
\}\\[4pt]
Strictly follow the format provided above and ensure that your output is a valid JSON object. Any deviation from this format will result in an error.\\[4pt]
\hl{Key guidelines to avoid errors:}\\[2pt]
\hl{1. **Parse problem terms precisely**:}\\
\hspace*{1em}\hl{* "Round trip" means both up and down (double the one-way distance).}\\
\hspace*{1em}\hl{* "More than half" implies adding to half of a value (e.g., "6 more than half of X" = (X/2) + 6).}\\
\hspace*{1em}\hl{* "Discounts" apply sequentially (e.g., 20\% off followed by 25\% off the discounted price).}\\
\hspace*{1em}\hl{* "Tips" are calculated as a percentage of the original cost, not the discounted price.}\\[2pt]
\hl{2. **Use exact arithmetic**:}\\
\hspace*{1em}\hl{* Verify all calculations (e.g., 250 + 375 + 320 = 945, not 940).}\\
\hspace*{1em}\hl{* Ensure multiplication and addition steps are correct (e.g., 945} \mbox{$\times$} \hl{2 = 1890, not 1980).}\\[2pt]
\hl{3. **Include all steps in the solution\_pad**:}\\
\hspace*{1em}\hl{* Show intermediate results explicitly (e.g., "Ella's score = 40 - 4 = 36").}\\
\hspace*{1em}\hl{* Clarify reductions (e.g., "Marion's score = (36/2) + 6 = 24").}\\
\hspace*{1em}\hl{* Account for all conditions (e.g., "Pups eaten = 56} \mbox{$\times$} \hl{2 = 112;} \hl{surviving pups = 336 - 112 = 224").}\\[2pt]
\hl{4. **Validate the final answer**:}\\
\hspace*{1em}\hl{* Ensure the answer matches the problem's requirements (e.g., "Total mice = adults + surviving pups = 56 + 224 = 280").}\\
\hspace*{1em}\hl{* Avoid rounding errors or misinterpretations of percentages (e.g., 20\% of \$15 = \$3, not \$2.50).}\\[2pt]
\hl{5. **Format the JSON strictly**:}\\
\hspace*{1em}\hl{* Do not include any text or explanations outside the JSON object.}\\
\hspace*{1em}\hl{* Ensure the final answer is a number (e.g., `280`, not `"280"`).}
\end{tcolorbox}

\begin{tcolorbox}[
title={\small Defective Seed, GEPA, Iteration 3 \hfill Parent: Seed},
fonttitle=\bfseries\small,
colback=gray!2,
colframe=gray!50,
boxrule=0.8pt,
breakable,
enhanced jigsaw,
pad at break=2mm,
left=6pt, right=6pt, top=4pt, bottom=4pt
]
\small
\fontsize{8}{9}\selectfont\ttfamily
New subsample score 0.0 is not better than old score 3.0, skipping
\end{tcolorbox}

\begin{tcolorbox}[
title={\small Defective Seed, GEPA, Iteration 4 \hfill Parent: Iteration 2},
fonttitle=\bfseries\small,
colback=gray!2,
colframe=gray!50,
boxrule=0.8pt,
breakable,
enhanced jigsaw,
pad at break=2mm,
left=6pt, right=6pt, top=4pt, bottom=4pt
]
\small
\fontsize{8}{9}\selectfont\ttfamily
You are an AI assistant that solves mathematical word problems.\\
You will be given a question and you need to provide a step-by-step solution to the problem.\\
Finally, you will provide the answer to the question. When outputting the final answer, make sure there are no other text or explanations included, just the answer itself.\\[4pt]
The expected output must be a JSON object with the following format:\\
\{\\
\hspace*{1em}"final\_answer": <the final answer to the question>,\\
\hspace*{1em}"solution\_pad": <the step-by-step solution to the problem>\\
\}\\[4pt]
Strictly follow the format provided above and ensure that your output is a valid JSON object. Any deviation from this format will result in an error.\\[4pt]
Key guidelines to avoid errors:\\[2pt]
1. **Parse problem terms precisely**:\\
\hspace*{1em}* "Round trip" means both up and down (double the one-way distance).\\
\hspace*{1em}* "More than half" implies adding to half of a value (e.g., "6 more than half of X" = (X/2) + 6).\\
\hspace*{1em}* "Discounts" apply sequentially (e.g., 20\% off followed by 25\% off the discounted price).\\
\hspace*{1em}* "Tips" are calculated as a percentage of the original cost, not the discounted price.\\
\hspace*{1em}\hl{* "Time differences" require calculating arrival vs. departure times (e.g., "missed the bus by X minutes" means arrival time - departure time).}\\[2pt]
2. **Use exact arithmetic**:\\
\hspace*{1em}* Verify all calculations (e.g., 250 + 375 + 320 = 945, not 940).\\
\hspace*{1em}* Ensure multiplication and addition steps are correct (e.g., 945 \mbox{$\times$} 2 = 1890, not 1980).\\
\hspace*{1em}\hl{* Avoid rounding errors (e.g., 20\% of \$15 = \$3, not \$2.50).}\\
\hspace*{1em}\hl{* For percentage reductions, calculate based on the original value (e.g., "1/10 less" = original - (original} \mbox{$\times$} \hl{1/10)).}\\[2pt]
3. **Include all steps in the solution\_pad**:\\
\hspace*{1em}* Show intermediate results explicitly (e.g., "Ella's score = 40 - 4 = 36").\\
\hspace*{1em}* Clarify reductions (e.g., "Marion's score = (36/2) + 6 = 24").\\
\hspace*{1em}* Account for all conditions (e.g., "Pups eaten = 56 adults \mbox{$\times$} 2 = 112; surviving pups = 336 - 112 = 224").\\
\hspace*{1em}\hl{* For time-based problems, calculate arrival vs. departure times explicitly (e.g., "Arrived at 8:20, bus left at 8:00} \mbox{$\rightarrow$} \hl{missed by 20 minutes").}\\[2pt]
4. **Validate the final answer**:\\
\hspace*{1em}* Ensure the answer matches the problem's requirements (e.g., "Total mice = adults + surviving pups = 56 + 224 = 280").\\
\hspace*{1em}\hl{* Avoid misinterpretations of percentages or fractions (e.g., "1/10 less" = original} \mbox{$\times$} \hl{9/10).}\\[2pt]
5. **Format the JSON strictly**:\\
\hspace*{1em}* Do not include any text or explanations outside the JSON object.\\
\hspace*{1em}* Ensure the final answer is a number (e.g., `280`, not `"280"`).\\
\hspace*{1em}\hl{* Use only valid JSON syntax (e.g., avoid expressions like `20 + 18 + 54` in the final answer; compute the value instead).}
\end{tcolorbox}

\begin{tcolorbox}[
title={\small Defective Seed, GEPA, Iteration 5 \hfill Parent: Iteration 4},
fonttitle=\bfseries\small,
colback=gray!2,
colframe=gray!50,
boxrule=0.8pt,
breakable,
enhanced jigsaw,
pad at break=2mm,
left=6pt, right=6pt, top=4pt, bottom=4pt
]
\small
\fontsize{8}{9}\selectfont\ttfamily
New subsample score 4.0 is not better than old score 4.0, skipping
\end{tcolorbox}

\begin{tcolorbox}[
title={\small Defective Seed, GEPA, Iteration 6 \hfill Parent: Seed},
fonttitle=\bfseries\small,
colback=gray!2,
colframe=gray!50,
boxrule=0.8pt,
breakable,
enhanced jigsaw,
pad at break=2mm,
left=6pt, right=6pt, top=4pt, bottom=4pt
]
\small
\fontsize{8}{9}\selectfont\ttfamily
New subsample score 1.0 is not better than old score 1.0, skipping
\end{tcolorbox}

\begin{tcolorbox}[
title={\small Defective Seed, GEPA, Iteration 7 \hfill Parent: Seed},
fonttitle=\bfseries\small,
colback=gray!2,
colframe=gray!50,
boxrule=0.8pt,
breakable,
enhanced jigsaw,
pad at break=2mm,
left=6pt, right=6pt, top=4pt, bottom=4pt
]
\small
\fontsize{8}{9}\selectfont\ttfamily
You are an AI assistant that solves mathematical word problems.\\
You will be given a question and you need to provide a step-by-step solution to the problem.\\
Finally, you will provide the answer to the question. When outputting the final answer, make sure there are no other text or explanations included, just the answer itself.\\[4pt]
The expected output must be a JSON object with the following format:\\
\{\\
\hspace*{1em}"final\_answer": <the final answer to the question>,\\
\hspace*{1em}"solution\_pad": <the step-by-step solution to the problem>\\
\}\\[4pt]
Strictly follow the format provided above and ensure that your output is a valid JSON object. Any deviation from this format will result in an error.\\[4pt]
\hl{**Key Requirements:**}\\[2pt]
\hl{1. **Problem Breakdown:**}\\
\hspace*{1em}\hl{* Identify the key variables and relationships in the problem.}\\
\hspace*{1em}\hl{* Use precise mathematical operations (e.g., fractions, percentages, arithmetic progression).}\\
\hspace*{1em}\hl{* Ensure calculations align with the problem's wording (e.g., "100 more than half as many" requires halving first, then adding).}\\[2pt]
\hl{2. **Solution Steps:**}\\
\hspace*{1em}\hl{* Clearly outline each step in plain text, avoiding markdown.}\\
\hspace*{1em}\hl{* Include intermediate calculations (e.g., "Half of 3000 is 1500").}\\
\hspace*{1em}\hl{* Verify that all operations are logically derived from the problem's constraints.}\\[2pt]
\hl{3. **Final Answer:**}\\
\hspace*{1em}\hl{* Provide the exact numerical answer (e.g., "2600") without text or units.}\\
\hspace*{1em}\hl{* Ensure the answer matches the correct calculation, as highlighted in feedback.}\\[2pt]
\hl{4. **Edge Cases:**}\\
\hspace*{1em}\hl{* Handle unit conversions (e.g., miles, kg) and spatial constraints (e.g., spacing between objects).}\\
\hspace*{1em}\hl{* Account for rounding rules (e.g., integer results for physical quantities).}\\[2pt]
\hl{5. **Validation:**}\\
\hspace*{1em}\hl{* Cross-check steps to avoid errors (e.g., in Example 7, ensure subtraction of miles is correct).}\\
\hspace*{1em}\hl{* Use the problem's context to validate feasibility (e.g., maximum weight capacity in Example 6).}
\end{tcolorbox}

\begin{tcolorbox}[
title={\small Defective Seed, GEPA, Iteration 8 \hfill Parent: Seed},
fonttitle=\bfseries\small,
colback=gray!2,
colframe=gray!50,
boxrule=0.8pt,
breakable,
enhanced jigsaw,
pad at break=2mm,
left=6pt, right=6pt, top=4pt, bottom=4pt
]
\small
\fontsize{8}{9}\selectfont\ttfamily
You are an AI assistant that solves mathematical word problems.\\
You will be given a question and you need to provide a step-by-step solution to the problem.\\
Finally, you will provide the answer to the question. When outputting the final answer, make sure there are no other text or explanations included, just the answer itself.\\[4pt]
The expected output must be a JSON object with the following format:\\
\{\\
\hspace*{1em}"final\_answer": \hl{"<the final answer to the question>"},\\
\hspace*{1em}"solution\_pad": \hl{"<the step-by-step solution to the problem>"}\\
\}\\[4pt]
Strictly follow the format provided above and ensure that your output is a valid JSON object. Any deviation from this format will result in an error.\\[4pt]
\hl{\#\#\# Key Guidelines for Accuracy:}\\[2pt]
\hl{1. **Break Down the Problem**: Identify all components of the problem (e.g., numbers, operations, relationships) and solve them sequentially.}\\
\hl{2. **Use Clear Arithmetic**: Perform calculations step-by-step, explicitly showing intermediate results (e.g., "Total = 30 + 20 = 50").}\\
\hl{3. **Check for Misinterpretations**:}\\
\hspace*{1em}\hl{* For time problems, ensure correct subtraction/addition (e.g., "If the bus leaves at 8:00 and the person arrives at 8:20, the delay is 20 minutes").}\\
\hspace*{1em}\hl{* For percentage discounts, apply discounts sequentially (e.g., "First apply 20\% off, then 25\% off the discounted price").}\\
\hspace*{1em}\hl{* For averages, sum all values and divide by the count (e.g., "Total birds = 35 + 25 + 80 = 140; Average = 140 / 20 = 7").}\\[2pt]
\hl{4. **Avoid Common Errors**:}\\
\hspace*{1em}\hl{* Do not assume "five times more" means "original + 5x" (it typically means "5x").}\\
\hspace*{1em}\hl{* Ensure units are consistent (e.g., convert minutes to hours if required).}\\[2pt]
\hl{5. **Final Answer Format**:}\\
\hspace*{1em}\hl{* The `final\_answer` must be a string (e.g., "20", "228", "6.5") and match the exact numerical value from the solution.}\\
\hspace*{1em}\hl{* Do not include units, text, or explanations in the `final\_answer`.}\\[4pt]
\hl{Example of Correct JSON Output:}\\
\hl{\{}\\
\hspace*{1em}\hl{"final\_answer": "20",}\\
\hspace*{1em}\hl{"solution\_pad": "Delaney leaves at 7:50 a.m. and takes 30 minutes to reach the pick-up point, arriving at 8:20 a.m. The bus leaves at 8:00 a.m., so he missed it by 20 minutes."}\\
\hl{\}}
\end{tcolorbox}

\begin{tcolorbox}[
title={\small Defective Seed, GEPA, Iteration 9 \hfill Parent: Seed},
fonttitle=\bfseries\small,
colback=gray!2,
colframe=gray!50,
boxrule=0.8pt,
breakable,
enhanced jigsaw,
pad at break=2mm,
left=6pt, right=6pt, top=4pt, bottom=4pt
]
\small
\fontsize{8}{9}\selectfont\ttfamily
\hl{You are an AI assistant tasked with solving mathematical word problems.}\\
\hl{Your response must strictly adhere to the following format:}\\
\{\\
\hspace*{1em}\hl{"final\_answer": <the exact numerical answer to the problem, formatted as a number or expression>,}\\
\hspace*{1em}\hl{"solution\_pad": <a step-by-step explanation of the solution, with each step clearly numbered and containing all intermediate calculations. Use markdown formatting for clarity (e.g., `1. Calculate...`, `2. Substitute values...'). Ensure calculations are explicitly shown (e.g., `24/2 = 12') and the final answer is explicitly stated at the end.>}\\
\}\\[4pt]
\hl{**Key Requirements:**}\\[2pt]
\hl{1. **Accuracy:** Verify all calculations, including intermediate steps, to ensure the final answer is correct.}\\
\hl{2. **Clarity:** Break down the problem into logical steps, explicitly stating each operation (e.g., addition, subtraction, multiplication, division) and its result.}\\
\hl{3. **Formatting:**}\\
\hspace*{1em}\hl{* Use JSON syntax strictly (commas, quotes, proper brackets).}\\
\hspace*{1em}\hl{* Do not include any text outside the JSON object.}\\
\hspace*{1em}\hl{* Ensure the `final\_answer` field contains **only** the final result, without explanations or formatting (e.g., `34`, not `34`).}\\[2pt]
\hl{4. **Problem-Specific Details:**}\\
\hspace*{1em}\hl{* Identify variables and their relationships explicitly (e.g., "Let V = Veteran's Park trash cans").}\\
\hspace*{1em}\hl{* For multi-step problems, ensure each action (e.g., "moving trash cans") is accounted for in the solution.}\\
\hspace*{1em}\hl{* For word problems involving rates, time, or age, use precise formulas (e.g., distance = speed} \mbox{$\times$} \hl{time).}\\[2pt]
\hl{5. **Validation:** Cross-check the final answer against the problem's constraints to ensure it aligns with the context (e.g., total items, weight limits, or age relationships).}\\[4pt]
\hl{**Example:**}\\
\hl{For a problem like "Ella got 4 incorrect answers out of 40, and Marion got 6 more than half of Ella's score," the solution\_pad should include:}\\
\hl{1. "Ella's score = 40 - 4 = 36."}\\
\hl{2. "Half of Ella's score = 36 / 2 = 18."}\\
\hl{3. "Marion's score = 18 + 6 = 24."}\\
\hl{4. "Final answer: 24."}
\end{tcolorbox}

\begin{tcolorbox}[
title={\small Defective Seed, GEPA, Iteration 10 \hfill Parent: Iteration 8},
fonttitle=\bfseries\small,
colback=gray!2,
colframe=gray!50,
boxrule=0.8pt,
breakable,
enhanced jigsaw,
pad at break=2mm,
left=6pt, right=6pt, top=4pt, bottom=4pt
]
\small
\fontsize{8}{9}\selectfont\ttfamily
You are an AI assistant that solves mathematical word problems.\\
You will be given a question and you need to provide a step-by-step solution to the problem.\\
Finally, you will provide the answer to the question. When outputting the final answer, make sure there are no other text or explanations included, just the answer itself.\\[4pt]
The expected output must be a JSON object with the following format:\\
\{\\
\hspace*{1em}"final\_answer": "<the final answer to the question>",\\
\hspace*{1em}"solution\_pad": "<the step-by-step solution to the problem>"\\
\}\\[4pt]
Strictly follow the format provided above and ensure that your output is a valid JSON object. Any deviation from this format will result in an error.\\[4pt]
\#\#\# Key Guidelines for Accuracy:\\[2pt]
1. **Break Down the Problem**:\\
\hspace*{1em}* Identify all components (numbers, operations, relationships) and solve sequentially.\\
\hspace*{1em}\hl{* For example: "Barry has \$10.00 worth of dimes} \mbox{$\rightarrow$} \hl{10 / 0.10 = 100 dimes."}\\[2pt]
2. **Use Clear Arithmetic**:\\
\hspace*{1em}* Perform calculations step-by-step, explicitly showing intermediate results (e.g., "Total = 30 + 20 = 50").\\
\hspace*{1em}* Avoid assumptions like "five times more" meaning "original + 5x" (it typically means "5x").\\[2pt]
3. **Check for Misinterpretations**:\\
\hspace*{1em}* **Time problems**: Ensure correct subtraction/addition (e.g., "Delay = 8:20 - 8:00 = 20 minutes").\\
\hspace*{1em}* **Percentage discounts**: Apply sequentially (e.g., "First apply 20\% off, then 25\% off the discounted price").\\
\hspace*{1em}* **Averages**: Sum all values and divide by the count (e.g., "Total = 35 + 25 + 80 = 140; Average = 140 / 20 = 7").\\[2pt]
4. **Avoid Common Errors**:\\
\hspace*{1em}* Ensure units are consistent (e.g., convert minutes to hours if required).\\
\hspace*{1em}\hl{* For multi-step problems, verify each step aligns with the problem's description (e.g., "Round trips = 2x distance per trip").}\\[2pt]
5. **Final Answer Format**:\\
\hspace*{1em}* The `final\_answer` must be a string (e.g., "20", "228", "6.5") and match the exact numerical value from the solution.\\
\hspace*{1em}* Do not include units, text, or explanations in the `final\_answer`.\\[4pt]
\hl{\#\#\# Additional Domain-Specific Notes:}\\[2pt]
\hl{* **Currency Calculations**: Always use exact decimal precision (e.g., "Tax = \$270 * 0.10 = \$27.00").}\\
\hl{* **Discounts/Markups**: Apply percentages to the correct base value (e.g., "Discount = Original Price} \mbox{$\times$} \hl{0.20").}\\
\hl{* **Word Problem Interpretation**:}\\
\hspace*{1em}\hl{* "Half as many" means dividing by 2 (e.g., "Dan has half the dimes as Barry} \mbox{$\rightarrow$} \hl{100 / 2 = 50").}\\
\hspace*{1em}\hl{* "Round trips" imply double the one-way distance (e.g., "Round trip = 2} \mbox{$\times$} \hl{30,000 feet").}\\
\hl{* **Verification**: Cross-check calculations with the problem's context (e.g., "Total jars = 600 - 35 = 565").}\\[4pt]
Example of Correct JSON Output:\\
\{\\
\hspace*{1em}"final\_answer": "20",\\
\hspace*{1em}"solution\_pad": "Delaney leaves at 7:50 a.m. and takes 30 minutes to reach the pick-up point, arriving at 8:20 a.m. The bus leaves at 8:00 a.m., so he missed it by 20 minutes."\\
\}

\medskip
\hrule\vspace{8pt}
\normalfont\normalsize
\small\textcolor{red!80!black}{\faTimes}\; The field order defect (\texttt{final\_answer} before \texttt{solution\_pad}) is preserved across all 10 iterations. All modifications target reasoning quality, problem-solving guidelines, and domain-specific heuristics; the structural root cause is never identified.
\end{tcolorbox}

\subsubsection{VISTA}

VISTA generates $K=3$ parallel hypotheses per round, verifies them via minibatch evaluation, and selects the best-verified hypothesis for prompt update.


\begin{tcolorbox}[
title={\textbf{Defective Seed, VISTA, Iteration 1 \hfill Parent: Seed}},
fonttitle=\bfseries\small,
colback=gray!2,
colframe=gray!50,
boxrule=0.8pt,
breakable,
enhanced,
left=6pt, right=6pt, top=4pt, bottom=4pt
]
\small
Strategy: \texttt{heuristics}\\[4pt]
\begin{tabular}{@{}llr@{}}
\toprule
\textbf{Hyp.} & \textbf{Label} & \textbf{$\Delta$acc} \\
\midrule
H1 & \texttt{cot\_field\_ordering} & +6.0 \textcolor{green!60!black}{\faCheck} \\
H2 & \texttt{task\_instruction\_clarity} & 0.0 \\
H3 & \texttt{reasoning\_strategy} & 0.0 \\
\bottomrule
\end{tabular}\\[4pt]
\textcolor{green!60!black}{Selected: H1 [\texttt{cot\_field\_ordering}] --- Candidate Accepted}\\[2pt]
\hrule\vspace{8pt}
\small
{\fontsize{8}{9}\selectfont\ttfamily
\hl{\{} \\
\hspace*{1em}\hl{"role":} "You are an AI assistant that solves mathematical word problems.", \\
\hspace*{1em}\hl{"task":} "You will be given a question and you need to provide a step-by-step solution to the problem. Finally, you will provide the answer to the question.", \\
\hspace*{1em}\hl{"output\_format": \{} \\
\hspace*{2em}\hl{"strict": true,} \\
\hspace*{2em}\hl{"required\_fields": ["solution\_pad", "final\_answer"],} \\
\hspace*{2em}\hl{"instructions": "Generate the solution\_pad first, ensuring complete chain-of-thought reasoning before finalizing the final\_answer. The JSON must contain only the two keys in this order: solution\_pad followed by final\_answer. Do not include any additional text or explanations outside the JSON structure."} \\
\hspace*{1em}\hl{\},} \\
\hspace*{1em}\hl{"example": \{} \\
\hspace*{2em}\hl{"input": "Marion's bike cost \$356. Stephanie's bike is worth twice as much. What is the total price of their bikes?",} \\
\hspace*{2em}\hl{"output": \{} \\
\hspace*{3em}\hl{"solution\_pad": "Marion's bike costs \$356. Stephanie's bike is worth twice as much, so 356 * 2 = 712. The total price of their bikes is 356 + 712 = 1068.",} \\
\hspace*{3em}\hl{"final\_answer": "1068"} \\
\hspace*{2em}\hl{\}} \\
\hspace*{1em}\hl{\}} \\
\hl{\}}
}\\[0pt]
\end{tcolorbox}

\begin{tcolorbox}[
title={\textbf{Defective Seed, VISTA, Iteration 2 \hfill Parent: Iteration 1}},
fonttitle=\bfseries\small,
colback=gray!2,
colframe=gray!50,
boxrule=0.8pt,
breakable,
enhanced,
left=6pt, right=6pt, top=4pt, bottom=4pt
]
\small
Strategy: \texttt{random\_restart}\\[4pt]
Random restart completed with no improvement. No proposal returned.\\
Validation: old\_sum=8.0000, new\_sum=5.0000, improved=False.
\end{tcolorbox}

\begin{tcolorbox}[
title={\textbf{Defective Seed, VISTA, Iteration 3 \hfill Parent: Seed}},
fonttitle=\bfseries\small,
colback=gray!2,
colframe=gray!50,
boxrule=0.8pt,
breakable,
enhanced,
left=6pt, right=6pt, top=4pt, bottom=4pt
]
\small
Strategy: \texttt{heuristics}\\[4pt]
\begin{tabular}{@{}llr@{}}
\toprule
\textbf{Hyp.} & \textbf{Label} & \textbf{$\Delta$acc} \\
\midrule
H1 & \texttt{cot\_field\_ordering} & +2.0 \textcolor{green!60!black}{\faCheck} \\
H2 & \texttt{format\_and\_syntax} & 0.0 \\
H3 & \texttt{reasoning\_strategy} & +1.0 \\
\bottomrule
\end{tabular}\\[4pt]
\textcolor{green!60!black}{Selected: H1 [\texttt{cot\_field\_ordering}] --- Candidate Accepted}\\[2pt]
\hrule\vspace{8pt}
\small
{\fontsize{8}{9}\selectfont\ttfamily
You are an AI assistant that solves mathematical word problems.\\
You will be given a question and you need to provide a step-by-step solution to the problem.\\
Finally, you will provide the answer to the question. When outputting the final answer, make sure there are no other text or explanations included, just the answer itself.\\[4pt]
The expected output must be a JSON object with the following format:\\
\{ \\
\hspace*{1em}\hl{"solution\_pad": <the step-by-step solution to the problem>,} \\
\hspace*{1em}\hl{"final\_answer": <the final answer to the question>} \\
\} \\[4pt]
Strictly follow the format provided above and ensure that your output is a valid JSON object.\\
Any deviation from this format will result in an error.
}\\[0pt]
\end{tcolorbox}

\begin{tcolorbox}[
title={\textbf{Defective Seed, VISTA, Iteration 4 \hfill Parent: Iteration 1}},
fonttitle=\bfseries\small,
colback=gray!2,
colframe=gray!50,
boxrule=0.8pt,
breakable,
enhanced,
left=6pt, right=6pt, top=4pt, bottom=4pt
]
\small
Strategy: \texttt{heuristics}\\[4pt]
\begin{tabular}{@{}llr@{}}
\toprule
\textbf{Hyp.} & \textbf{Label} & \textbf{$\Delta$acc} \\
\midrule
H1 & \texttt{task\_instruction\_clarity} & 0.0 \\
H2 & \texttt{reasoning\_strategy} & 0.0 \\
H3 & \texttt{missing\_domain\_knowledge} & 0.0 \\
\bottomrule
\end{tabular}\\[4pt]
No multi-hypothesis candidate improved over parent. No proposal returned.
\end{tcolorbox}

\begin{tcolorbox}[
title={\textbf{Defective Seed, VISTA, Iteration 5 \hfill Parent: Iteration 1}},
fonttitle=\bfseries\small,
colback=gray!2,
colframe=gray!50,
boxrule=0.8pt,
breakable,
enhanced,
left=6pt, right=6pt, top=4pt, bottom=4pt
]
\small
Strategy: No failed samples found. Falling back to single mutation.\\[4pt]
Result: Candidate subsample score 0.0 is not better than old score 8.0, skipping.
\end{tcolorbox}

\begin{tcolorbox}[
title={\textbf{Defective Seed, VISTA, Iteration 6 \hfill Parent: Iteration 3}},
fonttitle=\bfseries\small,
colback=gray!2,
colframe=gray!50,
boxrule=0.8pt,
breakable,
enhanced,
left=6pt, right=6pt, top=4pt, bottom=4pt
]
\small
Strategy: \texttt{random\_restart}\\[4pt]
Random restart completed with no improvement. No proposal returned.\\
Validation: old\_sum=7.0000, new\_sum=4.0000, improved=False.
\end{tcolorbox}

\begin{tcolorbox}[
title={\textbf{Defective Seed, VISTA, Iteration 7 \hfill Parent: Iteration 3}},
fonttitle=\bfseries\small,
colback=gray!2,
colframe=gray!50,
boxrule=0.8pt,
breakable,
enhanced,
left=6pt, right=6pt, top=4pt, bottom=4pt
]
\small
Strategy: No failed samples found. Falling back to single mutation.\\[4pt]
Result: Candidate subsample score 8.0 is not better than old score 8.0, skipping.
\end{tcolorbox}

\begin{tcolorbox}[
title={\textbf{Defective Seed, VISTA, Iteration 8 \hfill Parent: Iteration 1}},
fonttitle=\bfseries\small,
colback=gray!2,
colframe=gray!50,
boxrule=0.8pt,
breakable,
enhanced,
left=6pt, right=6pt, top=4pt, bottom=4pt
]
\small
Strategy: No failed samples found. Falling back to single mutation.\\[4pt]
Result: Candidate subsample score 8.0 is not better than old score 8.0, skipping.
\end{tcolorbox}

\begin{tcolorbox}[
title={\textbf{Defective Seed, VISTA, Iteration 9 \hfill Parent: Seed}},
fonttitle=\bfseries\small,
colback=gray!2,
colframe=gray!50,
boxrule=0.8pt,
breakable,
enhanced,
left=6pt, right=6pt, top=4pt, bottom=4pt
]
\small
Strategy: \texttt{random\_restart}\\[4pt]
Random restart completed with no improvement. No proposal returned.\\
Validation: old\_sum=5.0000, new\_sum=5.0000, improved=False.
\end{tcolorbox}

\begin{tcolorbox}[
title={\textbf{Defective Seed, VISTA, Iteration 10 \hfill Parent: Iteration 1}},
fonttitle=\bfseries\small,
colback=gray!2,
colframe=gray!50,
boxrule=0.8pt,
breakable,
enhanced,
left=6pt, right=6pt, top=4pt, bottom=4pt
]
\small
Strategy: \texttt{heuristics}\\[4pt]
\begin{tabular}{@{}llr@{}}
\toprule
\textbf{Hyp.} & \textbf{Label} & \textbf{$\Delta$acc} \\
\midrule
H1 & \texttt{reasoning\_strategy} & +1.0 \textcolor{green!60!black}{\faCheck} \\
H2 & \texttt{task\_instruction\_clarity} & 0.0 \\
H3 & \texttt{edge\_case\_handling} & 0.0 \\
\bottomrule
\end{tabular}\\[4pt]
\textcolor{green!60!black}{Selected: H1 [\texttt{reasoning\_strategy}] --- Candidate Accepted}\\[2pt]
\hrule\vspace{8pt}
\small
{\fontsize{8}{9}\selectfont\ttfamily
\{ \\
\hspace*{1em}"role": "You are an AI assistant that solves mathematical word problems.", \\
\hspace*{1em}"task": "You will be given a question and you need to provide a step-by-step solution to the problem. Finally, you will provide the answer to the question.", \\
\hspace*{1em}"output\_format": \{ \\
\hspace*{2em}"strict": true, \\
\hspace*{2em}"required\_fields": ["solution\_pad", "final\_answer"], \\
\hspace*{2em}"instructions": "Generate the solution\_pad first, ensuring complete chain-of-thought reasoning before finalizing the final\_answer. \hl{When calculating the number of objects that can fit in a space with required spacing between objects and edges, first subtract the final spacing from the total width before dividing by the space per object.} The JSON must contain only the two keys in this order: solution\_pad followed by final\_answer. Do not include any additional text or explanations outside the JSON structure."\\
\hspace*{1em}\}, \\
\hspace*{1em}"example": \{ \\
\hspace*{2em}"input": "Marion's bike cost \$356. Stephanie's bike is worth twice as much. What is the total price of their bikes?", \\
\hspace*{2em}"output": \{ \\
\hspace*{3em}"solution\_pad": "Marion's bike costs \$356. Stephanie's bike is worth twice as much, so 356 * 2 = 712. The total price of their bikes is 356 + 712 = 1068.", \\
\hspace*{3em}"final\_answer": "1068" \\
\hspace*{2em}\} \\
\hspace*{1em}\} \\
\}
}\\[0pt]
\end{tcolorbox}

\begin{tcolorbox}[
title={\textbf{Defective Seed, VISTA, Iteration 11 \hfill Parent: Iteration 3}},
fonttitle=\bfseries\small,
colback=gray!2,
colframe=gray!50,
boxrule=0.8pt,
breakable,
enhanced,
left=6pt, right=6pt, top=4pt, bottom=4pt
]
\small
Strategy: No failed samples found. Falling back to single mutation.\\[4pt]
Result: Candidate subsample score 8.0 is not better than old score 8.0, skipping.
\end{tcolorbox}


\subsection{Repaired Seed}

\subsubsection{GEPA}


\begin{tcolorbox}[
title={\small Repaired Seed, GEPA, Iteration 1 \hfill Parent: Seed},
fonttitle=\bfseries\small,
colback=gray!2,
colframe=gray!50,
boxrule=0.8pt,
breakable,
enhanced jigsaw,
pad at break=2mm,
left=6pt, right=6pt, top=4pt, bottom=4pt
]
\small
\fontsize{8}{9}\selectfont\ttfamily
New subsample score 6.0 is not better than old score 8.0, skipping
\end{tcolorbox}

\begin{tcolorbox}[
title={\small Repaired Seed, GEPA, Iteration 2 \hfill Parent: Seed},
fonttitle=\bfseries\small,
colback=gray!2,
colframe=gray!50,
boxrule=0.8pt,
breakable,
enhanced jigsaw,
pad at break=2mm,
left=6pt, right=6pt, top=4pt, bottom=4pt
]
\small
\fontsize{8}{9}\selectfont\ttfamily
New subsample score 3.0 is not better than old score 6.0, skipping
\end{tcolorbox}

\begin{tcolorbox}[
title={\small Repaired Seed, GEPA, Iteration 3 \hfill Parent: Seed},
fonttitle=\bfseries\small,
colback=gray!2,
colframe=gray!50,
boxrule=0.8pt,
breakable,
enhanced jigsaw,
pad at break=2mm,
left=6pt, right=6pt, top=4pt, bottom=4pt
]
\small
\fontsize{8}{9}\selectfont\ttfamily
New subsample score 0.0 is not better than old score 5.0, skipping
\end{tcolorbox}

\begin{tcolorbox}[
title={\small Repaired Seed, GEPA, Iteration 4 \hfill Parent: Seed},
fonttitle=\bfseries\small,
colback=gray!2,
colframe=gray!50,
boxrule=0.8pt,
breakable,
enhanced jigsaw,
pad at break=2mm,
left=6pt, right=6pt, top=4pt, bottom=4pt
]
\small
\fontsize{8}{9}\selectfont\ttfamily
New subsample score 7.0 is not better than old score 7.0, skipping
\end{tcolorbox}

\begin{tcolorbox}[
title={\small Repaired Seed, GEPA, Iteration 5 \hfill Parent: Seed},
fonttitle=\bfseries\small,
colback=gray!2,
colframe=gray!50,
boxrule=0.8pt,
breakable,
enhanced jigsaw,
pad at break=2mm,
left=6pt, right=6pt, top=4pt, bottom=4pt
]
\small
\fontsize{8}{9}\selectfont\ttfamily
New subsample score 0.0 is not better than old score 8.0, skipping
\end{tcolorbox}

\begin{tcolorbox}[
title={\small Repaired Seed, GEPA, Iteration 6 \hfill Parent: Seed},
fonttitle=\bfseries\small,
colback=gray!2,
colframe=gray!50,
boxrule=0.8pt,
breakable,
enhanced jigsaw,
pad at break=2mm,
left=6pt, right=6pt, top=4pt, bottom=4pt
]
\small
\fontsize{8}{9}\selectfont\ttfamily
New subsample score 5.0 is not better than old score 8.0, skipping
\end{tcolorbox}

\begin{tcolorbox}[
title={\small Repaired Seed, GEPA, Iteration 7 \hfill Parent: Seed},
fonttitle=\bfseries\small,
colback=gray!2,
colframe=gray!50,
boxrule=0.8pt,
breakable,
enhanced jigsaw,
pad at break=2mm,
left=6pt, right=6pt, top=4pt, bottom=4pt
]
\small
\fontsize{8}{9}\selectfont\ttfamily
New subsample score 8.0 is not better than old score 8.0, skipping
\end{tcolorbox}

\begin{tcolorbox}[
title={\small Repaired Seed, GEPA, Iteration 8 \hfill Parent: Seed},
fonttitle=\bfseries\small,
colback=gray!2,
colframe=gray!50,
boxrule=0.8pt,
breakable,
enhanced jigsaw,
pad at break=2mm,
left=6pt, right=6pt, top=4pt, bottom=4pt
]
\small
\fontsize{8}{9}\selectfont\ttfamily
\hl{You are an AI assistant tasked with solving mathematical word problems.}\\
\hl{When given a problem, follow these steps:}\\
\hl{1. **Parse the problem carefully** to identify all numbers, operations, and relationships.}\\
\hl{2. **Break the problem into logical steps**, ensuring each step is explicitly stated and mathematically precise.}\\
\hl{3. **Use domain-specific terminology correctly**, such as:}\\
\hspace*{1em}\hl{- "More than" (e.g., "five times more" means 5x the original value, not 6x).}\\
\hspace*{1em}\hl{- "Discounts" (e.g., "20\% off" reduces the price by 20\%, not to 20\%).}\\
\hspace*{1em}\hl{- "Tripling" or "halving" (explicitly multiply/divide by the stated factor).}\\
\hl{4. **Perform calculations step-by-step**, showing intermediate results and avoiding errors in arithmetic or unit conversions.}\\
\hl{5. **Verify the final answer** against the problem's context to ensure it matches the question's requirements (e.g., "how much more" requires a difference, not a total).}\\
\hl{6. **Output strictly in JSON format** with:}\\
\hspace*{1em}\hl{- `solution\_pad`: A string containing the step-by-step explanation, using newline-separated steps for clarity.}\\
\hspace*{1em}\hl{- `final\_answer`: The numerical result as a string (e.g., "20", not 20).}\\
\hl{7. **Avoid any text or explanations outside the JSON**. Ensure the final answer is unambiguous and matches the problem's expected units (e.g., dollars, minutes).}\\[4pt]
\hl{Example: For "Alice doubles her \$2000," calculate 2000 * 2 = 4000. For "Bob makes five times more," interpret this as 2000 + (2000 * 5) = 12000. Always clarify ambiguous phrasing and ensure calculations align with standard mathematical conventions.}
\end{tcolorbox}

\begin{tcolorbox}[
title={\small Repaired Seed, GEPA, Iteration 9 \hfill Parent: Seed},
fonttitle=\bfseries\small,
colback=gray!2,
colframe=gray!50,
boxrule=0.8pt,
breakable,
enhanced jigsaw,
pad at break=2mm,
left=6pt, right=6pt, top=4pt, bottom=4pt
]
\small
\fontsize{8}{9}\selectfont\ttfamily
New subsample score 7.0 is not better than old score 7.0, skipping
\end{tcolorbox}

\begin{tcolorbox}[
title={\small Repaired Seed, GEPA, Iteration 10 \hfill Parent: Iteration 8},
fonttitle=\bfseries\small,
colback=gray!2,
colframe=gray!50,
boxrule=0.8pt,
breakable,
enhanced jigsaw,
pad at break=2mm,
left=6pt, right=6pt, top=4pt, bottom=4pt
]
\small
\fontsize{8}{9}\selectfont\ttfamily
New subsample score 7.0 is not better than old score 7.0, skipping
\end{tcolorbox}

\begin{tcolorbox}[
title={\small Repaired Seed, GEPA, Iteration 11 \hfill Parent: Iteration 8},
fonttitle=\bfseries\small,
colback=gray!2,
colframe=gray!50,
boxrule=0.8pt,
breakable,
enhanced jigsaw,
pad at break=2mm,
left=6pt, right=6pt, top=4pt, bottom=4pt
]
\small
\fontsize{8}{9}\selectfont\ttfamily
New subsample score 8.0 is not better than old score 8.0, skipping
\end{tcolorbox}

\begin{tcolorbox}[
title={\small Repaired Seed, GEPA, Iteration 12 \hfill Parent: Iteration 8},
fonttitle=\bfseries\small,
colback=gray!2,
colframe=gray!50,
boxrule=0.8pt,
breakable,
enhanced jigsaw,
pad at break=2mm,
left=6pt, right=6pt, top=4pt, bottom=4pt
]
\small
\fontsize{8}{9}\selectfont\ttfamily
New subsample score 7.0 is not better than old score 7.0, skipping
\end{tcolorbox}

\begin{tcolorbox}[
title={\small Repaired Seed, GEPA, Iteration 13 \hfill Parent: Iteration 8},
fonttitle=\bfseries\small,
colback=gray!2,
colframe=gray!50,
boxrule=0.8pt,
breakable,
enhanced jigsaw,
pad at break=2mm,
left=6pt, right=6pt, top=4pt, bottom=4pt
]
\small
\fontsize{8}{9}\selectfont\ttfamily
New subsample score 8.0 is not better than old score 8.0, skipping
\end{tcolorbox}

\begin{tcolorbox}[
title={\small Repaired Seed, GEPA, Iteration 14 \hfill Parent: Seed},
fonttitle=\bfseries\small,
colback=gray!2,
colframe=gray!50,
boxrule=0.8pt,
breakable,
enhanced jigsaw,
pad at break=2mm,
left=6pt, right=6pt, top=4pt, bottom=4pt
]
\small
\fontsize{8}{9}\selectfont\ttfamily
New subsample score 5.0 is not better than old score 6.0, skipping
\end{tcolorbox}

\begin{tcolorbox}[
title={\small Repaired Seed, GEPA, Iteration 15 \hfill Parent: Iteration 8},
fonttitle=\bfseries\small,
colback=gray!2,
colframe=gray!50,
boxrule=0.8pt,
breakable,
enhanced jigsaw,
pad at break=2mm,
left=6pt, right=6pt, top=4pt, bottom=4pt
]
\small
\fontsize{8}{9}\selectfont\ttfamily
You are an AI assistant tasked with solving mathematical word problems.\\
When given a problem, follow these steps:\\
1. **Parse the problem carefully**, identifying all numbers, operations, and relationships. \hl{Pay special attention to ambiguous phrasing like "more than" or "times more," which often indicate addition or multiplication (e.g., "five times more" means 5x the original value, not 6x).}\\
2. **Break the problem into logical steps**, explicitly stating each step and ensuring mathematical precision. \hl{For example, interpret "8 more solar systems for every planet" as 8 additional systems per planet (totaling 9 systems per planet).}\\
3. **Use domain-specific terminology correctly**, such as:\\
\hspace*{1em}- "Discounts" (e.g., "20\% off" reduces the price by 20\%, not to 20\%).\\
\hspace*{1em}- "Tripling" or "halving" (explicitly multiply/divide by the stated factor).\\
\hspace*{1em}- "More than" (e.g., "five times more" means 5x the original value, not 6x).\\
4. **Perform calculations step-by-step**, showing intermediate results and avoiding arithmetic errors. \hl{For example, if the problem involves multiple stages (e.g., loading crates with weight limits), calculate total weight, capacity, and excess separately.}\\
5. **Verify the final answer** against the problem's context to ensure it matches the question's requirements. For instance, "how much more" requires a difference, not a total.\\
6. **Output strictly in JSON format** with:\\
\hspace*{1em}- `solution\_pad`: step-by-step explanation.\\
\hspace*{1em}- `final\_answer`: numerical result as a string (e.g., "20", not 20).\\
7. **Avoid any text or explanations outside the JSON**. Ensure the final answer is unambiguous and matches the problem's expected units (e.g., dollars, minutes).\\[4pt]
\hl{**Key clarifications to avoid common mistakes**:}\\
\hl{- "Five times more" means 5x the original value (not 6x).}\\
\hl{- "Twice as much" means 2x the original value.}\\
\hl{- "More than" often indicates addition (e.g., "8 more for every planet" means 8 added to the base quantity).}\\
\hl{- Always check for hidden constraints (e.g., weight limits, subdivisions of items).}\\
\hl{- Use equations to model relationships (e.g., for discounts: original price - discount = final price).}
\end{tcolorbox}

\begin{tcolorbox}[
title={\small Repaired Seed, GEPA, Iteration 16 \hfill Parent: Iteration 8},
fonttitle=\bfseries\small,
colback=gray!2,
colframe=gray!50,
boxrule=0.8pt,
breakable,
enhanced jigsaw,
pad at break=2mm,
left=6pt, right=6pt, top=4pt, bottom=4pt
]
\small
\fontsize{8}{9}\selectfont\ttfamily
New subsample score 7.0 is not better than old score 7.0, skipping
\end{tcolorbox}

\begin{tcolorbox}[
title={\small Repaired Seed, GEPA, Iteration 17 \hfill Parent: Seed},
fonttitle=\bfseries\small,
colback=gray!2,
colframe=gray!50,
boxrule=0.8pt,
breakable,
enhanced jigsaw,
pad at break=2mm,
left=6pt, right=6pt, top=4pt, bottom=4pt
]
\small
\fontsize{8}{9}\selectfont\ttfamily
New subsample score 8.0 is not better than old score 8.0, skipping
\end{tcolorbox}

\begin{tcolorbox}[
title={\small Repaired Seed, GEPA, Iteration 18 \hfill Parent: Iteration 8},
fonttitle=\bfseries\small,
colback=gray!2,
colframe=gray!50,
boxrule=0.8pt,
breakable,
enhanced jigsaw,
pad at break=2mm,
left=6pt, right=6pt, top=4pt, bottom=4pt
]
\small
\fontsize{8}{9}\selectfont\ttfamily
New subsample score 8.0 is not better than old score 8.0, skipping
\end{tcolorbox}

\begin{tcolorbox}[
title={\small Repaired Seed, GEPA, Iteration 19 \hfill Parent: Seed},
fonttitle=\bfseries\small,
colback=gray!2,
colframe=gray!50,
boxrule=0.8pt,
breakable,
enhanced jigsaw,
pad at break=2mm,
left=6pt, right=6pt, top=4pt, bottom=4pt
]
\small
\fontsize{8}{9}\selectfont\ttfamily
New subsample score 7.0 is not better than old score 7.0, skipping
\end{tcolorbox}

\begin{tcolorbox}[
title={\small Repaired Seed, GEPA, Iteration 20 \hfill Parent: Seed},
fonttitle=\bfseries\small,
colback=gray!2,
colframe=gray!50,
boxrule=0.8pt,
breakable,
enhanced jigsaw,
pad at break=2mm,
left=6pt, right=6pt, top=4pt, bottom=4pt
]
\small
\fontsize{8}{9}\selectfont\ttfamily
\hl{You are an AI assistant tasked with solving mathematical word problems. When given a problem, you must generate a JSON object with two keys: "solution\_pad" and "final\_answer".}\\[4pt]
\hl{For "solution\_pad", provide a clear, step-by-step explanation of the solution. Break down each calculation explicitly, using arithmetic operations (e.g., "0.4 * 60 = 24") and logical steps (e.g., "Subtract to find the remaining quantity"). Ensure all intermediate steps are shown, even for simple operations. Avoid markdown and use plain text.}\\[4pt]
\hl{For "final\_answer", output only the numerical result of the problem, without any text, units, or explanations. Ensure the answer is correctly formatted as a number (e.g., 36, 77.00, 565).}\\[4pt]
\hl{Key requirements:}\\
\hl{1. **Strict JSON format**: Ensure the output is a valid JSON object with no extra text, comments, or formatting.}\\
\hl{2. **Correctness**: Verify all calculations are accurate and align with the problem's context (e.g., percentages, fractions, cost totals, averages).}\\
\hl{3. **Domain-specific handling**: Account for problem-specific details (e.g., "each carton has 20 jars," "10\% less than the average").}\\
\hl{4. **Generalizable strategy**: Apply logical steps like identifying given values, determining operations, and solving sequentially.}\\[4pt]
\hl{Example: If the problem is "A pie is sliced into 8 pieces. 1/2 is given to Joe, 1/4 to Carl. How many slices remain?", the solution\_pad should detail each step (e.g., "1/2 of 8 = 4 slices; 1/4 of 8 = 2 slices; 8 - 4 - 2 = 2") and the final\_answer is "2".}\\[2pt]
\hl{Always validate that the final answer matches the problem's requirements exactly, including units or decimal precision if specified.}

\medskip
\hrule\vspace{8pt}
\normalfont\normalsize
\small\textcolor{green!60!black}{\faCheck}\; The correct field order (\texttt{solution\_pad} before \texttt{final\_answer}) is preserved across all 20 iterations. Optimizations progressively improve reasoning guidelines, domain-specific heuristics, and disambiguation strategies without disrupting the structural format.
\end{tcolorbox}

\subsubsection{VISTA}

\begin{tcolorbox}[
title={\textbf{Repaired Seed, VISTA, Iteration 1 \hfill Parent: Seed}},
fonttitle=\bfseries\small,
colback=gray!2,
colframe=gray!50,
boxrule=0.8pt,
breakable,
enhanced,
left=6pt, right=6pt, top=4pt, bottom=4pt
]
\small
Strategy: No failed samples found. Falling back to single mutation.\\[4pt]
Result: Candidate subsample score 8.0 is not better than old score 8.0, skipping.
\end{tcolorbox}

\begin{tcolorbox}[
title={\textbf{Repaired Seed, VISTA, Iteration 2 \hfill Parent: Seed}},
fonttitle=\bfseries\small,
colback=gray!2,
colframe=gray!50,
boxrule=0.8pt,
breakable,
enhanced,
left=6pt, right=6pt, top=4pt, bottom=4pt
]
\small
Strategy: No failed samples found. Falling back to single mutation.\\[4pt]
Result: Candidate subsample score 6.0 is not better than old score 8.0, skipping.
\end{tcolorbox}

\begin{tcolorbox}[
title={\textbf{Repaired Seed, VISTA, Iteration 3 \hfill Parent: Seed}},
fonttitle=\bfseries\small,
colback=gray!2,
colframe=gray!50,
boxrule=0.8pt,
breakable,
enhanced,
left=6pt, right=6pt, top=4pt, bottom=4pt
]
\small
Strategy: \texttt{heuristics}\\[4pt]
\begin{tabular}{@{}llr@{}}
\toprule
\textbf{Hyp.} & \textbf{Label} & \textbf{$\Delta$acc} \\
\midrule
H1 & \texttt{task\_instruction\_clarity} & +1.0 \textcolor{green!60!black}{\faCheck} \\
H2 & \texttt{reasoning\_strategy} & -1.0 \\
H3 & \texttt{edge\_case\_handling} & +1.0 \\
\bottomrule
\end{tabular}\\[4pt]
\textcolor{green!60!black}{Selected: H1 [\texttt{task\_instruction\_clarity}] --- Candidate Accepted}\\[2pt]
\hrule\vspace{8pt}
\small
{\fontsize{8}{9}\selectfont\ttfamily
\hl{\{} \\
\hspace*{1em}\hl{"role":} "You are an AI assistant that solves mathematical word problems.", \\
\hspace*{1em}\hl{"task":} "You will be given a question and you need to provide a step-by-step solution to the problem. Finally, you will provide the answer to the question. When outputting the final answer, make sure there are no other text or explanations included, just the answer itself.", \\
\hspace*{1em}\hl{"clarification": "When a problem states that there are `X more' of something for every Y, interpret this as X additional units per Y, and include both the Y and the X units in the total count. If the problem uses phrases like `X times more' or `X times as many,' clarify that these typically mean X times the original amount, not X times plus the original.",} \\
\hspace*{1em}\hl{"output\_format": "The expected output must be a JSON object with the following format: \{\"solution\_pad\": <the step-by-step solution to the problem>, \"final\_answer\": <the final answer to the question>\}. Strictly follow the format provided above and ensure that your output is a valid JSON object. Any deviation from this format will result in an error.",} \\
\hspace*{1em}\hl{"schema": "The JSON output must include two fields: `solution\_pad' (a string containing the step-by-step reasoning) and `final\_answer' (a numerical value representing the final result). Ensure that the final answer is correctly formatted as a number, not a text string."} \\
\hl{\}}
}\\[0pt]
\end{tcolorbox}

\begin{tcolorbox}[
title={\textbf{Repaired Seed, VISTA, Iteration 4 \hfill Parent: Seed}},
fonttitle=\bfseries\small,
colback=gray!2,
colframe=gray!50,
boxrule=0.8pt,
breakable,
enhanced,
left=6pt, right=6pt, top=4pt, bottom=4pt
]
\small
Strategy: \texttt{heuristics}\\[4pt]
\begin{tabular}{@{}llr@{}}
\toprule
\textbf{Hyp.} & \textbf{Label} & \textbf{$\Delta$acc} \\
\midrule
H1 & \texttt{task\_instruction\_clarity} & 0.0 \\
H2 & \texttt{format\_and\_syntax} & 0.0 \\
H3 & \texttt{missing\_domain\_knowledge} & 0.0 \\
\bottomrule
\end{tabular}\\[4pt]
No multi-hypothesis candidate improved over parent. No proposal returned.
\end{tcolorbox}

\begin{tcolorbox}[
title={\textbf{Repaired Seed, VISTA, Iteration 5 \hfill Parent: Seed}},
fonttitle=\bfseries\small,
colback=gray!2,
colframe=gray!50,
boxrule=0.8pt,
breakable,
enhanced,
left=6pt, right=6pt, top=4pt, bottom=4pt
]
\small
Strategy: No failed samples found. Falling back to single mutation.\\[4pt]
Result: Candidate subsample score 1.0 is not better than old score 8.0, skipping.
\end{tcolorbox}

\begin{tcolorbox}[
title={\textbf{Repaired Seed, VISTA, Iteration 6 \hfill Parent: Iteration 3}},
fonttitle=\bfseries\small,
colback=gray!2,
colframe=gray!50,
boxrule=0.8pt,
breakable,
enhanced,
left=6pt, right=6pt, top=4pt, bottom=4pt
]
\small
Strategy: \texttt{random\_restart}\\[4pt]
Random restart completed with no improvement. No proposal returned.\\
Validation: old\_sum=8.0000, new\_sum=6.0000, improved=False.
\end{tcolorbox}

\begin{tcolorbox}[
title={\textbf{Repaired Seed, VISTA, Iteration 7 \hfill Parent: Iteration 3}},
fonttitle=\bfseries\small,
colback=gray!2,
colframe=gray!50,
boxrule=0.8pt,
breakable,
enhanced,
left=6pt, right=6pt, top=4pt, bottom=4pt
]
\small
Strategy: No failed samples found. Falling back to single mutation.\\[4pt]
Result: Candidate subsample score 8.0 is not better than old score 8.0, skipping.
\end{tcolorbox}

\begin{tcolorbox}[
title={\textbf{Repaired Seed, VISTA, Iteration 8 \hfill Parent: Seed}},
fonttitle=\bfseries\small,
colback=gray!2,
colframe=gray!50,
boxrule=0.8pt,
breakable,
enhanced,
left=6pt, right=6pt, top=4pt, bottom=4pt
]
\small
Strategy: No failed samples found. Falling back to single mutation.\\[4pt]
Result: Candidate subsample score 7.0 is not better than old score 8.0, skipping.
\end{tcolorbox}

\begin{tcolorbox}[
title={\textbf{Repaired Seed, VISTA, Iteration 9 \hfill Parent: Seed}},
fonttitle=\bfseries\small,
colback=gray!2,
colframe=gray!50,
boxrule=0.8pt,
breakable,
enhanced,
left=6pt, right=6pt, top=4pt, bottom=4pt
]
\small
Strategy: \texttt{heuristics}\\[4pt]
\begin{tabular}{@{}llr@{}}
\toprule
\textbf{Hyp.} & \textbf{Label} & \textbf{$\Delta$acc} \\
\midrule
H1 & \texttt{task\_instruction\_clarity} & +1.0 \\
H2 & \texttt{reasoning\_strategy} & +2.0 \textcolor{green!60!black}{\faCheck} \\
H3 & \texttt{missing\_domain\_knowledge} & +1.0 \\
\bottomrule
\end{tabular}\\[4pt]
\textcolor{green!60!black}{Selected: H2 [\texttt{reasoning\_strategy}] --- Candidate Accepted}\\[2pt]
\hrule\vspace{8pt}
\small
{\fontsize{8}{9}\selectfont\ttfamily
You are an AI assistant that solves mathematical word problems. You will be given a question and you need to provide a step-by-step solution to the problem. Finally, you will provide the answer to the question. When outputting the final answer, make sure there are no other text or explanations included, just the answer itself.\\[4pt]
The expected output must be a JSON object with the following format:\\
\{ \\
\hspace*{1em}"solution\_pad": <the step-by-step solution to the problem>, \\
\hspace*{1em}"final\_answer": <the final answer to the question> \\
\} \\[4pt]
Strictly follow the format provided above and ensure that your output is a valid JSON object. Any deviation from this format will result in an error.\\[4pt]
\hl{When solving problems involving time, work rates, or quantities with pauses, explicitly account for interruptions by first calculating the effective working time (total time minus break duration) before computing total output. Ensure all steps are clearly articulated in the solution\_pad, including any adjustments for pauses or breaks. Maintain the general structure of the solution, but prioritize accuracy in time-based calculations by adhering to this principle.}
}\\[0pt]
\end{tcolorbox}

\begin{tcolorbox}[
title={\textbf{Repaired Seed, VISTA, Iteration 10 \hfill Parent: Iteration 3}},
fonttitle=\bfseries\small,
colback=gray!2,
colframe=gray!50,
boxrule=0.8pt,
breakable,
enhanced,
left=6pt, right=6pt, top=4pt, bottom=4pt
]
\small
Strategy: \texttt{heuristics}\\[4pt]
\begin{tabular}{@{}llr@{}}
\toprule
\textbf{Hyp.} & \textbf{Label} & \textbf{$\Delta$acc} \\
\midrule
H1 & \texttt{task\_instruction\_clarity} & +1.0 \textcolor{green!60!black}{\faCheck} \\
H2 & \texttt{edge\_case\_handling} & +1.0 \\
H3 & \texttt{reasoning\_strategy} & 0.0 \\
\bottomrule
\end{tabular}\\[4pt]
\textcolor{green!60!black}{Selected: H1 [\texttt{task\_instruction\_clarity}] --- Candidate Accepted}\\[2pt]
\hrule\vspace{8pt}
\small
{\fontsize{8}{9}\selectfont\ttfamily
You are an AI assistant that solves mathematical word problems. You will be given a question and you need to provide a step-by-step solution to the problem. Finally, you will provide the final answer to the question. When outputting the final answer, make sure there are no other text or explanations included, just the answer itself.\\[4pt]
When solving problems involving relative speed (e.g., one entity catching up to another), calculate the time based on the relative speed (difference in speeds) and the initial distance covered by the head start. Ensure units are consistent (e.g., convert minutes to hours if necessary).\\[4pt]
\hl{For problems involving spacing or positioning of entities in a confined space (e.g., boats, vehicles, objects), explicitly account for edge cases by subtracting the clearance required for the last entity from the total dimension before dividing by the per-entity space. This ensures accurate calculations for scenarios where the final entity's clearance overlaps with boundaries.}\\[4pt]
For all other problems, follow standard arithmetic or algebraic reasoning.\\[4pt]
The expected output must be a JSON object with the following format:\\
\{ \\
\hspace*{1em}"solution\_pad": <the step-by-step solution to the problem>, \\
\hspace*{1em}"final\_answer": <the final answer to the question> \\
\} \\[4pt]
Strictly follow the format provided above and ensure that your output is a valid JSON object. Any deviation from this format will result in an error.
}\\[0pt]
\end{tcolorbox}

\subsection{Minimal Seed}

\subsubsection{GEPA}


\begin{tcolorbox}[
title={\small Minimal Seed, GEPA, Iteration 1 \hfill Parent: Seed},
fonttitle=\bfseries\small,
colback=gray!2,
colframe=gray!50,
boxrule=0.8pt,
breakable,
enhanced jigsaw,
pad at break=2mm,
left=6pt, right=6pt, top=4pt, bottom=4pt
]
\small
\fontsize{8}{9}\selectfont\ttfamily
New subsample score 1.0 is not better than old score 1.0, skipping
\end{tcolorbox}

\begin{tcolorbox}[
title={\small Minimal Seed, GEPA, Iteration 2 \hfill Parent: Seed},
fonttitle=\bfseries\small,
colback=gray!2,
colframe=gray!50,
boxrule=0.8pt,
breakable,
enhanced jigsaw,
pad at break=2mm,
left=6pt, right=6pt, top=4pt, bottom=4pt
]
\small
\fontsize{8}{9}\selectfont\ttfamily
New subsample score 0.0 is not better than old score 2.0, skipping
\end{tcolorbox}

\begin{tcolorbox}[
title={\small Minimal Seed, GEPA, Iteration 3 \hfill Parent: Seed},
fonttitle=\bfseries\small,
colback=gray!2,
colframe=gray!50,
boxrule=0.8pt,
breakable,
enhanced jigsaw,
pad at break=2mm,
left=6pt, right=6pt, top=4pt, bottom=4pt
]
\small
\fontsize{8}{9}\selectfont\ttfamily
\hl{Solve the given math problem step-by-step, ensuring all calculations are correct. Output only a single JSON object with the key "final\_answer" and the correct numerical value as the value. Do not include any explanations, equations, or text inside the JSON value. Ensure the JSON is valid and free of syntax errors. If the answer requires multiple steps, compute the result accurately and present it as a single number.}
\end{tcolorbox}

\begin{tcolorbox}[
title={\small Minimal Seed, GEPA, Iteration 4 \hfill Parent: Iteration 3},
fonttitle=\bfseries\small,
colback=gray!2,
colframe=gray!50,
boxrule=0.8pt,
breakable,
enhanced jigsaw,
pad at break=2mm,
left=6pt, right=6pt, top=4pt, bottom=4pt
]
\small
\fontsize{8}{9}\selectfont\ttfamily
Solve the given math problem step-by-step, ensuring all calculations are correct. Output only a single JSON object with the key "final\_answer" and the correct numerical value as the value. Do not include any explanations, equations, or text inside the JSON value. Ensure the JSON is valid and free of syntax errors. If the answer requires multiple steps, compute the result accurately and present it as a single number.
\hl{Verify that the final answer is a fully calculated numeric value (e.g., 42, not 2+2) and matches the exact expected result. Avoid any expressions or intermediate steps in the JSON value. If the problem involves averages, totals, or comparisons, ensure the final answer reflects the correct aggregation or difference as per the problem's requirements.}
\end{tcolorbox}

\begin{tcolorbox}[
title={\small Minimal Seed, GEPA, Iteration 5 \hfill Parent: Seed},
fonttitle=\bfseries\small,
colback=gray!2,
colframe=gray!50,
boxrule=0.8pt,
breakable,
enhanced jigsaw,
pad at break=2mm,
left=6pt, right=6pt, top=4pt, bottom=4pt
]
\small
\fontsize{8}{9}\selectfont\ttfamily
New subsample score 0.0 is not better than old score 2.0, skipping
\end{tcolorbox}

\begin{tcolorbox}[
title={\small Minimal Seed, GEPA, Iteration 6 \hfill Parent: Iteration 3},
fonttitle=\bfseries\small,
colback=gray!2,
colframe=gray!50,
boxrule=0.8pt,
breakable,
enhanced jigsaw,
pad at break=2mm,
left=6pt, right=6pt, top=4pt, bottom=4pt
]
\small
\fontsize{8}{9}\selectfont\ttfamily
New subsample score 1.0 is not better than old score 1.0, skipping
\end{tcolorbox}

\begin{tcolorbox}[
title={\small Minimal Seed, GEPA, Iteration 7 \hfill Parent: Iteration 3},
fonttitle=\bfseries\small,
colback=gray!2,
colframe=gray!50,
boxrule=0.8pt,
breakable,
enhanced jigsaw,
pad at break=2mm,
left=6pt, right=6pt, top=4pt, bottom=4pt
]
\small
\fontsize{8}{9}\selectfont\ttfamily
New subsample score 2.0 is not better than old score 2.0, skipping
\end{tcolorbox}

\begin{tcolorbox}[
title={\small Minimal Seed, GEPA, Iteration 8 \hfill Parent: Iteration 3},
fonttitle=\bfseries\small,
colback=gray!2,
colframe=gray!50,
boxrule=0.8pt,
breakable,
enhanced jigsaw,
pad at break=2mm,
left=6pt, right=6pt, top=4pt, bottom=4pt
]
\small
\fontsize{8}{9}\selectfont\ttfamily
Solve the given math problem step-by-step, ensuring all calculations are correct. Output only a single JSON object with the key "final\_answer" and the correct numerical value as the value. Do not include any explanations, equations, or text inside the JSON value. Ensure the JSON is valid and free of syntax errors. If the answer requires multiple steps, compute the result accurately and present it as a single number.
\hl{For time-based problems, calculate durations using absolute time differences (e.g., 8:20 - 8:00 = 20 minutes). For percentage discounts, apply each discount sequentially to the updated price (e.g., 20\% off then 25\% off). For averages, sum all values and divide by the total count of items. Always verify intermediate steps to avoid errors in arithmetic or logical sequencing.}
\end{tcolorbox}

\begin{tcolorbox}[
title={\small Minimal Seed, GEPA, Iteration 9 \hfill Parent: Iteration 3},
fonttitle=\bfseries\small,
colback=gray!2,
colframe=gray!50,
boxrule=0.8pt,
breakable,
enhanced jigsaw,
pad at break=2mm,
left=6pt, right=6pt, top=4pt, bottom=4pt
]
\small
\fontsize{8}{9}\selectfont\ttfamily
New subsample score 2.0 is not better than old score 2.0, skipping
\end{tcolorbox}

\begin{tcolorbox}[
title={\small Minimal Seed, GEPA, Iteration 10 \hfill Parent: Iteration 4},
fonttitle=\bfseries\small,
colback=gray!2,
colframe=gray!50,
boxrule=0.8pt,
breakable,
enhanced jigsaw,
pad at break=2mm,
left=6pt, right=6pt, top=4pt, bottom=4pt
]
\small
\fontsize{8}{9}\selectfont\ttfamily
New subsample score 3.0 is not better than old score 3.0, skipping
\end{tcolorbox}

\begin{tcolorbox}[
title={\small Minimal Seed, GEPA, Iteration 11 \hfill Parent: Iteration 3},
fonttitle=\bfseries\small,
colback=gray!2,
colframe=gray!50,
boxrule=0.8pt,
breakable,
enhanced jigsaw,
pad at break=2mm,
left=6pt, right=6pt, top=4pt, bottom=4pt
]
\small
\fontsize{8}{9}\selectfont\ttfamily
New subsample score 1.0 is not better than old score 2.0, skipping
\end{tcolorbox}

\begin{tcolorbox}[
title={\small Minimal Seed, GEPA, Iteration 12 \hfill Parent: Iteration 4},
fonttitle=\bfseries\small,
colback=gray!2,
colframe=gray!50,
boxrule=0.8pt,
breakable,
enhanced jigsaw,
pad at break=2mm,
left=6pt, right=6pt, top=4pt, bottom=4pt
]
\small
\fontsize{8}{9}\selectfont\ttfamily
New subsample score 4.0 is not better than old score 4.0, skipping
\end{tcolorbox}

\begin{tcolorbox}[
title={\small Minimal Seed, GEPA, Iteration 13 \hfill Parent: Iteration 4},
fonttitle=\bfseries\small,
colback=gray!2,
colframe=gray!50,
boxrule=0.8pt,
breakable,
enhanced jigsaw,
pad at break=2mm,
left=6pt, right=6pt, top=4pt, bottom=4pt
]
\small
\fontsize{8}{9}\selectfont\ttfamily
Solve the given math problem step-by-step, ensuring all calculations are correct. Output only a single JSON object with the key "final\_answer" and the correct numerical value as the value. Do not include any explanations, equations, or text inside the JSON value. Ensure the JSON is valid and free of syntax errors. If the answer requires multiple steps, compute the result accurately and present it as a single number. Verify that the final answer is a fully calculated numeric value (e.g., 42, not 2+2) and matches the exact expected result. Avoid any expressions or intermediate steps in the JSON value. If the problem involves averages, totals, or comparisons, ensure the final answer reflects the correct aggregation or difference as per the problem's requirements.\\[4pt]
\hl{**Key Details to Include in Your Reasoning:**}\\
\hl{1. **Step-by-Step Breakdown:** Explicitly outline all required calculations, including intermediate steps (e.g., adding quantities, applying rates, calculating spacing).}\\
\hl{2. **Domain-Specific Rules:** Account for niche constraints (e.g., spacing between objects, rate adjustments, cumulative weight increments).}\\
\hl{3. **Verification:** Double-check arithmetic, ensure units are consistent, and confirm that all problem-specific conditions are satisfied.}\\
\hl{4. **Edge Cases:** Address scenarios like partial spaces (e.g., last boat requiring reduced spacing), rounding rules, or hidden totals (e.g., summing multiple categories).}\\
\hl{5. **Final Validation:** Ensure the answer is a single numeric value, not a formula or text, and matches the problem's exact requirements (e.g., "how many more miles," "total expenses").}\\[4pt]
\hl{**Examples of Correct/Incorrect Patterns:**}\\
\hl{- **Correct:** \{"final\_answer": 8\} (e.g., river spacing problem where 42ft river allows 8 boats).}\\
\hl{- **Incorrect:** \{"final\_answer": 4\} (same problem but missing adjustment for final boat spacing).}\\
\hl{- **Correct:** \{"final\_answer": 86\} (e.g., test score problem where Marco's 81 + 5 = 86).}\\
\hl{- **Incorrect:** \{"final\_answer": 81\} (same problem but omitting Margaret's 5-point addition).}
\end{tcolorbox}

\begin{tcolorbox}[
title={\small Minimal Seed, GEPA, Iteration 14 \hfill Parent: Iteration 13},
fonttitle=\bfseries\small,
colback=gray!2,
colframe=gray!50,
boxrule=0.8pt,
breakable,
enhanced jigsaw,
pad at break=2mm,
left=6pt, right=6pt, top=4pt, bottom=4pt
]
\small
\fontsize{8}{9}\selectfont\ttfamily
Solve the given math problem step-by-step, ensuring all calculations are correct. Output only a single JSON object with the key "final\_answer" and the correct numerical value as the value. Do not include any explanations, equations, or text inside the JSON value. Ensure the JSON is valid and free of syntax errors. If the answer requires multiple steps, compute the result accurately and present it as a single number. Verify that the final answer is a fully calculated numeric value (e.g., 42, not 2+2) and matches the exact expected result. Avoid any expressions or intermediate steps in the JSON value. If the problem involves averages, totals, or comparisons, ensure the final answer reflects the correct aggregation or difference as per the problem's requirements.\\[4pt]
**Key Details to Include in Your Reasoning:**\\
1. **Step-by-Step Breakdown:** Explicitly outline all required calculations, including intermediate steps (e.g., adding quantities, applying rates, calculating spacing).\\
\hl{2. **Domain-Specific Rules:** Account for niche constraints (e.g., spacing between objects, rate adjustments, cumulative weight increments). For example:}\\
\hspace*{1em}\hl{- In spacing problems, subtract the final object's space if partial spacing is required (e.g., last boat reduces spacing).}\\
\hspace*{1em}\hl{- In percentage problems, ensure "less than" or "more than" is applied correctly (e.g., 1/10 less = 90\% of original).}\\
3. **Verification:** Double-check arithmetic, ensure units are consistent, and confirm that all problem-specific conditions are satisfied.\\
\hl{4. **Edge Cases:** Address scenarios like partial spaces, rounding rules, or hidden totals (e.g., summing multiple categories). For example:}\\
\hspace*{1em}\hl{- If a problem involves multiple categories, ensure all are summed explicitly.}\\
\hspace*{1em}\hl{- For tip calculations, add the tip to the base cost.}\\
5. **Final Validation:** Ensure the answer is a single numeric value, not a formula or text, and matches the problem's exact requirements (e.g., "how many more miles," "total expenses").\\[4pt]
\hl{**Examples of Correct/Incorrect Patterns:**}\\
\hl{- **Correct:** \{"final\_answer": 18\} (e.g., taxi cost with 20\% tip: \$15 + \$3 = \$18).}\\
\hl{- **Incorrect:** \{"final\_answer": 26\} (same problem but missing tip calculation).}\\
\hl{- **Correct:** \{"final\_answer": 296\} (e.g., total marks: 70 + 63 + 83 + 80 = 296).}\\
\hl{- **Incorrect:** \{"final\_answer": 160\} (same problem but omitting Social Studies or Arts scores).}\\
\hl{- **Correct:** \{"final\_answer": 1890\} (e.g., total expenses: (250+375+320) * 2 = 1890).}\\
\hl{- **Incorrect:** \{"final\_answer": 2000\} (same problem but incorrect flower count).}

\medskip
\hrule\vspace{8pt}
\normalfont\normalsize
\small\textcolor{red!80!black}{\faTimes}\; The missing \texttt{solution\_pad} field is never introduced across all 14 iterations. The JSON output remains \texttt{\{"final\_answer": <number>\}} throughout. All modifications target reasoning quality and domain-specific heuristics; the structural incompleteness is never addressed.
\end{tcolorbox}

\subsubsection{VISTA}

\begin{tcolorbox}[
title={\textbf{Minimal Seed, VISTA, Iteration 1 \hfill Parent: Seed}},
fonttitle=\bfseries\small,
colback=gray!2,
colframe=gray!50,
boxrule=0.8pt,
breakable,
enhanced,
left=6pt, right=6pt, top=4pt, bottom=4pt
]
\small
Strategy: \texttt{heuristics}\\[4pt]
\begin{tabular}{@{}llr@{}}
\toprule
\textbf{Hyp.} & \textbf{Label} & \textbf{$\Delta$acc} \\
\midrule
H1 & \texttt{cot\_field\_ordering} & -2.0 \\
H2 & \texttt{format\_and\_syntax} & 0.0 \\
H3 & \texttt{task\_instruction\_clarity} & -1.0 \\
\bottomrule
\end{tabular}\\[4pt]
No multi-hypothesis candidate improved over parent. No proposal returned.
\end{tcolorbox}

\begin{tcolorbox}[
title={\textbf{Minimal Seed, VISTA, Iteration 2 \hfill Parent: Seed}},
fonttitle=\bfseries\small,
colback=gray!2,
colframe=gray!50,
boxrule=0.8pt,
breakable,
enhanced,
left=6pt, right=6pt, top=4pt, bottom=4pt
]
\small
Strategy: \texttt{heuristics}\\[4pt]
\begin{tabular}{@{}llr@{}}
\toprule
\textbf{Hyp.} & \textbf{Label} & \textbf{$\Delta$acc} \\
\midrule
H1 & \texttt{cot\_field\_ordering} & +6.0 \textcolor{green!60!black}{\faCheck} \\
H2 & \texttt{task\_instruction\_clarity} & -1.0 \\
H3 & \texttt{missing\_domain\_knowledge} & 0.0 \\
\bottomrule
\end{tabular}\\[4pt]
\textcolor{green!60!black}{Selected: H1 [\texttt{cot\_field\_ordering}] --- Candidate Accepted}\\[2pt]
\hrule\vspace{8pt}
\small
{\fontsize{8}{9}\selectfont\ttfamily
\hl{Solve the mathematical word problem by first generating a detailed "solution\_pad" field that outlines all intermediate steps and calculations, then providing the final answer in the "final\_answer" field. The JSON output must strictly follow this structure: \{"solution\_pad": "<step-by-step reasoning with calculations>", "final\_answer": "<correct final answer>"\}.}\\[4pt]
\hl{For problems involving multiple steps, ensure each calculation is explicitly written in the solution\_pad (e.g., "3/4 * 40,000 = <<3/4*40000=30000>>30000"). For percentage or ratio problems, show all conversion steps. For multi-part problems, break down each component into separate calculations. Only after fully documenting the reasoning process should the final\_answer be provided. The final\_answer must match the exact numerical value and formatting specified in the problem's expected solution.}
}\\[0pt]
\end{tcolorbox}

\begin{tcolorbox}[
title={\textbf{Minimal Seed, VISTA, Iteration 3 \hfill Parent: Seed}},
fonttitle=\bfseries\small,
colback=gray!2,
colframe=gray!50,
boxrule=0.8pt,
breakable,
enhanced,
left=6pt, right=6pt, top=4pt, bottom=4pt
]
\small
Strategy: \texttt{heuristics}\\[4pt]
\begin{tabular}{@{}llr@{}}
\toprule
\textbf{Hyp.} & \textbf{Label} & \textbf{$\Delta$acc} \\
\midrule
H1 & \texttt{format\_and\_syntax} & +3.0 \textcolor{green!60!black}{\faCheck} \\
H2 & \texttt{task\_instruction\_clarity} & +2.0 \\
H3 & \texttt{missing\_domain\_knowledge} & -1.0 \\
\bottomrule
\end{tabular}\\[4pt]
\textcolor{green!60!black}{Selected: H1 [\texttt{format\_and\_syntax}] --- Candidate Accepted}\\[2pt]
\hrule\vspace{8pt}
\small
{\fontsize{8}{9}\selectfont\ttfamily
\hl{Solve the mathematical problem and output a strictly formatted JSON object with the following structure:}\\
\hl{\{}\\
\hspace*{1em}\hl{"final\_answer": <numeric\_value>}\\
\hl{\}}\\[4pt]
\hl{**Rules**:}\\
\hl{1. **Numeric Final Answer**: The value in the "final\_answer" field must be a single numeric value (e.g., 42, 3.14, -5). Do not include expressions, calculations, or text (e.g., 5 + 3, 10 * 2).}\\
\hl{2. **JSON Validation**: Ensure the output is valid JSON with proper syntax (e.g., commas, brackets).}\\
\hl{3. **No Extra Fields**: Only include the "final\_answer" key. Do not add additional fields, explanations, or formatting (e.g., markdown, code blocks).}\\
\hl{4. **Accuracy**: Compute the exact numerical result of the problem, ensuring alignment with the problem's context and mathematical principles.}\\[4pt]
\hl{**Example**:}\\
\hl{For a problem like "What is 10 + 20?", the correct output is:}\\
\hl{\{"final\_answer": 30\}}
}\\[0pt]
\end{tcolorbox}

\begin{tcolorbox}[
title={\textbf{Minimal Seed, VISTA, Iteration 4 \hfill Parent: Iteration 2}},
fonttitle=\bfseries\small,
colback=gray!2,
colframe=gray!50,
boxrule=0.8pt,
breakable,
enhanced,
left=6pt, right=6pt, top=4pt, bottom=4pt
]
\small
Strategy: \texttt{heuristics}\\[4pt]
\begin{tabular}{@{}llr@{}}
\toprule
\textbf{Hyp.} & \textbf{Label} & \textbf{$\Delta$acc} \\
\midrule
H1 & \texttt{format\_and\_syntax} & 0.0 \\
H2 & \texttt{task\_instruction\_clarity} & -1.0 \\
H3 & \texttt{reasoning\_strategy} & -1.0 \\
\bottomrule
\end{tabular}\\[4pt]
No multi-hypothesis candidate improved over parent. No proposal returned.
\end{tcolorbox}

\begin{tcolorbox}[
title={\textbf{Minimal Seed, VISTA, Iteration 5 \hfill Parent: Iteration 2}},
fonttitle=\bfseries\small,
colback=gray!2,
colframe=gray!50,
boxrule=0.8pt,
breakable,
enhanced,
left=6pt, right=6pt, top=4pt, bottom=4pt
]
\small
Strategy: \texttt{heuristics}\\[4pt]
\begin{tabular}{@{}llr@{}}
\toprule
\textbf{Hyp.} & \textbf{Label} & \textbf{$\Delta$acc} \\
\midrule
H1 & \texttt{format\_and\_syntax} & +2.0 \textcolor{green!60!black}{\faCheck} \\
H2 & \texttt{reasoning\_strategy} & +2.0 \\
H3 & \texttt{task\_instruction\_clarity} & -5.0 \\
\bottomrule
\end{tabular}\\[4pt]
\textcolor{green!60!black}{Selected: H1 [\texttt{format\_and\_syntax}] --- Candidate Accepted}\\[2pt]
\hrule\vspace{8pt}
\small
{\fontsize{8}{9}\selectfont\ttfamily
Solve the mathematical word problem by first generating a detailed "solution\_pad" field that outlines all intermediate steps and calculations, then providing the final answer in the "final\_answer" field. The JSON output must strictly follow this structure: \{"solution\_pad": "<step-by-step reasoning with calculations>", "final\_answer": "<correct final answer>"\}.\\[4pt]
For problems involving multiple steps, ensure each calculation is explicitly written in the solution\_pad (e.g., "3/4 * 40,000 = <<3/4*40000=30000>>30000"). For percentage or ratio problems, show all conversion steps. For multi-part problems, break down each component into separate calculations. Only after fully documenting the reasoning process should the final\_answer be provided. The final\_answer must match the exact numerical value and formatting specified in the problem's expected solution.\\[4pt]
\hl{**Strict JSON Formatting Requirements**:}\\
\hl{1. All strings must be enclosed in double quotes ("), not single quotes (').}\\
\hl{2. Special characters (e.g., mathematical symbols, parentheses) must be properly escaped if needed.}\\
\hl{3. Validate the JSON structure before outputting to ensure no syntax errors (e.g., missing commas, unmatched braces).}\\
\hl{4. Use the exact field names "solution\_pad" and "final\_answer" in the JSON output.}\\
\hl{5. Avoid any trailing commas or incomplete JSON fragments.}\\[4pt]
\hl{Example of valid JSON:}\\
\hl{\{"solution\_pad": "Step 1: Calculate 1/3 of 18. 18 * (1/3) = <<6>>6. Step 2: Subtract from total. 18 - 6 = <<12>>12.", "final\_answer": "12"\}}
}\\[0pt]
\end{tcolorbox}

\begin{tcolorbox}[
title={\textbf{Minimal Seed, VISTA, Iteration 6 \hfill Parent: Iteration 5}},
fonttitle=\bfseries\small,
colback=gray!2,
colframe=gray!50,
boxrule=0.8pt,
breakable,
enhanced,
left=6pt, right=6pt, top=4pt, bottom=4pt
]
\small
Strategy: \texttt{heuristics}\\[4pt]
\begin{tabular}{@{}llr@{}}
\toprule
\textbf{Hyp.} & \textbf{Label} & \textbf{$\Delta$acc} \\
\midrule
H1 & \texttt{format\_and\_syntax} & +3.0 \textcolor{green!60!black}{\faCheck} \\
H2 & \texttt{reasoning\_strategy} & +2.0 \\
H3 & \texttt{edge\_case\_handling} & +2.0 \\
\bottomrule
\end{tabular}\\[4pt]
\textcolor{green!60!black}{Selected: H1 [\texttt{format\_and\_syntax}] --- Candidate Accepted}\\[2pt]
\hrule\vspace{8pt}
\small
{\fontsize{8}{9}\selectfont\ttfamily
Solve the mathematical word problem by first generating a detailed "solution\_pad" field that outlines all intermediate steps and calculations, then providing the final answer in the "final\_answer" field. The JSON output must strictly follow this structure: \{"solution\_pad": "<step-by-step reasoning with calculations>", "final\_answer": "<correct final answer>"\}.\\[4pt]
For problems involving multiple steps, ensure each calculation is explicitly written in the solution\_pad (e.g., "3/4 * 40,000 = <<3/4*40000=30000>>30000"). For percentage or ratio problems, show all conversion steps. For multi-part problems, break down each component into separate calculations. Only after fully documenting the reasoning process should the final\_answer be provided. The final\_answer must match the exact numerical value and formatting specified in the problem's expected solution, including decimal places, currency symbols, or unit labels.\\[4pt]
**Critical Clarifications**:\\
1. **Arithmetic Order**: Always perform operations in the correct sequence (PEMDAS/BODMAS). For subtraction-based problems (e.g., discounts, differences), explicitly write the operation as "original value - final value" to avoid reversed calculations. Example: "Discount = Original Price - Paid Price".\\
2. **Explicit Steps**: For any subtraction, addition, multiplication, or division, write the full expression with operands and operator first, then the result in the placeholder format. Example: "22 - 16 = <<22-16=6>>6".\\
3. **Unit Consistency**: Ensure all units are explicitly converted and documented in the solution\_pad if required (e.g., minutes to hours).\\
4. **Equation Solving**: For algebraic problems, isolate variables step-by-step, showing each transformation (e.g., "22 - x = 16 $\to$ x = 22 - 16").\\[4pt]
\hl{**Additional Requirements**:}\\
\hl{- The JSON output must be syntactically valid, with all opening and closing braces properly matched and no trailing commas.}\\
\hl{- The "solution\_pad" and "final\_answer" fields must be enclosed in double quotes and separated by a comma.}\\
\hl{- The final\_answer must exactly match the problem's expected solution, including formatting (e.g., currency symbols, decimal places, unit labels).}\\[4pt]
The final\_answer must match the exact numerical value and formatting specified in the problem's expected solution, including decimal places, currency symbols, or unit labels.
}\\[0pt]
\end{tcolorbox}

\end{document}